%% file: acl_latex.tex
\documentclass[11pt]{article}

\usepackage[preprint]{acl}

\usepackage{times}
\usepackage{latexsym}

\usepackage[T1]{fontenc}

\usepackage[utf8]{inputenc}

\usepackage{microtype}

\usepackage{inconsolata}

\usepackage{graphicx}

\input{preamble}

\title{TEVI: Text-Conditioned Editing of Visual Representations via Sparse Autoencoders for Improved Vision-Language Alignment}

\author{
Sweta Mahajan$^{*,1}$, Sukrut Rao$^{*,1}$, Jiahao Xie$^1$, Alexander Koller$^{1,2}$, Bernt Schiele$^1$ \\
$^1$Max Planck Institute for Informatics, Saarland Informatics Campus, Saarbrücken, Germany \\
$^2$Department of Language Science and Technology, Saarland University, Saarbrücken, Germany \\
\texttt{\{sweta.mahajan,sukrut.rao,schiele\}@mpi-inf.mpg.de} \\
\footnotesize{$^*$Equal contribution}
}

\begin{document}
\maketitle

\input{sec/0_abstract} 
\begin{figure*}[!ht]
    \centering
    \begin{subfigure}[c]{0.38\linewidth}
        \includegraphics[width=\linewidth]{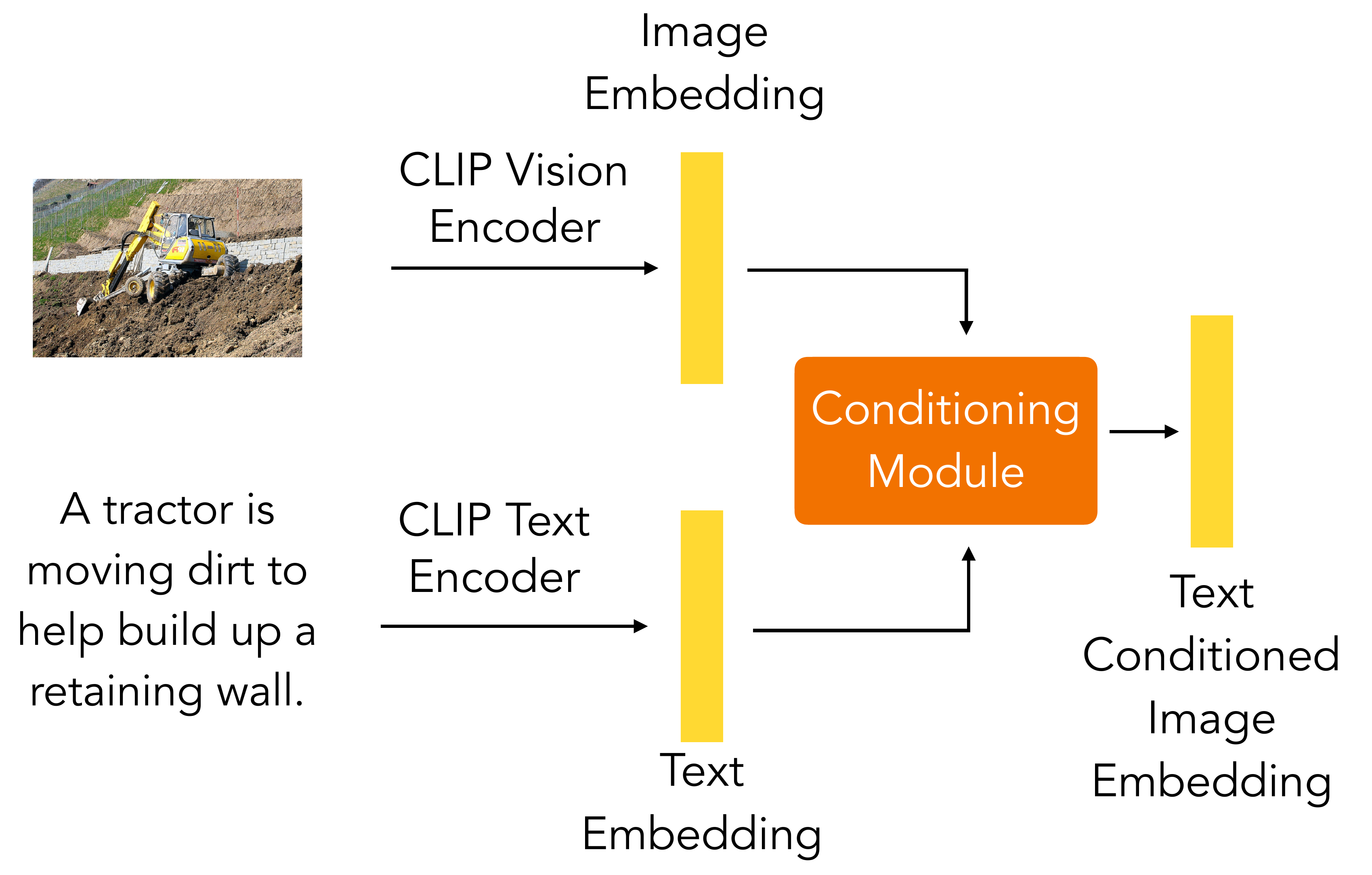}
        \label{fig:teaser:overview_left}
    \end{subfigure}
    \hfill
    \begin{subfigure}[c]{0.40\linewidth}
        \includegraphics[width=\linewidth]{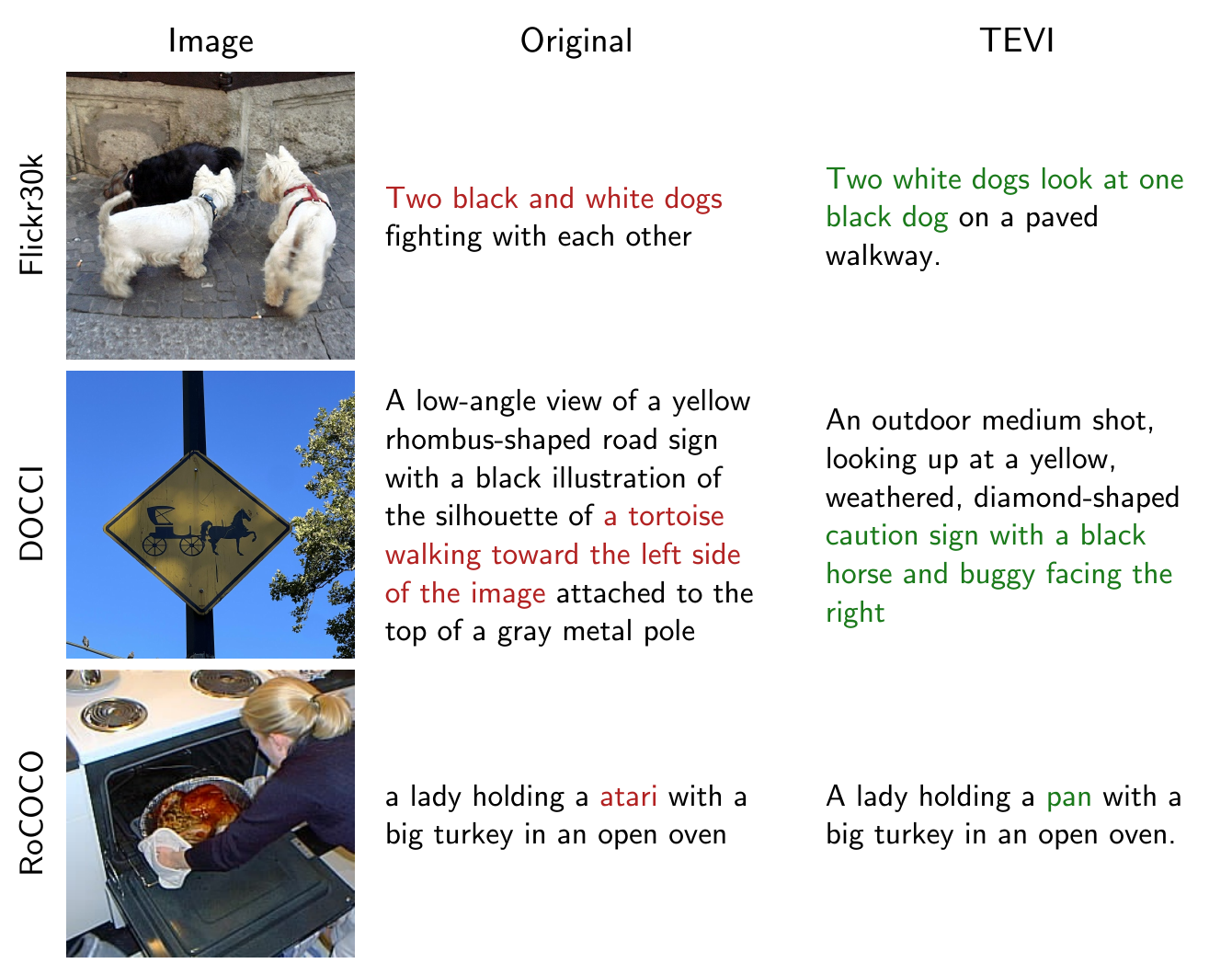}
        \label{fig:teaser:overview_right}
    \end{subfigure}
    \hfill
    \begin{subfigure}[c]{0.18\linewidth}
        \includegraphics[width=\linewidth]{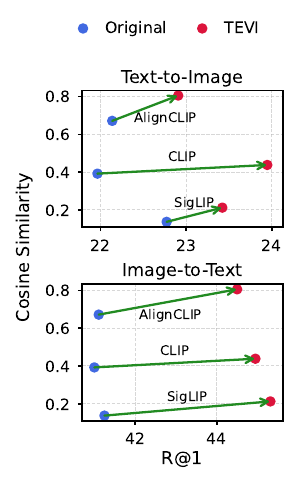}
        \label{fig:teaser:t2i}
        \label{fig:teaser:i2t}
    \end{subfigure}
    \caption{\textbf{\ours: using captions to edit image embeddings.} \textit{Left:} An overview of our approach. We use text captions as a signal to modify image embeddings from CLIP~\cite{radford2021learning}. For details, see \cref{fig:method:lmaskedit}. \textit{Middle:} Qualitative examples of top texts retrieved for an image by CLIP and \ours.  \textit{Right:} \ours helps improve cross-modal alignment as well as downstream retrieval performance. We report mean performance across datasets, for full results see \cref{tab:retrieval:coco_flickr,tab:retrieval:docci_iiw,fig:alignment_plot_cc12m}.
    }
    \label{fig:teaser}
\end{figure*}
\input{sec/1_intro}
\input{sec/2_related}
\input{sec/3_background}
\input{sec/4_toy}

\input{sec/5_real}
\input{sec/6_conclusion}
{
    \bibliography{main}
}
\clearpage
\input{supplement/0_main}

\end{document}

%% file: preamble.tex
\usepackage[table]{xcolor}

\definecolor{cvprblue}{rgb}{0.21,0.49,0.74}
\usepackage{amsmath}
\usepackage{amsfonts}
\usepackage[capitalize]{cleveref}

\usepackage{xspace}
\crefname{section}{Sec.}{Secs.}
\Crefname{section}{Section}{Sections}
\crefname{table}{Tab.}{Tabs.}
\Crefname{table}{Table}{Tables}
\makeatletter
\DeclareRobustCommand\onedot{\futurelet\@let@token\@onedot}
\def\@onedot{\ifx\@let@token.\else.\null\fi\xspace}
\def\eg{e.g\onedot} 
\def\ie{i.e\onedot}

\makeatother

\usepackage{changepage}
\usepackage{enumitem}
\usepackage{multirow}
\usepackage{float}
\usepackage{stfloats}
\usepackage{placeins}
\usepackage{balance}
\usepackage{subcaption}
\usepackage{booktabs}

\usepackage{etoolbox}
\usepackage[export]{adjustbox}
\usepackage[font=small]{caption}
\usepackage{bbm}
\usepackage{multirow}
\usepackage{hhline}

\makeatletter
\DeclareRobustCommand\onedot{\futurelet\@let@token\@onedot}
\def\@onedot{\ifx\@let@token.\else.\null\fi\xspace}
\def\eg{e.g\onedot} 
\def\ie{i.e\onedot}

\makeatother

\newcommand{\myparagraph}[1]{\vspace{3pt}\noindent{\bf #1}}

\usepackage{array}
\newcommand{\PreserveBackslash}[1]{\let\temp=\\#1\let\\=\temp}
\newcolumntype{C}[1]{>{\PreserveBackslash\centering}p{#1}}
\newcolumntype{R}[1]{>{\PreserveBackslash\raggedleft}p{#1}}
\newcolumntype{L}[1]{>{\PreserveBackslash\raggedright}p{#1}}

\makeatletter
\newcommand\footnoteref[1]{\protected@xdef\@thefnmark{\ref{#1}}\@footnotemark}
\makeatother

\usepackage{afterpage}

\usepackage{pifont}

\newcommand{\ours}{TEVI\xspace}

\definecolor{hirow}{HTML}{e6ffe6}

\definecolor{forestgreen}{rgb}{0.0, 0.50, 0.13}

\usepackage{tikz}
\definecolor{iconink}{HTML}{444444}
\newsavebox{\iconboxa}\newsavebox{\iconboxb}
\newsavebox{\iconboxc}\newsavebox{\iconboxd}
\sbox{\iconboxa}{\tikz[baseline=-0.55ex]{\fill[iconink] (0,0) circle (0.24em);}}
\sbox{\iconboxb}{\tikz[baseline=-0.55ex]{\fill[iconink] (-0.213em,-0.213em) rectangle (0.213em,0.213em);}}
\sbox{\iconboxc}{\tikz[baseline=-0.55ex]{\draw[iconink,line width=0.075em] (0,0) circle (0.2025em);}}
\sbox{\iconboxd}{\tikz[baseline=-0.55ex]{\draw[iconink,line width=0.075em] (-0.176em,-0.176em) rectangle (0.176em,0.176em);}}
\newcommand{\iconcirc}{\usebox{\iconboxa}}
\newcommand{\iconsq}{\usebox{\iconboxb}}

\newcommand{\std}[1]{\,{\tiny$\pm$#1}}

\usepackage[normalem]{ulem}

%% file: sec/0_abstract.tex
\begin{abstract}
Vision-language models such as CLIP are highly useful for diverse tasks due to their shared image-text embedding space. Despite this, the image and text embeddings are often poorly aligned, affecting downstream performance. Recent work has hypothesized that this can be attributed to an information imbalance: images contain more information than their captions describe. In this work, we propose \ours, a framework that uses captions as a signal for what to retain from image embeddings. Specifically, we use sparse autoencoders to disentangle image embeddings and train a masking module to selectively reconstruct the embedding based on a given caption. In a controlled setup with synthetic captions, we show that \ours is effective at preserving caption-described attributes while discarding others. We find that this extends to CLIP models trained on natural images, where \ours learns to mask meaningfully and allows retrieval based on conditioning. Finally, we use \ours to achieve improved retrieval performance across coarse-grained and fine-grained benchmarks. Code available at \href{https://github.com/neuroexplicit-saar/TEVI}{https://github.com/neuroexplicit-saar/TEVI}.
\end{abstract}

%% file: sec/1_intro.tex
\section{Introduction}
\label{sec:introduction}

Vision-language models such as the CLIP family~\cite{radford2021learning,ilharco_gabriel_2021_5143773,zhai2023sigmoid,tschannen2025siglip} are trained using a contrastive loss to align images and text to a shared embedding space\footnote{Throughout this work, we use ``vision-language model'' to refer to CLIP-like models with aligned image-text encoders, and not autoregressive multimodal models that generate text.}. Such models have been widely successful for a variety of multimodal applications, such as zero-/few-shot classification and cross-modal retrieval. Their embeddings serve as a vital bridge between vision and language for generative models such as text-to-image diffusion models~\cite{razzhigaev2023kandinsky} and large multimodal models (LMMs)~\cite{liu2023visual}. However, it has been shown that the learnt embedding space suffers from a modality gap~\cite{liang2022mind}, where image and text embeddings lie in different regions of the embedding space, often leading to poor downstream performance~\cite{liang2022mind,eslami2024mitigate,schrodi2024two}.

While various possible causes have been discussed~\cite{eslami2024mitigate,schrodi2024two}, a recent work~\cite{schrodi2024two} hypothesizes that this is caused by an \emph{information imbalance}---the fact that images contain more information than their corresponding captions, which forces models to push their embeddings apart when minimizing training loss---and shows evidence for this hypothesis via systematically controlled experiments on the synthetic MAD dataset. However, despite finding that only a few embedding dimensions characterize the gap, removing these dimensions also significantly degrades performance. Since CLIP dimensions do not directly correspond to concepts, this may not map to selectively removing information as the hypothesis suggests.

In this work, we explore whether captions could instead determine what information should be preserved, by using text as a signal to modify CLIP image embeddings to preserve only what the caption describes. Specifically, we use sparse autoencoders (SAEs)~\cite{bricken2023monosemanticity,cunningham2023sparse} to decompose image embeddings into constituent concepts, and train a conditioning module that selects SAE latents to be used for reconstruction based on the conditioning.
Through a controlled setup, we show that our proposed approach, \underline{t}ext-conditioned \underline{e}diting of 
\underline{v}isual representations for \underline{i}mage-text alignment (\textbf{\ours}) (\cref{fig:teaser}, left), can learn to preserve attributes presented in the text while discarding information about other attributes and improve cross-modal alignment and retrieval ordering (\cref{fig:mad_ordering_combined}). We then apply \ours to CLIP~\cite{radford2021learning,ilharco_gabriel_2021_5143773}, SigLIP~\cite{zhai2023sigmoid}, SharedCLIP, and AlignCLIP~\cite{eslami2024mitigate} models trained on CC12M~\cite{changpinyo2021conceptual}. We find that \ours learns to selectively and meaningfully mask latents based on the conditioning, which helps retrieval with known attributes (\cref{fig:coco_qualitative}). Further, \ours improves (\cref{fig:teaser}, middle and right) image-to-text and text-to-image retrieval performance across coarse-grained (MS COCO~\cite{lin2014microsoft}, Flickr30k~\cite{plummer2015flickr30k}) and fine-grained (DOCCI~\cite{onoe2024docci}, IIW~\cite{garg2024imageinwords}) benchmarks, by serving as an effective reranking method of retrieved candidates from the base model. Using the RoCOCO~\cite{park2024rococo} benchmark, which augments captions with linguistically perturbed alternatives, we also show that \ours may help improve robustness against such perturbations. On the whole, we find that having a concept basis makes the information imbalance controllable.

\noindent In summary, our \textbf{contributions} are as follows:
\begin{itemize}
    \item CLIP-Guided SAEs (CG-SAEs), a controlled setup with latents that encode predefined text concepts, to study their utility for preserving targeted concepts.
    \item \ours, a framework that uses text as a signal to selectively retain caption-described content in image embeddings, via a learnt mask over sparse autoencoder latents.
\end{itemize}

\noindent We use CG-SAEs to show a proof-of-concept of our proposed approach through controlled experiments on MAD~\cite{schrodi2024two}. We then use \ours with MAD to show that it leads to improved alignment and retrieval, and then show that this extends to CLIP models trained on natural images.

%% file: sec/2_related.tex
\section{Related Work}
\label{sec:related}

\myparagraph{Vision-Language Models (VLMs)}~\cite{radford2021learning,zhai2023sigmoid,ilharco_gabriel_2021_5143773,yu2022coca,jia2021scaling,tschannen2025siglip} learn a joint aligned embedding space between images and texts. They typically consist of separate unimodal image and text encoders that each provide an embedding, and are trained using contrastive losses so that embeddings of similar image-text pairs are placed close to each other and dissimilar pairs are placed apart. Such models are useful for a diverse set of multimodal tasks such as cross-modal retrieval and zero-shot classification. Their embeddings are also used as a bridge between vision and language for text-to-image diffusion models~\cite{razzhigaev2023kandinsky} and large multimodal models (LMMs)~\cite{liu2023visual}. In this work, we focus on CLIP-family models.

\myparagraph{Modality Gap}~\cite{schrodi2024two,liang2022mind,eslami2024mitigate,mistretta2025cross,shi2023towards} is a phenomenon observed in trained VLMs, where, despite the contrastive training objective, image and text embeddings lie in different regions of the embedding space. While initially attributed to the `cone effect' at initialization~\cite{liang2022mind}, recent work~\cite{schrodi2024two} showed that a likely cause is the information imbalance between the two modalities, \ie, images contain more information than is described in their corresponding caption, which forces the model to push apart their embeddings to reduce the contrastive loss. Existing post hoc approaches to reduce the modality gap~\cite{liang2022mind,schrodi2024two} have been shown to come at the cost of degraded performance. Recently, AlignCLIP~\cite{eslami2024mitigate} proposed to use an intra-modal separation loss during CLIP training to improve alignment and downstream performance. In contrast, we explicitly control the information imbalance post hoc by using text as a signal to edit image embeddings, and show that this is complementary to methods such as AlignCLIP.
\input{figures/tex/2_method}

\myparagraph{Cross-Modal Conditioning} methods such as SmartCLIP~\cite{xie2025smartclip}, FLAIR~\cite{xiao2024flair}, and FILIP~\cite{yao2021filip} produce image embeddings conditioned on text for improved fine-grained retrieval, at the cost of a small increase in retrieval time. Concurrently, SteerViT~\cite{ruthardt2026steerable} conditions a frozen vision encoder on text by inserting cross-attention layers into its blocks, which steers the representation towards the objects that a prompt names. An orthogonal line of work \cite{xie2025fg,asokan2025finelip} performs fine-grained alignment at training without conditioning at inference. Among the former methods, all except SmartCLIP and SteerViT train models from scratch. In contrast, \ours is post-hoc: it adds a small module on top of pretrained CLIP, with the vision encoder kept frozen and only the text encoder fine-tuned. Similar to SmartCLIP, \ours learns a mask over vision embeddings conditioned on text, but in contrast, \ours masks disentangled SAE latents instead of raw representations. 

\myparagraph{Sparse Autoencoders (SAEs)} \cite{bricken2023monosemanticity,cunningham2023sparse,gao2024scaling,rajamanoharan2024improving,rajamanoharan2024jumping,bussmann2025learning} are a popular mechanistic interpretability tool to disentangle activations learnt by a deep network into constituent human understandable concepts. While originally used in the context of LLMs~\cite{bricken2023monosemanticity,cunningham2023sparse}, SAEs have recently been used to decompose concepts from CLIP vision embeddings~\cite{rao2024discover,zaigrajew2025interpreting} for use in downstream tasks such as constructing concept bottleneck models~\cite{rao2024discover}. SAEs have also been used as a tool for steering models by performing edits to their latents \cite{farrell2024applying,cywinski2025saeuron,pach2026sparse,joseph2025steering}. In contrast to these works which apply SAE-based interventions for interpretability or steering, we use them for text-conditioned editing aimed at improving cross-modal alignment.

%% file: figures/tex/2_method.tex
\begin{figure*}[!ht]
    \centering
    \includegraphics[trim=0 6cm 0 0,clip=false,width=0.7\linewidth]{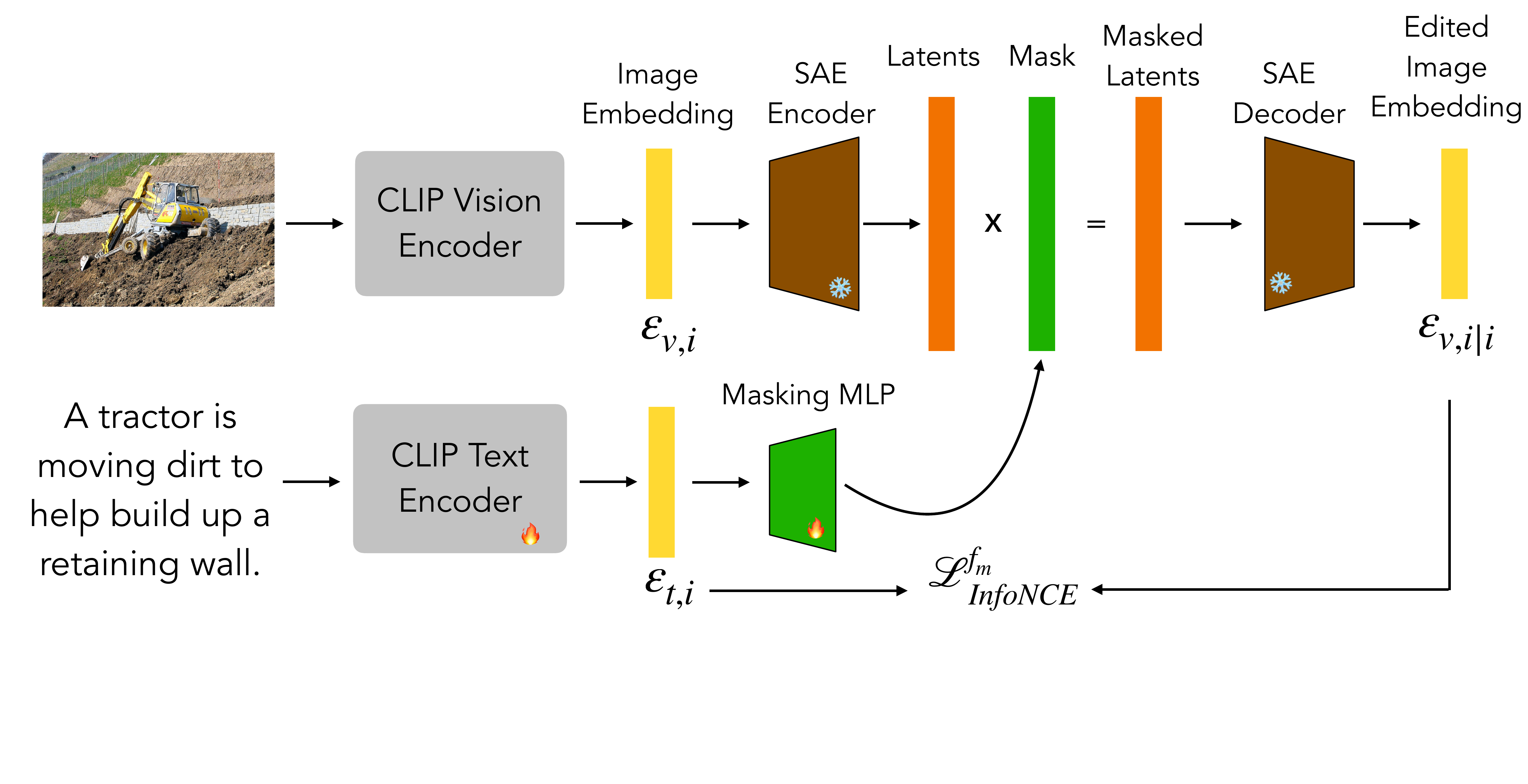}
    \caption{\textbf{Our proposed \ours framework for obtaining text-conditioned image embeddings.} We train a TopK SAE~\cite{gao2024scaling} over CLIP image embeddings, and then use an MLP trained using the InfoNCE loss to learn a mask over the SAE latents to obtain conditioned image embeddings. For details, see \cref{sec:method,sec:real_data}. 
    }
    \label{fig:method:lmaskedit}
\end{figure*}

%% file: sec/3_background.tex
\section{Using SAEs to Edit Representations}
\label{sec:background}
In this section, we motivate our text-conditioned image editing approach. In \cref{sec:background:preliminaries}, we first provide a brief overview on CLIP~\cite{radford2021learning} and sparse autoencoders (SAEs)~\cite{bricken2023monosemanticity}. Then, to test whether SAEs can be used to edit representations, we describe a controlled test on a synthetic setup in \cref{sec:background:sanity} and present our findings in \cref{sec:background:results}. We then present our proposed \ours in \cref{sec:method}, and in \cref{sec:real_data}, we show how we extend it to real data.

\subsection{Background}
\label{sec:background:preliminaries}

\paragraph{CLIP~\cite{radford2021learning}.} Let $\mathcal{D}=\{(v_i,t_i)\}_{i=1}^N$ be a paired dataset of images $v_i$ and their corresponding texts $t_i$. A CLIP model $M=(f_v,f_t)$ consists of a vision and text encoder respectively which provide corresponding embeddings $(\varepsilon_{v,i}, \varepsilon_{t,i})$, \ie $\varepsilon_{v,i}=f_v(v_i)\in \mathbb R^d$ and $\varepsilon_{t,i}=f_t(t_i)\in \mathbb R^d$. The model is then trained with a contrastive InfoNCE loss~\cite{oord2018representation}:
\begin{multline}
\label{eq:cliploss}
\mathcal{L}_{\text{InfoNCE}} = 
-\frac{1}{2N} \sum_{i=1}^N \Bigg[
    \log \frac{
        e^{\hat{\varepsilon}_{v,i} \cdot \hat{\varepsilon}_{t,i} / \tau}
    }{
        \sum_{j=1}^N e^{\hat{\varepsilon}_{v,i} \cdot \hat{\varepsilon}_{t,j} / \tau}
    }
\\
    + 
    \log \frac{
        e^{\hat{\varepsilon}_{v,i} \cdot \hat{\varepsilon}_{t,i} / \tau}
    }{
        \sum_{j=1}^N e^{\hat{\varepsilon}_{v,j} \cdot \hat{\varepsilon}_{t,i} / \tau}
    }
\Bigg]
\end{multline}
where $\tau$ is a learnable temperature hyperparameter and $\hat{x}=\frac{x}{\lVert x \rVert_2}$. \cref{eq:cliploss} pulls embeddings of corresponding (positive) image-text pairs $(\varepsilon_{v,i}, \varepsilon_{t,i})$ close to each other and pushes embeddings of other (negative) image-text pairs $(\varepsilon_{v,i}, \varepsilon_{t,j}), i\ne j$ away from each other to learn a shared semantic embedding space.

\paragraph{Sparse Autoencoders~\cite{bricken2023monosemanticity}} consist of a linear encoder $W_E\in\mathbb R^{d\times d_1}$ and decoder $W_D\in\mathbb R^{d_1\times d}$ where typically $d_1\gg d$. The encoder maps an input $x\in\mathbb R^d$ to latents $z=\text{ReLU}(W_E^T(x-b_{pre})+b_E)$, where $b_{pre}\in\mathbb R^d, b_{E}\in\mathbb R^{d_1}$ are learnt biases and $z$ is a sparse disentangled concept representation of $x$. The decoder then reconstructs $x$ using $z$, \ie $\tilde{x} = W_D^Tz+b_{pre}$. The SAE is trained with a combination of a reconstruction loss $\mathcal{L}_{\text{recon}}=\lVert x-\tilde{x} \rVert_2$ and sparsity loss $\mathcal{L}_{\text{sparse}}$, \ie
\begin{equation}
\label{eq:sae}
\mathcal{L}_{\text{SAE}}=\mathcal{L}_{\text{recon}} + \lambda_{\text{SAE}}\mathcal{L}_{\text{sparse}},
\end{equation}
where $\lambda_{\text{SAE}}$ is a hyperparameter.
For sparsity, we use $\text{TopK}(z)$~\cite{gao2024scaling} where $\text{TopK}(\cdot)$ selects the top $K$ activated latents in $z$ for hyperparameter $K$.

\subsection{Controlled Setup for Evaluating Representation Editing}
\label{sec:background:sanity}

To understand whether subselecting SAE latents before reconstruction can effectively edit embeddings, we construct a controlled setup using synthetic data with known attributes, where we induce each SAE latent to represent a predefined attribute. In contrast to a typical setup where SAEs automatically learn to disentangle concepts, this is designed to allow for targeted editing, and we refer to it as CLIP-guided SAE (CG-SAE). Let $C=\{c_j\}_{j=1}^S$ be a set of text concepts, and let $(W_E,W_D)$ be an SAE to be trained on image embeddings of CLIP. Then, for an embedding $\varepsilon_{v,i}$,  
\begin{equation}
\label{eq:z}
    z_i = \text{TopK}(\text{ReLU}(W_E^T(\varepsilon_{v,i}-b_{pre})+b_E))
\end{equation}
\begin{equation}
    \widetilde{\varepsilon_{v,i}} = W_D^Tz_i + b_{\text{pre}}
\end{equation}
where $z_i=[z_{i,r}]_{r=1}^{d_1}$ such that elements in $z_i$ outside the top $K$ elements are zeros.

Recent work~\cite{rao2024discover} showed that latents $z_{i,r}$ could be assigned meaningful concept names in $c_q\in C$ post hoc by selecting the text embedding that is closest to their corresponding decoder weight vector $W_{D,r}$, \ie  
\begin{equation}
    q = \arg\max_j \cos(W_{D,r}; \hat{\varepsilon}_{t,c_j})\,.
\end{equation}
Inspired by this, for our conditioning, we propose to attempt the opposite---given $C$, we fix the rows of the decoder weights $W_D$ to be text embeddings of concepts $c_j\in C$, \ie 
\begin{equation}
    W_{D,r} = \hat{\varepsilon}_{t,c_r}
\end{equation}
where $1\le r\le S$ and $d_1=S$, and then only train the SAE encoder.

Once trained, given an attribute set $C_A=\{c_{q}\}\subseteq C$, we select latents \emph{not} present in the text, \ie $\bar{C}=C\setminus C_A$ and set their activations to zero, \ie
\begin{equation}
\label{eq:cgsae0}
    z_{p|q} = [z_p]_{z_{p,c_{q'}} = \vec{0}},\,\,   \forall c_{q'}\in\bar{C}
\end{equation}
The edited image embedding of an image $v_p$ conditioned on text $t_q$ is then given by $\widetilde{\varepsilon_{v, p|q}}$, where
\begin{align}
    \widetilde{\varepsilon_{v,p|q}} &= W_D^Tz_{p|q} + b_{\text{pre}}\,.
\end{align}
We evaluate the following: \textbf{(R1)}
if each CG-SAE latent indeed encodes the concept assigned to it, and \textbf{(R2)} if masking out specific attributes removes information about that attribute from the reconstructed embedding from the SAE. We present our findings in \cref{sec:background:results}.

\input{figures/tex/auc_roc}

\subsection{Results}
\label{sec:background:results}
\myparagraph{Experimental Setup.} We use the synthetic MAD dataset~\cite{schrodi2024two}, which consists of MNIST digits with colors and morphological transforms. Every image is characterized by six attributes---digit, color, thickthinning, swelling, scaling, and fracture---each of which can be one among a set of predefined options, which are 26 in total. For example, the `thickthinning' attribute for an image could be `thickening', `thinning' or `nothickthinning', depending on the transform applied to the image.
For each image, we create captions that contain the digit and three randomly selected attributes out of the remaining five to simulate information imbalance~\cite{schrodi2024two}.
Following~\cite{schrodi2024two}, we train CLIP models~\cite{radford2021learning,ilharco_gabriel_2021_5143773} consisting of a 6-layer ViT~\cite{dosovitskiy2021an} as the vision encoder and a 6-layer transformer as the text encoder, with a shared embedding dimension of 18, for 200 epochs using the AdamW~\cite{loshchilov2017decoupled} optimizer. Given the 26 possible attributes per image, the text encoder uses a tokenizer with a vocabulary of size 28, after accounting for the start and end tokens. Then, we train TopK SAE~\cite{gao2024scaling} on CLIP image embeddings with a latent dimension of 26 (equal to the number of attributes for the dataset). We assign each of the decoder weights as the normalized text embedding of one of the attribute values, so that each SAE latent represents one of the 26 attributes. For the SAE, we ablate across values of $K$, expansion factors, and learning rates, and select models based on reconstruction accuracy and classification performance across the six attributes. 

\input{figures/tex/learntmask_freeform}

\myparagraph{Concept Disentanglement (R1).}
We evaluate if our CG-SAE learns to disentangle image embeddings into the fixed set of predefined concepts $C$. In \cref{fig:auc_roc}, we show  the area under the receiver operating characteristic (ROC) curve (AUC) across images in the dataset for the two latents assigned to the `swelling' attribute. We find that the latents are able to disentangle the attributes well, with the AUC being close to 1 for the attribute value the latent is assigned to, 0.5 for unrelated attributes, and 0 for attribute values anti-correlated to the assigned attribute (\ie `no swelling', for the `swelling' latent). We also show qualitative examples of top activating images for a selection of latents in \cref{fig:madqualitative}, and find that they are highly consistent, \eg for the `blue' latent, the top activating images are all digits of the color blue, while also being diverse in all other attributes.

\myparagraph{Editing Image Representations (R2).} We evaluate if masking latents corresponding to a single attribute discards information about that attribute from the reconstructed representations. To do this, we pick a single attribute `thickthinning', and set the three latents corresponding to it (`thickening', `thinning', `nothickthinning') to zero. This is similar to conditioning on all attributes $q$ except `thickthinning' (\cref{eq:cgsae0}). We then obtain edited embeddings for all images, and perform classification on all six attributes based on the cosine similarity with the text embeddings of the classes within the attribute, similar to zeroshot classification. For example, for `Thickthinning', given an edited embedding $\varepsilon_{v,p|q}$, $\text{Pred}_{\text{thickthinning}} = \arg\max \cos(\varepsilon_{v,p|q}; \varepsilon_{t,j})$,
where $t_j\in\{\text{thickening}, \text{thinning}, \text{nothickthinning}\}$. We find (\cref{fig:conditioned_accuracy_latent_ablate_cossim}, left) that the classification accuracy of the edited embeddings on `thickthinning' drops to near random chance while that of all other attributes remains close to that with the original embedding. This shows that our conditioning can be an effective way to remove information present in the image embeddings that is not in the text.

%% file: figures/tex/auc_roc.tex
\begin{figure}[!t]
    \centering
    \includegraphics[trim={0 0 0 1.8cm},clip,width=0.9\linewidth]{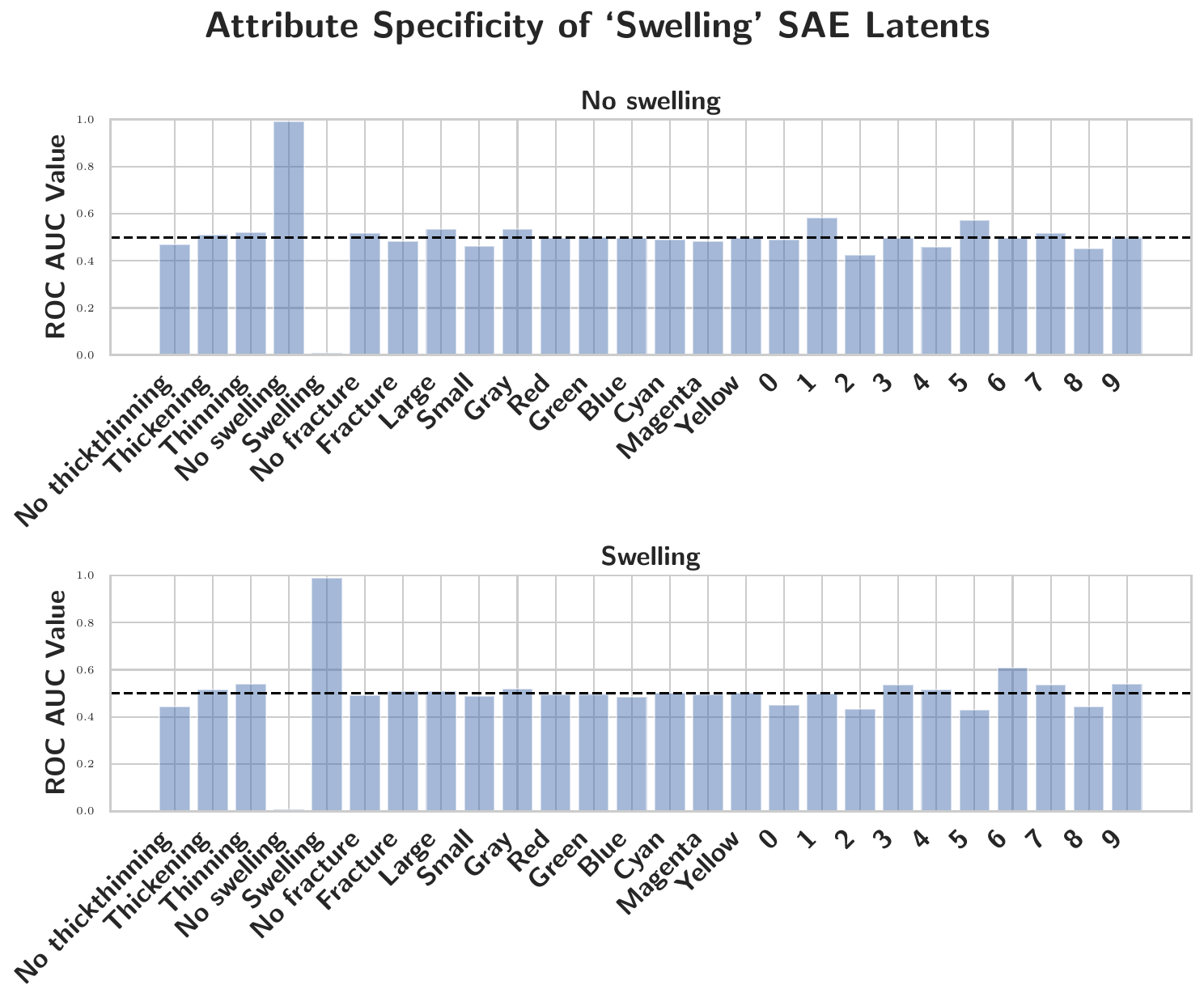} 
    \caption{\textbf{Attribute specificity of CG-SAE latents.} We plot the AUC for CG-SAE latents corresponding to the attribute values `No swelling' and `Swelling'. We find that the latents are highly attribute specific, which shows that our CG-SAE latents are highly disentangled, despite being assigned to a predefined concept. Full results in \cref{supp:sec:mad:disentanglement}.}
    \label{fig:auc_roc}
\end{figure}

%% file: figures/tex/learntmask_freeform.tex
\begin{figure*}[!t]
    \centering
    
    \begin{subfigure}[c]{0.35\linewidth}
    \centering
    \includegraphics[trim={0 0 0 0.25cm},clip,width=\linewidth]{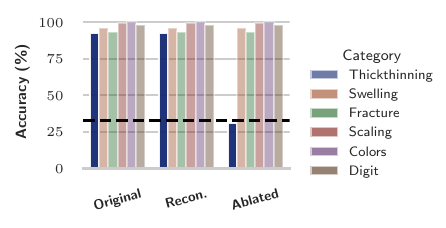}

    \end{subfigure}
    \hfill
    \begin{subfigure}[c]{0.38\linewidth}
        \centering
        \includegraphics[width=\linewidth]{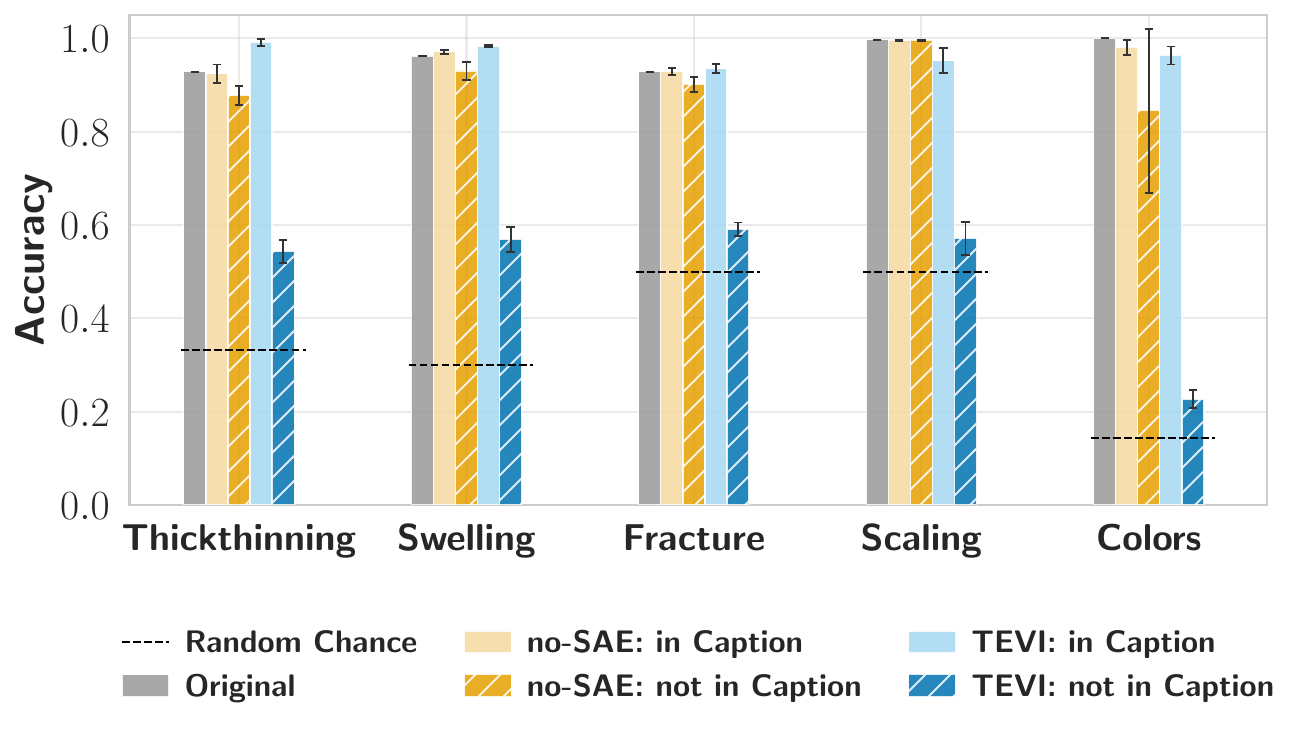}
    \end{subfigure}
    \begin{subfigure}[c]{0.25\linewidth}
        \centering
        \includegraphics[width=\linewidth]{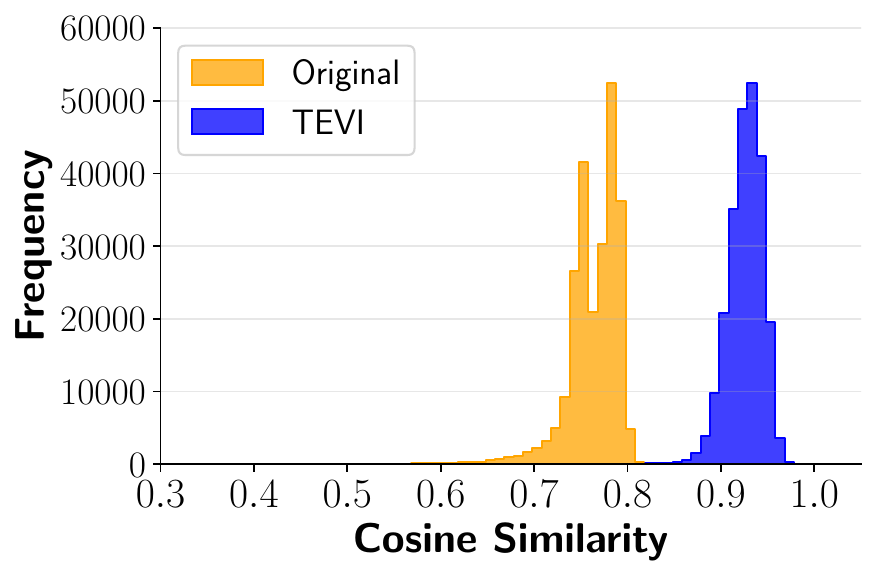}
    \end{subfigure}
    \hfill
    \caption{\textit{Left:}  \textbf{Accuracy after ablating a single latent.} Accuracy for `Thickthinning' drops to random chance (dotted line) when editing image embeddings to discard information about that attribute (right group),  while the other attributes continue to maintain high accuracy (\cref{sec:background}). \textit{Middle:} \textbf{Effectiveness of conditioning.} Attributes present in the conditioning text are preserved in the edited embedding, while attributes that are absent are classified at close to random chance accuracy (\cref{sec:method}), and the no-SAE baseline shows almost no effect. \textit{Right:} \textbf{Impact on vision-language alignment.} Pairwise image-text cosine similarities between positive pairs increase  on applying \ours (\cref{sec:method}).}
    \label{fig:conditioned_accuracy_latent_ablate_cossim}
\end{figure*}

%% file: sec/4_toy.tex
\input{figures/tex/mad_ordering_combined}

\section{\ours: Learning to Mask Latents based on Conditioned Text}
\label{sec:method}

In \cref{sec:background:sanity}, we showed that we could construct CG-SAEs for encoding and editing representations given predefined synthetic concepts. However, this is a restrictive assumption, since CLIP models are typically open vocabulary and a predefined set of concepts is not available. As a result, in this section, we relax this assumption and propose \ours, an approach to learn a mask over SAE latents $z$ to obtain the text-conditioned image embeddings. An overview of our approach is shown in \cref{fig:method:lmaskedit}.

\subsection{Optimization Objective}
Specifically, we keep the CLIP vision encoder and a trained SAE frozen, and train a small multi-layer perceptron (MLP) network consisting of linear transforms with ReLU activations $f_m: \mathbb R^d \to \mathbb R^{d_1}$ that maps text embeddings $\varepsilon_{t,i}$ to a mask $m_i\in [0,1]^{d_1}$,
\ie $m_i = \sigma(f_m(\varepsilon_{t,i}))$
where $\sigma(\cdot)$ is the sigmoid function.

Then, the edited image embedding of an image $v_p$ conditioned on text $t_q$ is given by $\varepsilon_{v, p|q}$, where
\begin{equation}
\label{eq:editours}
    \varepsilon_{v,p|q} = W_D^T(z_{p}\odot m_{q}) + b_{\text{pre}}
\end{equation}
where
$\odot$ is the element-wise product.

We train $f_m$ and fine-tune the CLIP text encoder $f_t$ using the InfoNCE loss (\cref{eq:cliploss}) to pull the edited embeddings towards their conditioning texts and away from other texts, \ie
\begin{multline}
\label{eq:learntlossours}
\mathcal{L}^{f_m}_{\text{InfoNCE}} =
-\frac{1}{2N} \sum_{i=1}^N \Bigg[
    \log \frac{
        e^{\hat{\varepsilon}_{v,i|i} \cdot \hat{\varepsilon}_{t,i} / \tau}
    }{
        \sum_{j=1}^N e^{\hat{\varepsilon}_{v,i|i} \cdot \hat{\varepsilon}_{t,j} / \tau}
    }
\\
    +
    \log \frac{
        e^{\hat{\varepsilon}_{v,i|i} \cdot \hat{\varepsilon}_{t,i} / \tau}
    }{
        \sum_{j=1}^N e^{\hat{\varepsilon}_{v,j|j} \cdot \hat{\varepsilon}_{t,i} / \tau}
    }
\Bigg]\,.
\end{multline}

\subsection{Results}
\vspace{-4pt}
\myparagraph{Experimental Setup.}
We follow the setup from \cref{sec:background:results}; however, we no longer fix the decoder weights of SAE and keep them learnable. We use 3-layer MLPs for the masking module. As a baseline, we train a mask on CLIP image embeddings.

\myparagraph{Effectiveness of Conditioning.}
We evaluate the effectiveness of conditioning for \ours on actual captions. First, we perform attribute-wise classification. Since each caption has a random subset of attributes present in the image (\cref{sec:background:results}), we bin the attributes for each image depending on whether the caption contains it, and report the classification accuracy across the two bins in \cref{fig:conditioned_accuracy_latent_ablate_cossim} (middle). We find that (\textcolor{blue}{blue}) the classification accuracy on attributes \emph{present} in the caption remains similar to the accuracy from the original embeddings, while the accuracy from attributes \emph{absent} from the caption reach near random chance (dotted black line). In contrast, the no-SAE baseline (\textcolor{orange}{orange}) shows almost no effect. Then, in \cref{fig:mad_ordering_combined}, we evaluate the impact of conditioning on retrieval. We do this by picking a pair of attributes, selecting a query that contains both attributes, and conditioning on one while keeping the other as control. Before conditioning, candidates containing both attributes should match the strongest, and those that contain just one should match weaker but equally well. However, after conditioning, we expect the conditioned attribute to be matched stronger than the control attribute, whose information should be discarded from the representation (see \cref{fig:mad_ordering_combined}, top right). We find that this holds when using \ours, and significantly outperforms the no-SAE baseline.
This shows that \ours is effective in performing conditioning such that the edited embedding preserves information present in the text while discarding the remaining attributes.

\myparagraph{Vision-language Alignment.}
Following \citet{eslami2024mitigate}, we also evaluate the pairwise image-text cosine similarities between positive image-text pairs before and after conditioning and the plots are shown in \cref{fig:conditioned_accuracy_latent_ablate_cossim} (right). We find that for \ours, the pairwise similarities increase significantly after conditioning (\textcolor{blue}{blue}) as compared to before (\textcolor{orange}{orange}).

%% file: figures/tex/mad_ordering_combined.tex
\begin{figure*}[!t]
    \centering
    \begin{minipage}[c]{0.51\linewidth}
        \centering
        \includegraphics[width=\linewidth]{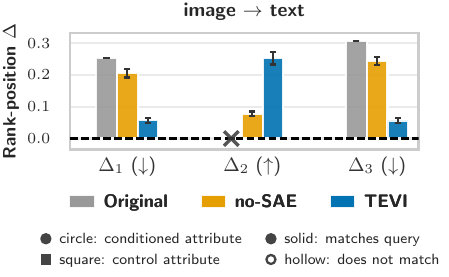}
    \end{minipage}\hfill
    \begin{minipage}[c]{0.47\linewidth}
        \centering
        \includegraphics[width=\linewidth]{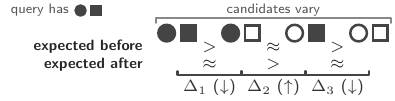}\\[0.1em]
        \includegraphics[width=\linewidth]{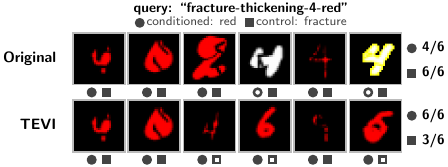}
    \end{minipage}
    \caption{\textbf{Conditioning reorders the retrieved set on MAD.} We condition on one attribute (\iconcirc) and use another as control (\iconsq), see top right. With a query containing both \iconcirc~and \iconsq~and varying candidates, we expect 
    $\Delta_1$ and $\Delta_3$ to fall towards zero and $\Delta_2$ to grow, as the edited embedding should preserve \iconcirc~and discard \iconsq. $\Delta$ denotes the difference in mean rank position between two groups of candidates, where each rank is normalized between 0 and 1.
    \textit{Left:} The rank-position differences $\Delta_1$, $\Delta_2$ and $\Delta_3$
    before and after conditioning, for I~$\rightarrow$~T retrieval. \ours moves all three in the expected direction, and significantly more than the no-SAE baseline. Full results in
    \cref{fig:mad_ordering_supp}.
    \textit{Right:} a T~$\rightarrow$~I example, showing the top-6 retrieved images
    before and after conditioning on `red'. The conditioned attribute is kept
    ($4/6 \rightarrow 6/6$) and the control attribute is dropped ($6/6 \rightarrow 3/6$).
    }
    \label{fig:mad_ordering_combined}
\end{figure*}

%% file: sec/5_real.tex
\input{tables/mask_real_data}

\input{tables/retrieval_results_fine}
\input{figures/tex/coco_qualitative}

\vspace{-6pt}
\section{Using \ours with Natural Images}
\label{sec:real_data}
In \cref{sec:method}, we showed the utility of our \ours on a synthetic setup. In this section, we extend to CLIP models trained on natural images with the goal of improving alignment and retrieval performance.

\subsection{Optimization Objective}
\label{sec:real_data:opt_obj}
Despite its effectiveness in the synthetic setup, the objective from \cref{eq:learntlossours} only uses edited image embeddings $\hat{\varepsilon}_{v,i|i}$ conditioned on \emph{their own} corresponding positive captions, \ie it does not model conditioning on negative captions. However, image-to-text and text-to-image retrieval involve selecting from a set of candidates, which would require conditioning both positive and negative pairs. So, we modify the objective to incorporate negative conditioning during training:
\vspace{-16pt}

\begin{multline}
\label{eq:learntlossours_real}
\mathcal{L}^{f_m}_{\text{InfoNCE}} =
-\frac{1}{2N} \sum_{i=1}^N \Bigg[
    \log \frac{
        e^{\hat{\varepsilon}_{v,i|i} \cdot \hat{\varepsilon}_{t,i} / \tau}
    }{
        \sum_{j=1}^N e^{\hat{\varepsilon}_{v,i|j} \cdot \hat{\varepsilon}_{t,j} / \tau}
    }
\\
    +
    \log \frac{
        e^{\hat{\varepsilon}_{v,i|i} \cdot \hat{\varepsilon}_{t,i} / \tau}
    }{
        \sum_{j=1}^N e^{\hat{\varepsilon}_{v,j|i} \cdot \hat{\varepsilon}_{t,i} / \tau}
    }
\Bigg]
\end{multline}

where \eg $e^{\hat{\varepsilon}_{v,i|i}}$ in the denominator of the first term from \cref{eq:learntlossours} is changed to $e^{\hat{\varepsilon}_{v,i|j}}$. Conditioning on negative captions is essential, without which the model performs poorly, which could be due to the fact that the model sees both positive and negative captions during inference. We discuss further in \cref{supp:sec:cc12m:without_negative_conditioning_retrieval}. 

\subsection{Results}
\label{sec:real_data:results}

\myparagraph{Experimental Setup.}
Following previous works \cite{eslami2024mitigate, goel2022cyclip, mu2022slip, li2021supervision}, we use the CC12M dataset~\cite{changpinyo2021conceptual} to train a CLIP~\cite{radford2021learning} ViT-B/16~\cite{dosovitskiy2021an}, CLIP ViT-L/14, and a SigLIP~\cite{zhai2023sigmoid} ViT-B/16 model.
To train the SAE, we ablate across values for the expansion factors and learning rates and select the best configuration based on the mean reconstruction accuracy and classification accuracy on the reconstructed features. We then train the masking module by sweeping across learning rates and selecting the configuration that is best for retrieval on the validation split of CC3M~\cite{sharma2018conceptual}, to ensure generalization to different datasets. Full details are provided in the supplement in \cref{supp:sec:cc12m:implementation}. As in \cref{sec:method}, we compare against a no-SAE baseline with a mask trained on CLIP image embeddings.

\myparagraph{Mask Analysis.} We first evaluate if the learnt mask is meaningful and selective. As a sanity check, we evaluate the distribution of mask values, and find that it is non-uniform and high only for a few latents, as expected (\cref{supp:table:mask_sparsity}). We next evaluate whether (i) latents selected by the mask, given by mask values, correlate to (ii) the presence of the concepts they encode in the conditioning caption. In other words, if a latent encodes a particular concept, we want to evaluate if captions containing that concept correspond to high mask values for that latent. As a proxy for the concept encoded by a latent, we follow \citet{rao2024discover} and compute the cosine similarity between the text embedding of a caption and the corresponding SAE decoder dictionary vector, where a higher value indicates that the latent encodes the concepts present in the caption. We then compute the Pearson correlation $\rho$ between this similarity and the mask value across captions, with positive correlation indicating that the mask is meaningful. We find that 20\% of active latents on DOCCI show $\rho > 0.1$, as compared to 0\% for a caption-shuffled null baseline. While this number may appear small, not all latents are likely to be equally important. To evaluate this, we zero out mask values for the top, random, and bottom $m$ latents, and evaluate the impact on downstream retrieval. We sort the latents in two ways, by $\rho$ and by the mask values themselves. If the mask keeps latents important for alignment, then one would expect removing the top latents to lead to a strong drop in retrieval, and the opposite when removing the bottom latents. We find (\cref{tab:mask_real_data}) that the results consistently match expectations for both rankings: ablating the top latents leads to strong drops in retrieval, followed by random, and followed by the bottom latents. The latents that are well correlated are also the most impactful, \eg ablating the top 32 by $\rho$ leaves 5.2 R@1, as compared to 21.0 with the bottom 32. Results for text-to-image retrieval are provided in \cref{supp:table:ablation_full}.

\myparagraph{Selectivity of Conditioning.} Next, we evaluate whether conditioning is selective of the concepts present in the caption. For each DOCCI image whose caption names two objects, we condition the image on each of them in turn, and evaluate whether each edited embedding is closer to its own object. Over 1{,}061 images and 1{,}528 object pairs, we find that \ours selects the named object 79.4\% of the time, as compared to 55.6\% for the no-SAE baseline, where chance is 50\%.

\myparagraph{Vision-Language Alignment.} To evaluate the impact of \ours on cross-modal alignment, we plot the cosine similarities between image and corresponding text embeddings, both before and after editing. We find (\cref{fig:alignment_plot_cc12m}) that alignment improves across datasets after applying \ours (\eg 0.417$\to$0.470 for MSCOCO). Further results on different models are provided in \cref{supp:sec:cc12m:crossmodal}. Note however that alignment is a diagnostic metric, with the primary goal being to improve downstream performance, which we discuss next.

\myparagraph{Retrieval Performance.} We evaluate the performance of \ours for text-to-image and image-to-text retrieval across fine-grained long-caption (DOCCI~\cite{onoe2024docci}, IIW~\cite{garg2024imageinwords}) and coarse-grained short-caption (MS COCO~\cite{lin2014microsoft}, Flickr30k~\cite{plummer2015flickr30k}) retrieval benchmarks.  Similar to other cross-modal conditioning methods (\eg~\citealp{xie2025smartclip,xiao2024flair}), \ours incurs an additional cost for retrieval, since every image is conditioned on every text. To mitigate this, we only use \ours to rerank the top-10 candidates obtained via the base model. We find (\cref{tab:retrieval:docci_iiw,tab:retrieval:coco_flickr}) that \ours consistently improves retrieval performance across nearly all datasets and models. Qualitative examples are provided in \cref{fig:teaser,supp:sec:cc12m:qualitative}. Interestingly, the no-SAE baseline, despite performing worse for latent selectivity and steering, performs comparably well. We believe this could be because of the added challenge of disentangling attributes and the noisiness of captions in real data, and we leave further investigation of this for future work. To further understand this, we evaluate on a more controlled setup, where we fix the conditioning text to a single object and use the exhaustive MS COCO instance annotations as ground truth. We find (\cref{fig:coco_qualitative}) that while both \ours and no-SAE promote the conditioned object, only \ours removes the object it was not conditioned on, which suggests that \ours is more selective. 

\myparagraph{Applicability to Existing Methods.}
We additionally compare against SharedCLIP and AlignCLIP~\cite{eslami2024mitigate}, which recently proposed using intra-modal separation objectives during CLIP training to improve cross-modal alignment and retrieval performance. Even though these training objectives already provide gains, we find (\cref{tab:retrieval:docci_iiw,tab:retrieval:coco_flickr}) that our approach can additionally help these methods with consistent and significant improvement on fine-grained datasets and competitive performance on coarse-grained datasets.

\input{tables/robust_retrieval_results}

\myparagraph{Comparison to SmartCLIP.} We compare \ours against SmartCLIP~\cite{xie2025smartclip}, which learns a text-conditioned mask on raw vision embeddings. For a fair comparison, we train on a similar setup with CC12M. We find (\cref{tab:retrieval:smartclip_main}, and \cref{supp:sec:cc12m:smartclip}) that \ours consistently outperforms it on fine-grained datasets and is competitive on coarse-grained datasets, highlighting the benefit of disentangling representations. Additionally, similar to \citet{xie2025smartclip}, we obtain comparable zeroshot performance as the baseline; for details refer to \cref{supp:sec:cc12m:zeroshot}.

\myparagraph{Robust Retrieval.} We additionally evaluate \ours for robust retrieval, using the RoCOCO~\cite{park2024rococo} benchmark. This augments the caption set of MS COCO~\cite{lin2014microsoft} with perturbed captions containing irrelevant concepts that alter their meaning, and should not be retrieved by the model. Specifically, it contains four sets---`Rand-voca', `Danger', `Same-concept',  and `Diff-concept'---which replace words in the original caption with alternatives based on the set.
Following \cite{park2024rococo}, we report (\cref{tab:retrieval:robustness_full}) the drop rate (percentage drop in retrieval after augmenting captions), and Recall Score of Manipulated Samples (RSMS) (fraction of data where a perturbed entity was retrieved on top). We find that \ours provides consistent improvements across these metrics on all data splits. Additional results on other backbones are provided in \cref{supp:sec:cc12m:rococo_additional}.

\input{tables/smartclip_main}

%% file: tables/mask_real_data.tex
\begin{table}[!t]
\footnotesize
\setlength{\tabcolsep}{4pt}
    \centering
    \caption{\textbf{Ablating mask latents by ranking.}
    We zero the mask for $m$/128 latents and report DOCCI
    I~$\rightarrow$~T R@1, where the baseline performance is 24.20. The latents are
    ranked in two ways: by their mask value, and by $\rho$, the correlation between a
    latent's mask value and the presence of its concept in the conditioning caption.
    We find that zeroing out top latents leads to the sharpest drops in performance for both rankings, which shows that important latents are ranked higher.
    }
    \vspace{0.4em}
    \begin{tabular}{@{}llccc@{}}
        \toprule
        Ranked by & $m$ & top & random & bottom \\
        \midrule
        \multirow{2}{*}{mask value}
          & 32 & \textbf{8.62} & 15.58 & 22.94 \\
          & 64 & \textbf{0.80} & \phantom{0}7.24  & 16.64 \\
        \midrule
        \multirow{3}{*}{grounding $\rho$}
          & \phantom{0}8  & \textbf{18.0} & 22.1 & 23.6 \\
          & 16 & \textbf{12.1} & 20.4 & 22.8 \\
          & 32 & \textbf{5.2}  & 15.2 & 21.0 \\
        \bottomrule
    \end{tabular}
    \label{tab:mask_real_data}
\end{table}

%% file: tables/retrieval_results_fine.tex
\begin{table*}[!t]
\footnotesize
\setlength{\tabcolsep}{3pt}
    \centering
     \caption{\textbf{Fine-grained, long-caption retrieval on DOCCI and IIW.} All models are trained on CC12M dataset. We report R@1 and R@5 for both image-to-text and text-to-image retrieval. For \ours and No-SAE, we use re-ranking, and report mean $\pm$ std over 5 random seeds. We see that \ours improves retrieval performance across models and datasets.}
    \begin{adjustbox}{width=\textwidth}
    \begin{tabular}{C{2.9cm}C{1.5cm}C{1.5cm}C{1.5cm}C{1.5cm}C{1.5cm}C{1.5cm}C{1.5cm}C{1.5cm}}
        \toprule
        \multirow{3}{*}{Model}
        & \multicolumn{4}{c}{DOCCI \cite{onoe2024docci}}
        & \multicolumn{4}{c}{IIW \cite{garg2024imageinwords}}
         \\
        & \multicolumn{2}{c}{I $\rightarrow$ T} & \multicolumn{2}{c}{T $\rightarrow$ I}
        & \multicolumn{2}{c}{I $\rightarrow$ T} & \multicolumn{2}{c}{T $\rightarrow$ I} \\
        & R@1 & R@5 & R@1 & R@5 & R@1 & R@5 & R@1 & R@5 \\

        \midrule
        CLIP ViT-B/16 & $20.38$ & $42.36$ & $7.16$ & $16.96$ & $50.98$ & $77.94$ & $16.88$ & $32.66$ \\
        No SAE & $\mathbf{24.29}$\std{0.22} & $\mathbf{47.24}$\std{0.18} & $\mathbf{8.40}$\std{0.08} & $\mathbf{18.66}$\std{0.04} & $\mathbf{56.86}$\std{0.31} & $\mathbf{81.83}$\std{0.50} & $\mathbf{18.76}$\std{0.15} & $\mathbf{34.84}$\std{0.14} \\
        \rowcolor{hirow}
        \ours & $\mathbf{24.52}$\std{0.15} & $\mathbf{47.30}$\std{0.14} & $\mathbf{8.30}$\std{0.04} & $18.55$\std{0.03} & $55.49$\std{0.36} & $\mathbf{82.42}$\std{0.19} & $\mathbf{18.77}$\std{0.08} & $34.57$\std{0.05} \\
        \hline
        CLIP ViT-L/14 & $23.60$ & $46.68$ & $8.21$ & $18.70$ & $53.92$ & $81.37$ & $18.05$ & $35.59$ \\
        \rowcolor{hirow}
        \ours & $\mathbf{26.84}$\std{0.47} & $\mathbf{50.04}$\std{0.20} & $\mathbf{9.09}$\std{0.06} & $\mathbf{20.44}$\std{0.03} & $\mathbf{60.78}$\std{0.60} & $\mathbf{84.80}$\std{0.35} & $\mathbf{19.72}$\std{0.05} & $\mathbf{37.63}$\std{0.08} \\
        \hline
        SigLIP ViT-B/16 & $20.68$ & $41.84$ & $7.41$ & $17.17$ & $48.20$ & $76.14$ & $16.94$ & $33.08$ \\
        \rowcolor{hirow}
        \ours & $\mathbf{25.29}$\std{0.22} & $\mathbf{47.37}$\std{0.09} & $\mathbf{8.32}$\std{0.06} & $\mathbf{18.59}$\std{0.02} & $\mathbf{57.84}$\std{0.95} & $\mathbf{80.98}$\std{0.19} & $\mathbf{18.90}$\std{0.11} & $\mathbf{35.68}$\std{0.10} \\
        \hline
        SharedCLIP ViT-B/16 & $19.02$ & $41.06$ & $7.16$ & $17.52$ & $51.14$ & $79.08$ & $17.10$ & $33.78$ \\
        \rowcolor{hirow}
        \ours & $\mathbf{22.64}$\std{0.27} & $\mathbf{45.23}$\std{0.07} & $\mathbf{7.94}$\std{0.06} & $\mathbf{18.42}$\std{0.03} & $\mathbf{55.59}$\std{0.31} & $\mathbf{80.98}$\std{0.25} & $\mathbf{17.91}$\std{0.18} & $\mathbf{34.61}$\std{0.10} \\
        \hline
        AlignCLIP ViT-B/16 & $20.32$ & $41.88$ & $7.39$ & $17.43$ & $53.92$ & $82.03$ & $17.46$ & $34.27$ \\
        \rowcolor{hirow}
        \ours & $\mathbf{23.65}$\std{0.26} & $\mathbf{46.02}$\std{0.15} & $\mathbf{8.13}$\std{0.08} & $\mathbf{18.68}$\std{0.05} & $\mathbf{58.92}$\std{0.65} & $\mathbf{84.31}$\std{0.12} & $\mathbf{19.75}$\std{0.20} & $\mathbf{36.06}$\std{0.21} \\
        \bottomrule
    \end{tabular}
    \end{adjustbox}
    \label{tab:retrieval:docci_iiw}
\end{table*}

%% file: figures/tex/coco_qualitative.tex
\begin{figure*}[!t]
    \centering
    \begin{minipage}[c]{0.42\linewidth}
        \centering
        \footnotesize
        \setlength{\tabcolsep}{3pt}
        \begin{tabular}{@{}lcccccc@{}}
            \toprule
            & & \multicolumn{3}{c}{$\Delta$} & \multicolumn{2}{c}{Classes} \\
            \cmidrule(lr){3-5} \cmidrule(lr){6-7}
            & Exp. & Orig. & no-SAE & \ours & no-SAE & \ours \\
            \midrule
            $\Delta_1$ & $\downarrow$ & 0.118 & 0.119 & \textbf{0.095} & 32/62 & \textbf{47/62} \\
            $\Delta_2$ & $\uparrow$   & 0.120 & 0.162 & \textbf{0.195} & 45/62 & \textbf{57/62} \\
            $\Delta_3$ & $\downarrow$ & 0.203 & 0.221 & \textbf{0.188} & 18/62 & \textbf{42/62} \\
            \bottomrule
        \end{tabular}
    \end{minipage}\hfill
    \begin{minipage}[c]{0.56\linewidth}
        \centering
        \includegraphics[width=\linewidth]{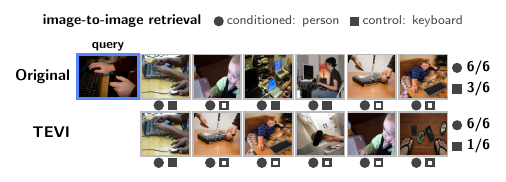}
    \end{minipage}
    \caption{\textbf{Conditioning helps retrieval on MS COCO captions given by attributes.}
    \textit{Left:} Similar to \cref{fig:mad_ordering_combined}, we compute rank-position differences for image~$\rightarrow$~text retrieval, averaged
    over 62 object classes, with equal weight per class and 400 query images. We find that \ours reorders as per the conditioning direction, and outperforms the no-SAE baseline, especially significantly in discarding information not present in the conditioning ($\Delta_1, \Delta_3$). The last
    two columns count how many of the 62 classes move in the expected direction, where \ours consistently outperforms the baseline.
    \textit{Right:} A qualitative example for image~$\rightarrow$~image retrieval. The query image contains a person and a keyboard, and we condition on
    `person'. The retrieved set keeps the person (circle, $6/6 \rightarrow 6/6$) and
    loses the keyboard (square, $3/6 \rightarrow 1/6$).
    }
    \label{fig:coco_qualitative}
\end{figure*}

%% file: tables/robust_retrieval_results.tex
\begin{table*}[!t]
\footnotesize
    \centering
    \caption{\textbf{Robust retrieval performance of \ours on the RoCOCO benchmark~\cite{park2024rococo}}. We find that \ours improves performance over the baseline across settings.
    }
    \begin{adjustbox}{width=\textwidth}
    \begin{tabular}{L{0.7cm}C{0.7cm}C{0.5cm}C{1.15cm}C{0.6cm}C{0.5cm}C{1.15cm}C{0.6cm}C{0.5cm}C{1.15cm}C{0.6cm}C{0.5cm}C{1.15cm}C{0.6cm}}
        \toprule
        \multirow{2}{*}{Model} & \multirow{2}{*}{\shortstack{COCO\\R@1}}
        & \multicolumn{3}{c}{Rand-voca} 
        & \multicolumn{3}{c}{Same-concept}
        & \multicolumn{3}{c}{Diff-concept}
        & \multicolumn{3}{c}{Danger} 
         \\ 
        & & R@1 $(\uparrow)$ & drop rate $(\downarrow)$ & RSMS $(\downarrow)$ & R@1 $(\uparrow)$ & drop rate $(\downarrow)$ & RSMS $(\downarrow)$ & R@1 $(\uparrow)$ & drop rate $(\downarrow)$ & RSMS $(\downarrow)$ & R@1 $(\uparrow)$ & drop rate $(\downarrow)$ & RSMS $(\downarrow)$\\
        \midrule
        CLIP & 33.26 & 21.26 & 12.00 & 46.90 & 21.74 & 11.52 & 43.36 & 22.12 & 11.14 & 43.98 & 22.92 & 10.34 & 39.74 \\
        \rowcolor{hirow}
        +\ours & \textbf{36.00} & \textbf{26.14} & \textbf{9.86} & \textbf{33.98}  & \textbf{25.74}  & \textbf{10.26} & \textbf{37.12} & \textbf{25.60} & \textbf{10.40} & \textbf{35.70} & \textbf{27.12} & \textbf{8.88} & \textbf{31.00}  \\
        \bottomrule
    \end{tabular}
    \end{adjustbox}
    \label{tab:retrieval:robustness_full}
\end{table*}

%% file: tables/smartclip_main.tex
\begin{table}[!t]
\footnotesize
\centering
\caption{\textbf{Retrieval comparison with SmartCLIP.} \ours outperforms on fine-grained datasets and is competitive on coarse-grained datasets. Both methods are applied without re-ranking. Full results in \cref{supp:sec:cc12m:smartclip}.}
\label{tab:retrieval:smartclip_main}
\setlength{\tabcolsep}{3pt}
\begin{tabular}{l*{8}{C{0.55cm}}}
\toprule
Model
& \multicolumn{4}{c}{DOCCI I$\rightarrow$T, T$\rightarrow$I}
& \multicolumn{4}{c}{Flickr30k I$\rightarrow$T, T$\rightarrow$I} \\
& R@1 & R@5 & R@1 & R@5 & R@1 & R@5 & R@1 & R@5 \\
\midrule
CLIP
& 20.4 & 42.4 & 7.2 & 17.0 
& 59.7 & 83.7 & 42.5 & 70.3 \\
\rowcolor{hirow}
+\ours
& \textbf{24.2} & \textbf{48.7} & \textbf{8.6} & \textbf{20.0}
& \textbf{64.2} & \textbf{85.7} & 44.8 & 72.1 \\
SmartCLIP
& 21.8 & 45.0 & 7.5 & 17.8
& 61.0 & 85.1 & \textbf{45.3} & \textbf{72.8} \\
\bottomrule
\end{tabular}
\end{table}

%% file: sec/6_conclusion.tex
\section{Conclusion}
\label{sec:conclusion}

In this work, we explored whether captions can guide what information is preserved in CLIP image embeddings, in order to improve cross-modal alignment. We proposed a framework, \ours, to use text as a signal to edit image embeddings, such that the resultant edited embedding is better aligned with the text.
We first used a synthetic setup, where we could explicitly control the information imbalance, to show that text-conditioned editing preserves information described in the caption. 
We then applied \ours to models trained on natural images and showed its effectiveness for learning meaningful and sparse masks that are selective of the conditioning attributes. Finally, we showed that \ours helps improve retrieval, and complements existing approaches such as AlignCLIP. On the whole, we showed that decomposing embeddings into concepts makes the information imbalance controllable.

\clearpage
\section*{Limitations}
\label{sec:limitations}

While our proposed \ours framework provides a controllable setup that shows strong benefits for a variety of CLIP models trained on CC12M, extending to large-scale models trained on billions of data points remains a challenge, likely owing to the complexity of training sufficiently large and diverse SAEs. Nevertheless, our work intends to provide a clear proof of concept. Scaling to models trained on large-scale data would be a fruitful direction for future research, and this might involve using state-of-the-art SAEs and extracting concepts more granularly, such as at the image patch level. In addition, our work relies on SAEs for disentanglement, and is limited by its accuracy. Gains in some setups, \eg retrieval on benchmark data, remain comparable to learning a mask directly without disentanglement from SAEs, which could be a result of this. As with all cross-modal conditioning-based methods, \ours also incurs an additional computational cost as compared to standard retrieval.  Exploring how to extend \ours to practical setups remains an interesting direction for future research.

\section*{Acknowledgements}
Funded in part by the Deutsche Forschungsgemeinschaft (DFG, German Research Foundation) - GRK 2853/1 ``Neuroexplicit Models of Language, Vision, and Action'' - project number 471607914. We thank Mayank Jobanputra for the helpful discussions and feedback.

%% file: supplement/0_main.tex
\appendix
\renewcommand\thesection{\Alph{section}}
\numberwithin{equation}{section}
\numberwithin{figure}{section}
\numberwithin{table}{section}
\renewcommand{\thefigure}{\thesection\arabic{figure}}
\renewcommand{\thetable}{\thesection\arabic{table}}

\raggedbottom

\setcounter{topnumber}{4}
\setcounter{bottomnumber}{2}
\setcounter{totalnumber}{6}
\setcounter{dbltopnumber}{5}
\renewcommand{\topfraction}{0.95}
\renewcommand{\bottomfraction}{0.85}
\renewcommand{\textfraction}{0.05}
\renewcommand{\floatpagefraction}{0.75}
\renewcommand{\dbltopfraction}{0.95}
\renewcommand{\dblfloatpagefraction}{0.75}
\makeatletter
\setlength{\@fptop}{0pt}
\setlength{\@fpsep}{12pt}
\setlength{\@fpbot}{0pt plus 1fil}
\setlength{\@dblfptop}{0pt}
\setlength{\@dblfpsep}{11pt}
\setlength{\@dblfpbot}{0pt plus 1fil}
\makeatother
\crefname{appendix}{Sec.}{Secs.}

\onecolumn
{\begin{center}
\Large\bf
{TEVI: Text-Conditioned Editing of Visual Representations via Sparse Autoencoders for Improved Vision-Language Alignment}\\[1em]
\large
Appendix
\end{center}
}
\newcommand{\additem}[2]{%
\item[\textbf{(\ref{#1})}]
    \textbf{#2} \dotfill\makebox{\textbf{\pageref{#1}}
    }
}

\newcommand{\myindent}{.5em}
\newcommand{\addsubitem}[2]{%
\vspace{.5em}
    \textbf{(\ref{#1})}
        \hspace{\myindent} #2 \\
}

\newcommand{\adddescription}[1]{\vspace{.1em}
\begin{adjustwidth}{0cm}{0cm}
#1
\end{adjustwidth}
}
\setlist[itemize]{noitemsep,leftmargin=*,topsep=0em}
\setlist[enumerate]{noitemsep,leftmargin=*,topsep=0em}

\noindent In this appendix, we provide implementation details and additional results. \cref{supp:sec:mad_results} covers our controlled synthetic setup and \cref{supp:sec:cc12m_results} provides additional details and results for \ours applied on CLIP models trained on CC12M. In \cref{supp:sec:limitations:broader_impact}, we briefly discuss broader impact, and in \cref{supp:sec:licenses}, we enumerate artifacts used.\\

\begin{adjustwidth}{0.15cm}{0.15cm}
\begin{enumerate}[label={({\arabic*})}, topsep=1em, itemsep=.2em]
    \additem{supp:sec:mad_results}{Controlled Synthetic Setup}\\[0.1em]\hfill
    \addsubitem{supp:sec:mad:implementation}{Implementation Details}
    \addsubitem{supp:sec:mad:disentanglement}{Additional Results for Concept Disentanglement with CG-SAE}
    \addsubitem{supp:sec:mad:conditioning}{Additional Results for Effectiveness of Conditioning}
    \additem{supp:sec:cc12m_results}{Evaluation on CLIP Models Trained with Natural Images}\\[0.1em]\hfill
    \addsubitem{supp:sec:cc12m:implementation}{Implementation Details}
    \addsubitem{supp:sec:cc12m:main_tables_copy}{Full Retrieval Results}
    \addsubitem{supp:sec:cc12m:baseline}{Baseline Comparisons}
    \addsubitem{supp:sec:cc12m:rococo_additional}{Additional Results on RoCOCO}
    \addsubitem{supp:sec:cc12m:zeroshot}{Zeroshot Classification Results}
    \addsubitem{supp:sec:cc12m:without_negative_conditioning_retrieval}{Ablation without Negative Conditioning}
    \addsubitem{supp:sec:cc12m:crossmodal}{Cross-modal Alignment}
    \addsubitem{supp:sec:cc12m:smartclip}{Comparison against SmartCLIP}
    \addsubitem{supp:sec:cc12m:qualitative}{Additional Qualitative Examples}

    \additem{supp:sec:limitations:broader_impact}{Broader Impact}\\

    \additem{supp:sec:licenses}{Licenses of Artifacts Used}
\end{enumerate}
\end{adjustwidth}

\twocolumn

\input{supplement/mad_results}

\input{supplement/cc12m_results}

\input{supplement/limitations_broader_impact}

%% file: supplement/mad_results.tex
\section{Controlled Synthetic Setup}
\label{supp:sec:mad_results}

More implementation details about the method on the MAD dataset~\cite{schrodi2024two} are as follows. This is in addition to the details provided in \cref{sec:background:results}.

\subsection{Implementation Details}
\label{supp:sec:mad:implementation}

\myparagraph{Dataset.} The MAD dataset~\cite{schrodi2024two} is a synthetic dataset that consists of images of digits with colors and morphological transforms. Specifically, it consists of the following attributes: `Digit': $\{$0,1,2,3,4,5,6,7,8,9$\}$, `Thickthinning': $\{$thickening, thinning, no thickthinning$\}$, `Scaling': $\{$large, small$\}$, `Fracture': $\{$fracture, no fracture$\}$, `Swelling': $\{$swelling, no swelling$\}$, and `Color': $\{$gray, red, green, blue, cyan, magenta, yellow$\}$, which gives a total of 6 attribute categories and an aggregate of 26 attribute values. The training dataset consists of 1.44 million images, and the test data consists of 240,000 images. For training CLIP models, captions are generated by using the digit and a random sample of three out of the remaining five attribute categories in the image, placed in a random order with a `-' separator. 

\myparagraph{CLIP models.}  Following~\cite{schrodi2024two}, we train the CLIP model using the AdamW~\cite{loshchilov2017decoupled} optimizer for 200 epochs with a batch size of 128 and a weight decay of 0.1. We sweep over learning rates of $\{10^{-5}, 5\times10^{-4}, 5\times 10^{-5}\}$ and pick the learning rate $5\times10^{-4}$ and the final checkpoint with the lowest loss. We use the cosine annealing for the learning rate. For finetuning the text encoder along with the learnt mask, we sweep over learning rates $\{10^{-2}, 10^{-3}, 10^{-4}, 5\times 10^{-4}, 10^{-5}, 10^{-6}\}$ and pick the learning rate $10^{-6}$. Our code is based on the implementation from OpenCLIP~\cite{ilharco_gabriel_2021_5143773} using PyTorch~\cite{paszke2019pytorch}.

\myparagraph{SAE models.} We use the TopK SAE implementation  of~\cite{gao2024scaling}. This particular type of SAE chooses the top few ($=K$) SAE latents to reconstruct the original CLIP embeddings and uses an auxiliary loss that approximates the reconstruction error using the top few (=auxK) dead latents. For our setup, we use K as 12, auxK as 18, after sweeping over these hyperparameters and we train the SAE for 200 epochs. We choose the SAE configuration with the lowest reconstruction error. We sweep over learning rates $\{10^{-1}, 10^{-2}\}$ and expansion factor $1, 2, 4$ and pick $10^{-2}$, $1$ respectively for both the SAEs used for CG-SAE and \ours. 

\myparagraph{Learnt masks.} We use a 3-layer MLP with a hidden dimension of 256 and ReLU activations between linear layers. This MLP predicts values to mask the SAE latents. We sweep over learning rates $0.1$, $0.01$, and $0.001$, and choose $0.01$ as the optimal learning rate. We train for 25 epochs with 5 warmup epochs using the AdamW optimizer~\cite{loshchilov2017decoupled}.

\subsection{Additional Results for Concept Disentanglement with CG-SAE}
\label{supp:sec:mad:disentanglement}

In this section, we provide full results for concept disentanglement from our CG-SAEs, as discussed in~\cref{sec:background:results}. In \cref{fig:auc_roc:supp_others,fig:auc_roc:supp_digit}, we show AUC ROC plots for other attribute categories, to supplement the results provided for the  `Swelling' category in~\cref{fig:auc_roc}. In \cref{fig:madqualitative_supp1,fig:madqualitative_supp2}, we show top activating images for each latent, expanding on ~\cref{fig:madqualitative}. Both quantitatively and qualitatively, we find that our CG-SAE effectively disentangles concepts and each latent activates highly only on the attribute value assigned to it.

\input{figures/tex/top_activating_images_fdw_mad}

\subsection{Additional Results for Effectiveness of Conditioning}
\label{supp:sec:mad:conditioning}

In this section, we provide additional results on our method's effectiveness of conditioning (\cref{table:different_seeds_fdw_freeform:cgsae,table:different_seeds_fdw_freeform:lmask}). Specifically, for robustness, we report results averaged across five runs, and find that the trends observed in \cref{fig:conditioned_accuracy_latent_ablate_cossim} continue to hold.

Further, in \cref{fig:mad_ordering_supp}, we report the retrieval ordering experiment of \cref{fig:mad_ordering_combined} for all three retrieval directions, \ie image~$\rightarrow$~text, text~$\rightarrow$~image and image~$\rightarrow$~image. We find that the rank-position differences move in the expected direction in each of the three settings.

Finally, in \cref{fig:mad_qualitative_page}, we show qualitative examples for text~$\rightarrow$~image retrieval, which expand on the single example shown in \cref{fig:mad_ordering_combined} (right). For each example, we show the top-8 retrieved images before and after conditioning, where the conditioned attribute is marked with \iconcirc~and the control attribute with \iconsq. In most examples, we find that conditioning may keep the conditioned attribute in the retrieved set while discarding the control attribute.

\input{figures/tex/supplement/auc_roc_fdw_all}
\input{figures/tex/supplement/mad_qualitative}
\input{figures/tex/supplement/mad_qualitative_page}

\FloatBarrier
\twocolumn[%
  \input{tables/different_seeds_conditioned_accuracy}  \vspace{1.5em}%
]

%% file: figures/tex/top_activating_images_fdw_mad.tex
\begin{figure*}[!p]
    \centering
    \begin{minipage}[c][0.90\textheight][c]{\textwidth}
    \centering
    \begin{minipage}[t]{0.94\textwidth}
        \centering
        \includegraphics[trim={0 0 0 1.8cm},clip,width=0.90\linewidth]{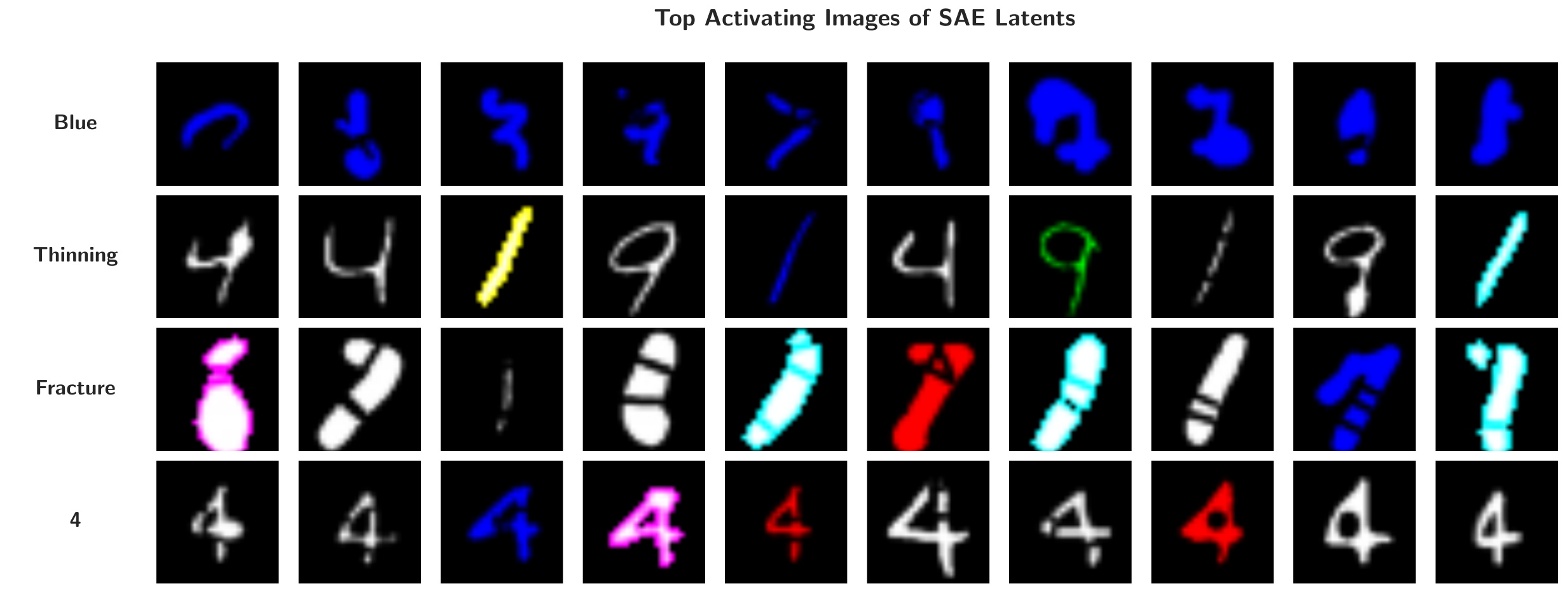}
        \captionof{figure}{\textbf{Qualitative examples of top activating images for the setup when the CG-SAE is trained with fixed semantics of latents.} Each row corresponds to an SAE latent and is labelled with the predefined concept that is assigned to that latent (\cref{sec:background:sanity}). The columns show examples of images that maximally activate these latents. We find that the SAE learns to disentangle the CLIP image features into concepts as specified by the fixed weight of the latent.}
        \label{fig:madqualitative}
    \end{minipage}

    \vspace{0.8em}
    \begin{minipage}[t]{0.48\textwidth}
        \centering
        \includegraphics[trim={0 0 0 1.8cm},clip,width=\linewidth]{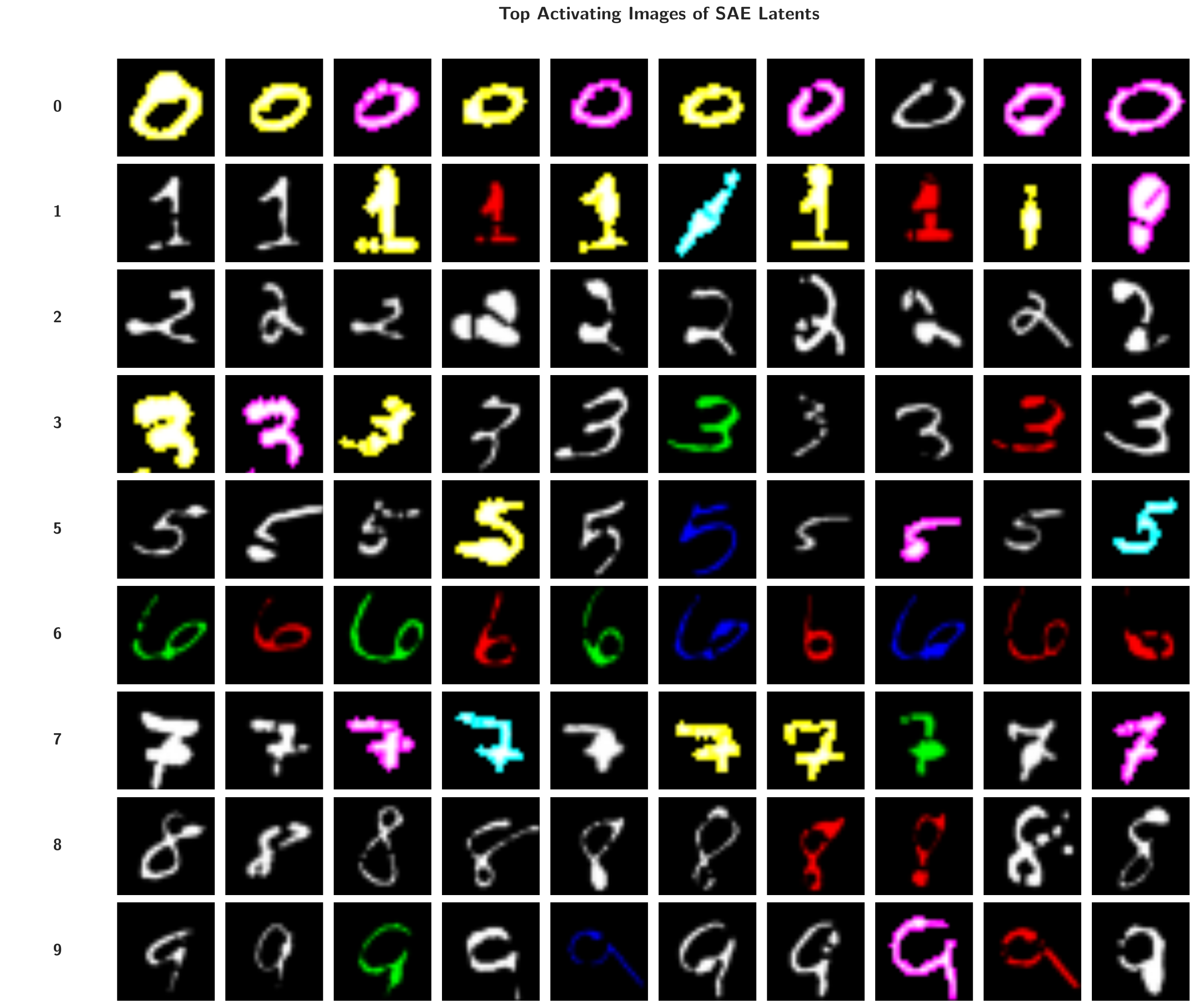}
        \captionof{figure}{\textbf{Qualitative examples of top activating images for all the digits of the CG-SAE latents.} Each row corresponds to a CG-SAE latent and is labelled with the predefined concept that is assigned to that latent (\cref{sec:background:sanity}). The columns show examples of images that maximally activate these latents. We find that the CG-SAE learns to disentangle the CLIP image features into concepts as specified by the fixed weight of the latent.}
        \label{fig:madqualitative_supp1}
    \end{minipage}
    \hfill
    \begin{minipage}[t]{0.48\textwidth}
        \centering
        \includegraphics[trim={0 0 0 1.8cm},clip,width=\linewidth]{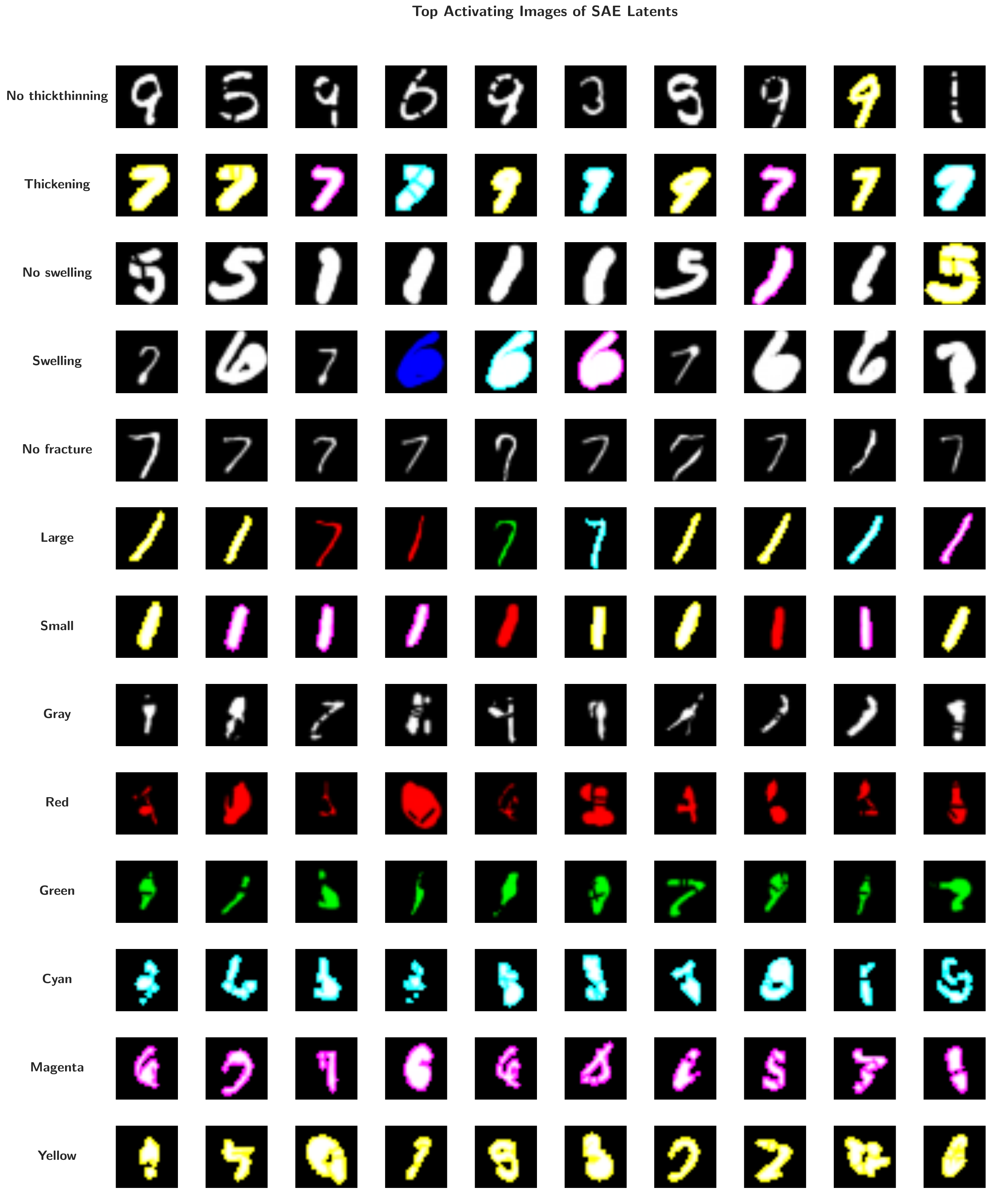}
        \captionof{figure}{\textbf{Qualitative examples of top activating images of the remaining CG-SAE latents.} Each row corresponds to a CG-SAE latent and is labelled with the predefined concept that is assigned to that latent (\cref{sec:background:sanity}). The columns show examples of images that maximally activate these latents. We find that the CG-SAE learns to disentangle the CLIP image features into concepts as specified by the fixed weight of the latent.}
        \label{fig:madqualitative_supp2}
    \end{minipage}
    \end{minipage}
\end{figure*}

%% file: figures/tex/supplement/auc_roc_fdw_all.tex
\begin{figure*}[ht]
    \centering
    \begin{minipage}[t]{0.39\textwidth}
        \centering
        \vspace{0.001cm}
        \includegraphics[width=\linewidth, valign=t]{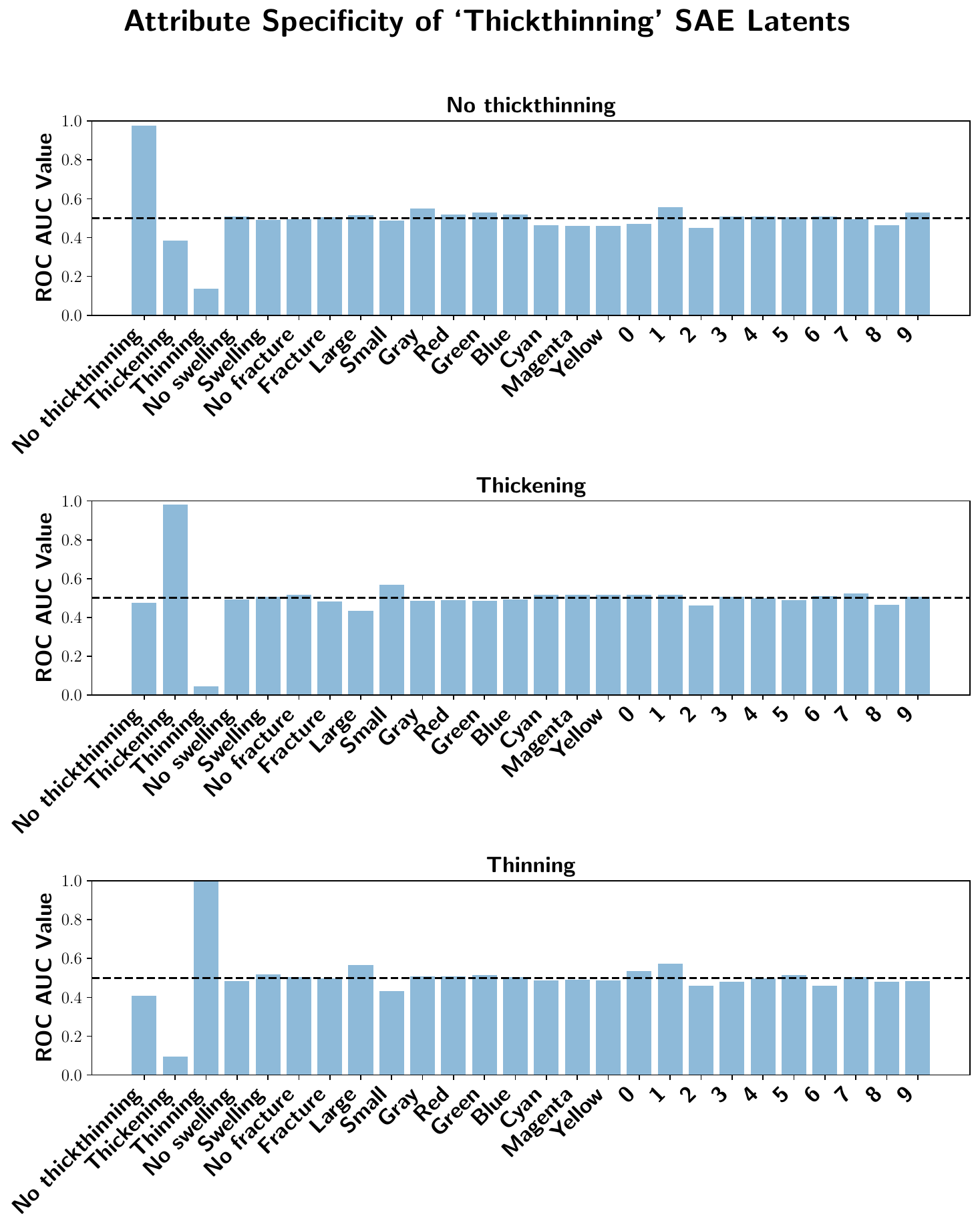}
        
        \includegraphics[width=\linewidth, valign=t]{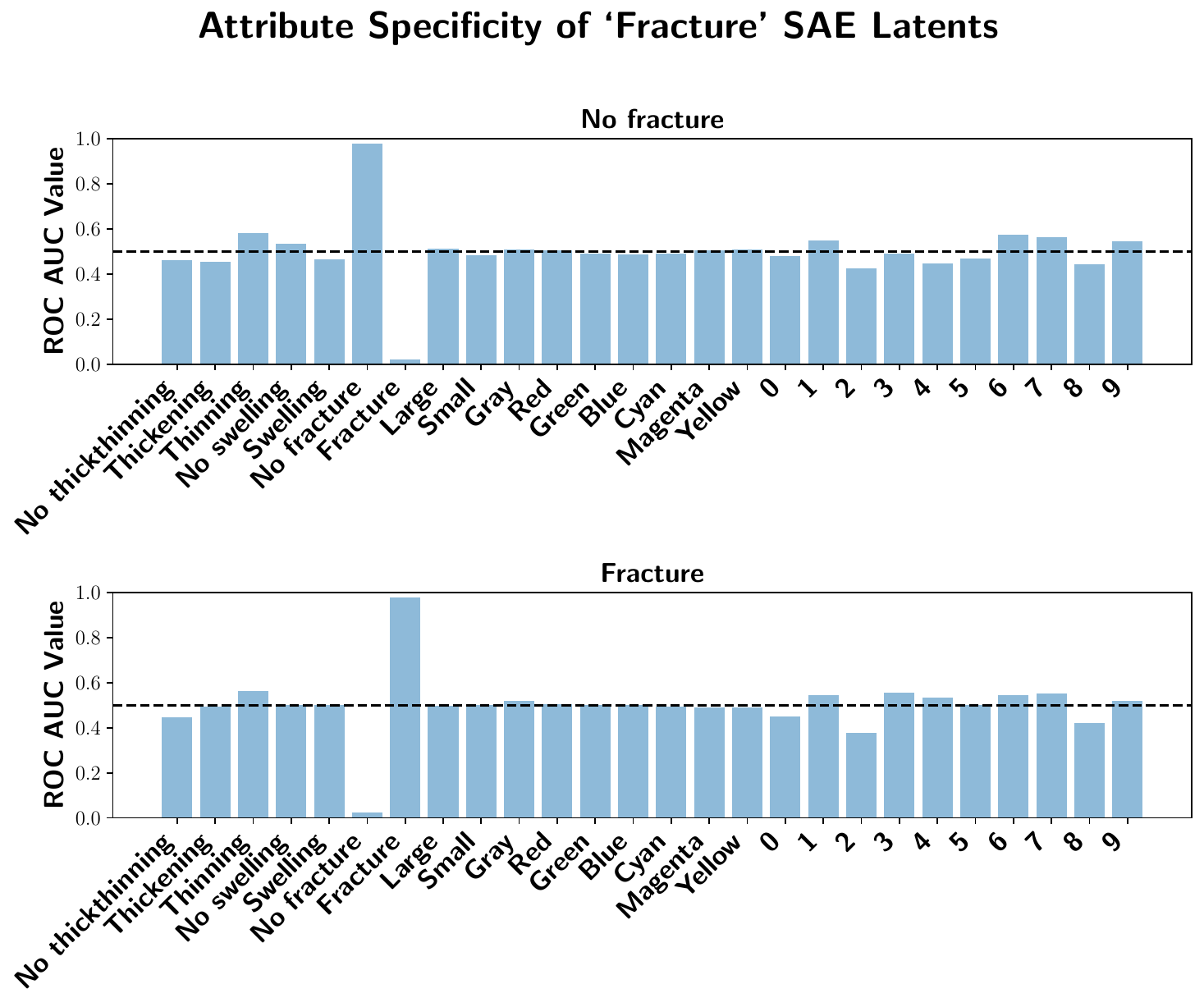}
        
        \includegraphics[width=\linewidth, valign=t]{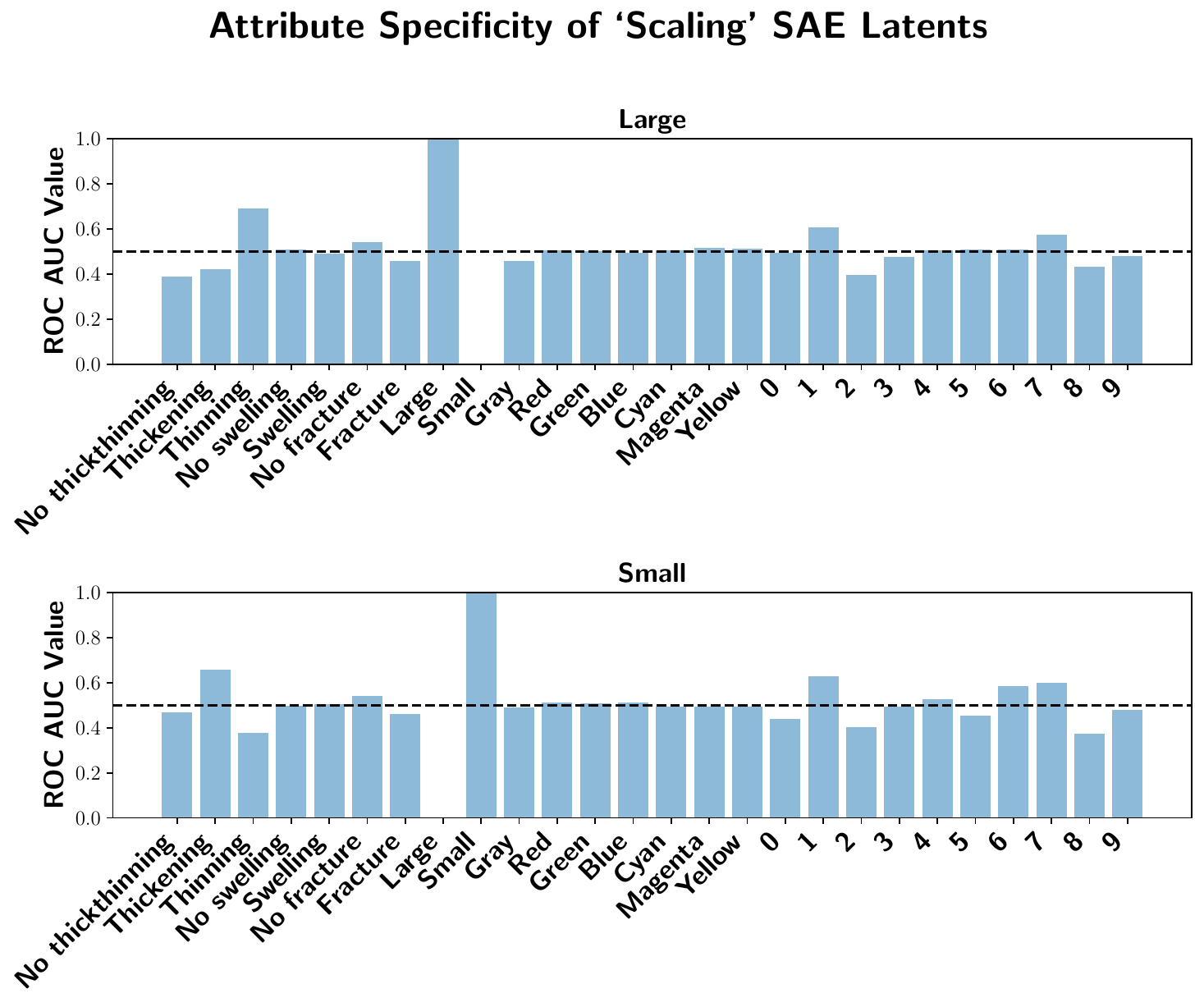}
    \end{minipage}
    \hfill
    \begin{minipage}[t]{0.39\textwidth}
        \centering
        \includegraphics[width=\linewidth, valign=t]{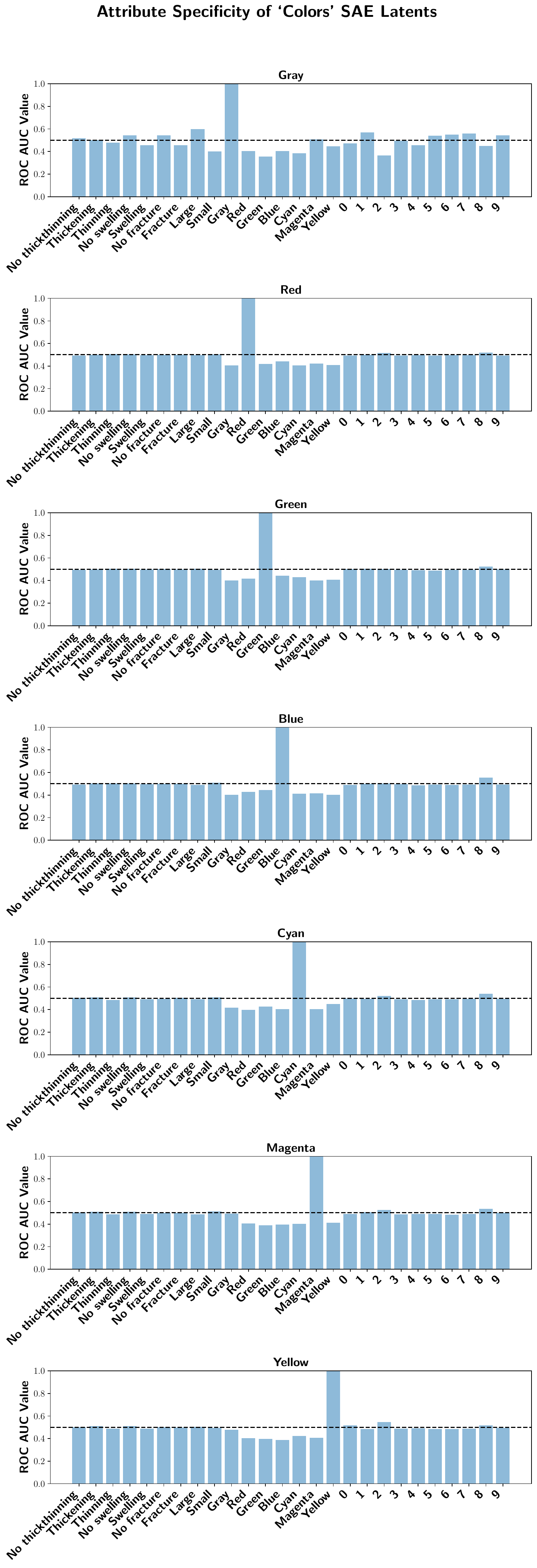}
    \end{minipage}
    
    \caption{\textbf{Attribute specificity of CG-SAE latents for attribute categories `Thickthinning', `Fracture', `Scaling', and `Color'.} We plot the area under the receiver operating characteristic
(ROC) curve (AUC) for CG-SAE latents corresponding to the attribute values of each attribute category.
We find that the latents are highly attribute-specific, with the AUC being close to 1 for the attribute the latent is
assigned to, and 0.5 for unrelated attributes. For results for other attribute categories, see \cref{fig:auc_roc} and \cref{fig:auc_roc:supp_digit}.}
\label{fig:auc_roc:supp_others}
\end{figure*}

\begin{figure*}[ht]
    \centering
    \includegraphics[width=0.6\textwidth]{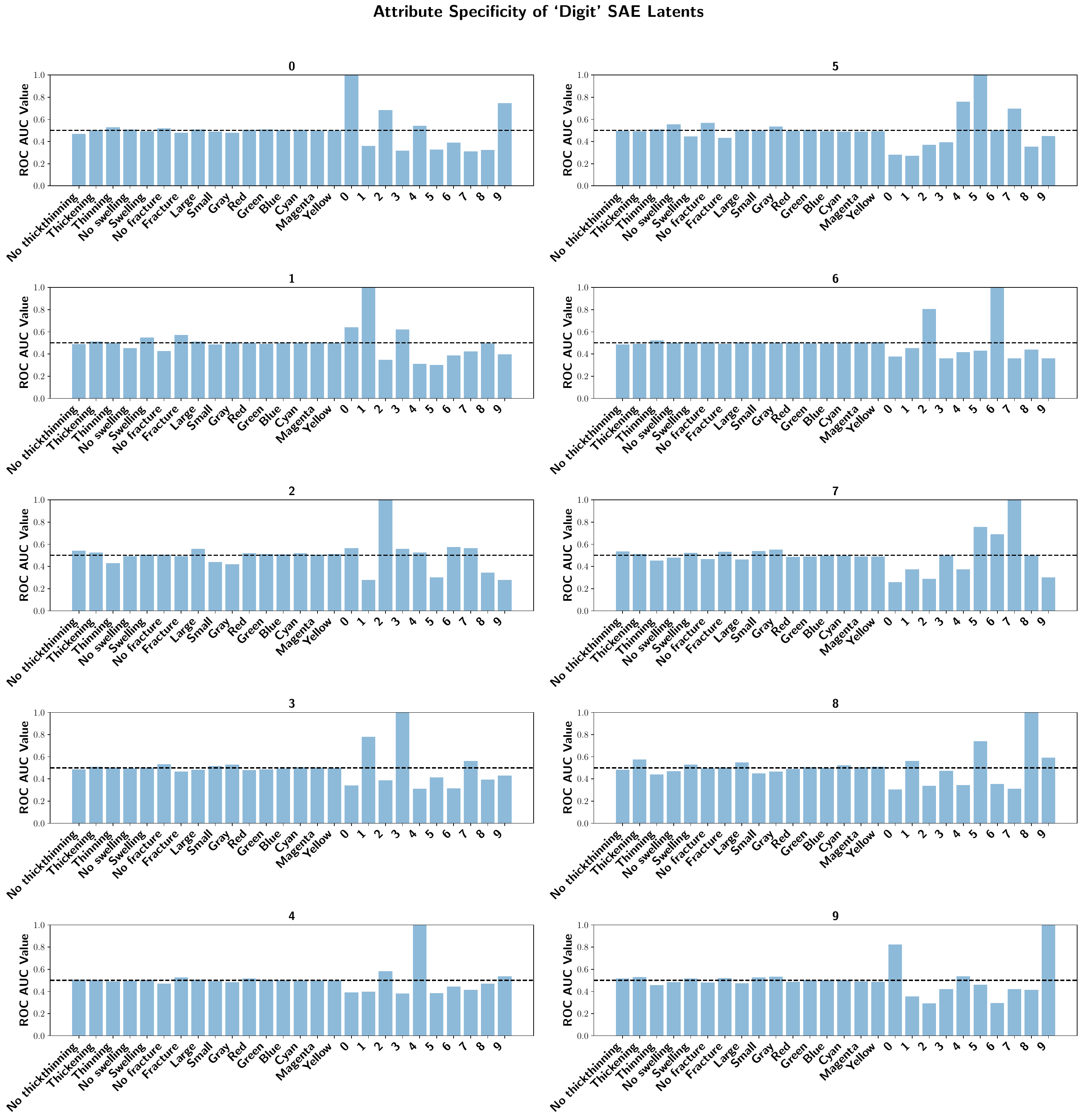}
    \caption{\textbf{Attribute specificity of CG-SAE latents for attribute category `Digit'.} We plot the area under the receiver operating characteristic
    (ROC) curve (AUC) for CG-SAE latents corresponding to each digit attribute.
    We find that the latents are fairly attribute-specific, with the AUC being close to 1 for the attribute the latent is
    assigned to, and close to 0.5 for unrelated attributes. For results for other attribute categories, see \cref{fig:auc_roc} and \cref{fig:auc_roc:supp_others}.}
    \label{fig:auc_roc:supp_digit}
\end{figure*}

%% file: figures/tex/supplement/mad_qualitative.tex
\begin{figure*}[!tb]
    \centering
    \includegraphics[width=0.93\linewidth]{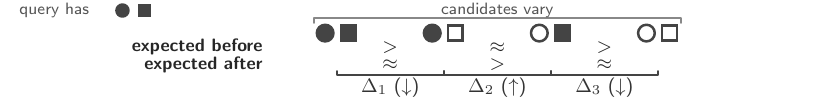}\\[0.5em]
    \includegraphics[width=\linewidth]{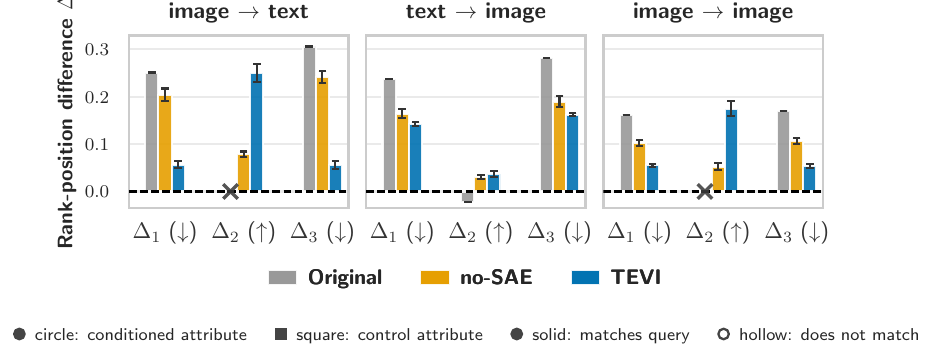}
    \caption{\textbf{Conditioning reorders the retrieved set on MAD.} We condition on one attribute (\iconcirc) and use another as control (\iconsq), see top right. With a query containing both \iconcirc~and \iconsq~and varying candidates, we expect 
    $\Delta_1$ and $\Delta_3$ to fall towards zero and $\Delta_2$ to grow, as the edited embedding should preserve \iconcirc~and discard \iconsq. $\Delta$ denotes the difference in mean rank position between two groups of candidates, where each rank is normalized between 0 and 1. The pattern holds in all three directions, for image~$\rightarrow$~text, text~$\rightarrow$~image and
    image~$\rightarrow$~image.  All
    bars are the mean over five training seeds and the error bars are tshe standard
    deviation over those seeds. See \cref{fig:mad_ordering_combined}.}
    
    \label{fig:mad_ordering_supp}
\end{figure*}

%% file: figures/tex/supplement/mad_qualitative_page.tex
\begin{figure*}[!t]
    \centering
    \includegraphics[width=\linewidth]{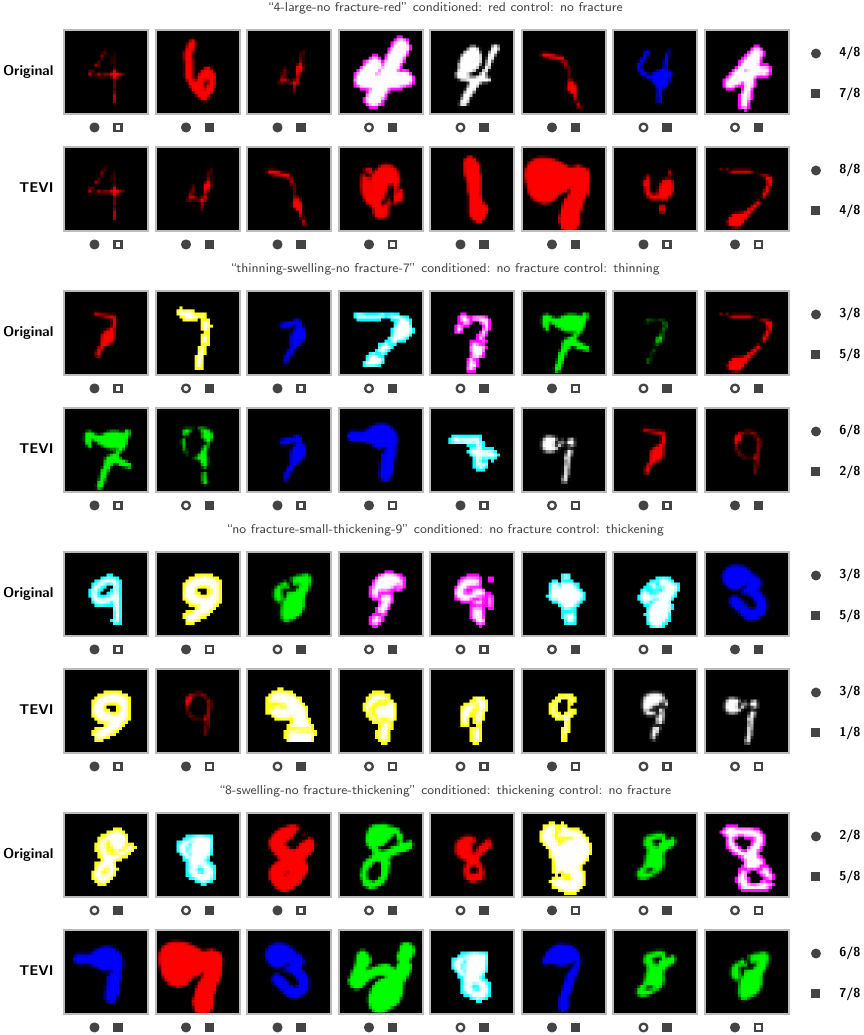}
    
    \caption{\textbf{MAD text~$\rightarrow$~image examples across the range of outcomes.} Each block shows the top-8 retrieved images before and after conditioning, for a different query caption and attribute pair. The circle marks the conditioned attribute and the square the control, filled when the retrieved image matches the query on that attribute. Blocks are a mix, so that weaker cases appear alongside strong ones. \cref{fig:mad_ordering_combined} shows one of these examples at full size.}
    \label{fig:mad_qualitative_page}
\end{figure*}

%% file: tables/different_seeds_conditioned_accuracy.tex
\begin{minipage}{\textwidth}
{\footnotesize
        \centering
        \captionof{table}{\textbf{Effectiveness of conditioning with CG-SAE for the MAD dataset.} For each attribute category (rows), we report the accuracy with the original embeddings (col.~1) and conditioned embeddings, where the attribute of that category is present in the caption (col.~2) and is absent from the caption (col.~3), averaged over five runs. We see that when the attribute is present in the caption, the accuracy is at par with the original embeddings. However, when the attribute is absent from the caption, the accuracy usually reaches close to random chance (col.~4), which shows that conditioning is effective in only preserving information about attributes present in the caption.}
        \vspace{0.4em}
        \begin{tabular}{lcccc}
            \toprule
            & Original ($\uparrow$) & Attribute in Caption ($\uparrow$) & Attribute not in Caption  & Random Chance \\
            \midrule
            Thickthinning & 92.9 & 97.8 $\pm$ 1.3 & 35.4 $\pm$ 2.1 & 33.3 \\
            Swelling      & 96.2 & 96.6 $\pm$ 6.8 & 50.2 $\pm$ 0.1 & 50.0 \\
            Fracture     & 92.8 & 97.5 $\pm$ 3.0 & 50.5 $\pm$ 0.7 & 50.0  \\
            Scaling      & 99.7 & 99.9 $\pm$ 0.1 & 60.7 $\pm$ 3.4 & 50.0 \\
            Color       & 100.0 & 100.0     $\pm$ 0.0   & 34.2 $\pm$ 1.7 & 14.3 \\
            \bottomrule
        \end{tabular}
        \label{table:different_seeds_fdw_freeform:cgsae}
        \label{fig:different_seeds_fdw}

        \vspace{0.8em}
        \centering
        \captionof{table}{\textbf{Effectiveness of conditioning with \ours for the MAD dataset.} For each attribute category (rows), we report the accuracy with the original embeddings (col.~1) and conditioned embeddings, where the attribute of that category is present in the caption (col.~2) and is absent from the caption (col.~3), averaged over five runs. We see that when the attribute is present in the caption, the accuracy is at par with the original embeddings. However, when the attribute is absent from the caption, the accuracy usually reaches close to random chance (col.~4), which shows that conditioning is effective in only preserving information about attributes present in the caption.}
        \vspace{0.4em}
        \begin{tabular}{lcccc}
            \toprule
            & Original ($\uparrow$) & Attribute in Caption ($\uparrow$) & Attribute not in Caption  & Random Chance \\
            \midrule
            Thickthinning & 92.9 & 99.1 $\pm$ 0.7 & 54.3 $\pm$ 2.4 & 33.3 \\
            Swelling      & 96.2 & 98.3 $\pm$ 0.2 & 57.0 $\pm$ 2.7 & 50.0 \\
            Fracture     & 92.8 & 93.5 $\pm$ 0.9 & 59.1 $\pm$ 1.5 & 50.0  \\
            Scaling      & 99.7 & 95.2 $\pm$ 2.7 & 57.1 $\pm$ 3.5 & 50.0 \\
            Color       & 100.0 & 96.3 $\pm$ 1.9 & 22.6 $\pm$ 1.9 & 14.3 \\ 
            \bottomrule
        \end{tabular}
        \label{table:different_seeds_fdw_freeform:lmask}
\par}
\end{minipage}

%% file: supplement/cc12m_results.tex
\section{Evaluation on CLIP Models Trained with Natural Images}
\label{supp:sec:cc12m_results}

\subsection{Implementation Details}
\label{supp:sec:cc12m:implementation}

In this section, we provide implementation details. 

\myparagraph{CLIP models.}  We train CLIP models using the AdamW~\cite{loshchilov2017decoupled} optimizer for 30 epochs with a batch size of 2048 and a weight decay of 0.1. We sweep over learning rates of $\{10^{-3}, 5\times10^{-4}\}$ and pick the learning rate $10^{-3}$ and the final checkpoint with the lowest loss. We use cosine annealing for the learning rate. For fine-tuning the text encoder along with training the masking module, we sweep over learning rates $\{10^{-4}, 10^{-5}, 10^{-6}\}$. Our code is based on the implementation from OpenCLIP~\cite{ilharco_gabriel_2021_5143773, eslami2024mitigate} using PyTorch~\cite{paszke2019pytorch}.

\myparagraph{SAE models.} We use the TopK SAE implementation  of~\cite{gao2024scaling}. This particular type of SAE chooses the top few ($=K$) SAE latents to reconstruct the original CLIP embeddings and uses an auxiliary loss that approximates the reconstruction error using the top few (=auxK) dead latents. For our setup, we vary the sparsity (K) value across values $\{ 64, 128, 256 \}$ and choose $K=128$, which yields the most monosemantic dictionary; we ablate this choice in \cref{fig:sae_topk}. We use auxK as 256, after sweeping over these hyperparameters and we train the SAE for 50 epochs. We sweep over learning rates $\{10^{-3}, 5 \times 10^{-3}, 10^{-4}, 5 \times 10^{-4}\}$, and use an expansion factor of $32$ for ViT-B/16 backbones and $16$ for ViT-L/14 backbones. For the learning rate and expansion factor, we discard configurations with a high reconstruction error or a large number of dead latents, and among the remaining ones we select the configuration with the best monosemanticity score. We calculate the monosemanticity score following \cite{pach2026sparse}. 

\myparagraph{Learnt masks.} We use a 3-layer MLP and ReLU activations between linear layers. This MLP predicts values to mask the SAE latents. We sweep over learning rates $\{10^{-4}, 10^{-5}, 10^{-6}\}$. We train for 3 epochs using the AdamW optimizer~\cite{loshchilov2017decoupled}.

\myparagraph{Baselines.} For the results on SharedCLIP and AlignCLIP in  \cref{tab:retrieval:docci_iiw,tab:retrieval:coco_flickr}, we take the checkpoints given by the authors \cite{eslami2024mitigate} and apply \ours on them. Note that these checkpoints were trained with a batch size of 512, and in contrast we train our own CLIP models with a batch size of 2048. As a result, the performance of AlignCLIP/SharedCLIP versus our CLIP models cannot be directly compared.

\myparagraph{Compute usage.} We used H100 and A100 GPUs for training the CLIP models on CC12M, training the SAEs, and the masking module. The maximum number of GPU hours needed for training the CLIP model is upper bounded by 18 hours, and for training the masking module is upper bounded by 15 hours. 

\myparagraph{Robust retrieval.} Following \cite{park2024rococo}, in \cref{tab:retrieval:robustness_full}, we report the drop rate and the RSMS metric in addition to the retrieval metrics. Drop rate is calculated as $\frac{(R@1-Rp@1)}{R@1}$, where $Rp$ is the retrieval score under the perturbed setting. RSMS calculates the percentage of newly added semantically perturbed captions that are retrieved in the first spot by the model. We note that due to the use of SAE to decompose CLIP image features into concepts, our method is unable to work with altered images as given by the benchmark, where the original image is superimposed with a random patch from a different image. As a result, we only report image-to-text retrieval performance in \cref{tab:retrieval:robustness_full}.
\input{figures/tex/supplement/sae_topk_ablation}

\FloatBarrier
\subsection{Full Retrieval Results}
\label{supp:sec:cc12m:main_tables_copy}

We report the full retrieval results here. \cref{supp:tab:retrieval:docci_iiw:copy} reports fine-grained, long-caption retrieval on DOCCI \cite{onoe2024docci} and IIW \cite{garg2024imageinwords}, and \cref{supp:tab:retrieval:coco_flickr:copy} reports coarse-grained, short-caption retrieval on MSCOCO \cite{lin2014microsoft} and Flickr30k \cite{plummer2015flickr30k}. For the fine-grained retrieval results using re-ranking, refer to \cref{tab:retrieval:docci_iiw} of the main paper. The corresponding coarse-grained results are reported in \cref{tab:retrieval:coco_flickr}.
\input{tables/retrieval_results_coarse}

\begin{table*}[!t]
\footnotesize
\setlength{\tabcolsep}{3pt}
    \centering
     \caption{\textbf{Fine-grained, long-caption retrieval without re-ranking on DOCCI and IIW.} All the models are trained on CC12M dataset. We report R@1 and R@5 for both image-to-text and text-to-image retrieval. For \ours we report mean $\pm$ std over 5 random seeds. We see that \ours improves retrieval performance across models and datasets.}
    \begin{adjustbox}{width=\textwidth}
    \begin{tabular}{C{2.9cm}C{1.5cm}C{1.5cm}C{1.5cm}C{1.5cm}C{1.5cm}C{1.5cm}C{1.5cm}C{1.5cm}}
        \toprule
        \multirow{3}{*}{Model}
        & \multicolumn{4}{c}{DOCCI \cite{onoe2024docci}}
        & \multicolumn{4}{c}{IIW \cite{garg2024imageinwords}}
         \\
        & \multicolumn{2}{c}{I $\rightarrow$ T} & \multicolumn{2}{c}{T $\rightarrow$ I}
        & \multicolumn{2}{c}{I $\rightarrow$ T} & \multicolumn{2}{c}{T $\rightarrow$ I} \\
        & R@1 & R@5 & R@1 & R@5 & R@1 & R@5 & R@1 & R@5 \\

        \midrule
        CLIP ViT-B/16 & $20.38$ & $42.36$ & $7.16$ & $16.96$ & $50.98$ & $77.94$ & $16.88$ & $32.66$ \\
        No SAE & $\mathbf{24.03}$\std{0.24} & $47.95$\std{0.29} & $\mathbf{8.60}$\std{0.07} & $\mathbf{20.18}$\std{0.09} & $\mathbf{56.80}$\std{0.82} & $79.74$\std{0.50} & $\mathbf{19.15}$\std{0.13} & $\mathbf{36.62}$\std{0.38} \\
        \rowcolor{hirow}
        \ours & $\mathbf{24.05}$\std{0.21} & $\mathbf{48.48}$\std{0.14} & $\mathbf{8.47}$\std{0.07} & $19.96$\std{0.06} & $\mathbf{55.75}$\std{0.35} & $\mathbf{81.05}$\std{0.59} & $\mathbf{19.14}$\std{0.14} & $\mathbf{36.36}$\std{0.10} \\
        \hline
        CLIP ViT-L/14 & $23.60$ & $46.68$ & $8.21$ & $18.70$ & $53.92$ & $81.37$ & $18.05$ & $35.59$ \\
        \rowcolor{hirow}
        \ours & $\mathbf{26.63}$\std{0.35} & $\mathbf{51.55}$\std{0.28} & $\mathbf{9.30}$\std{0.07} & $\mathbf{21.69}$\std{0.04} & $\mathbf{58.89}$\std{0.59} & $\mathbf{84.67}$\std{0.78} & $\mathbf{19.69}$\std{0.08} & $\mathbf{37.86}$\std{0.05} \\
        \hline
        SigLIP ViT-B/16 & $20.68$ & $41.84$ & $7.41$ & $17.17$ & $48.20$ & $76.14$ & $16.94$ & $33.08$ \\
        \rowcolor{hirow}
        \ours & $\mathbf{24.59}$\std{0.14} & $\mathbf{48.50}$\std{0.20} & $\mathbf{8.51}$\std{0.06} & $\mathbf{20.09}$\std{0.05} & $\mathbf{57.68}$\std{0.74} & $\mathbf{81.18}$\std{0.45} & $\mathbf{18.96}$\std{0.13} & $\mathbf{36.81}$\std{0.14} \\
        \hline
        SharedCLIP & $19.02$ & $41.06$ & $7.16$ & $17.52$ & $51.14$ & $79.08$ & $17.10$ & $33.78$ \\
        \rowcolor{hirow}
        \ours & $\mathbf{22.28}$\std{0.32} & $\mathbf{45.51}$\std{0.29} & $\mathbf{7.90}$\std{0.04} & $\mathbf{18.65}$\std{0.07} & $\mathbf{57.22}$\std{0.61} & $\mathbf{82.16}$\std{0.61} & $\mathbf{17.97}$\std{0.13} & $\mathbf{34.77}$\std{0.21} \\
        \hline
        AlignCLIP ViT-B/16 & $20.32$ & $41.88$ & $7.39$ & $17.43$ & $53.92$ & $82.03$ & $17.46$ & $34.27$ \\
        \rowcolor{hirow}
        \ours & $\mathbf{22.94}$\std{0.25} & $\mathbf{47.18}$\std{0.45} & $\mathbf{8.28}$\std{0.12} & $\mathbf{19.61}$\std{0.15} & $\mathbf{58.59}$\std{0.55} & $\mathbf{84.48}$\std{0.95} & $\mathbf{19.50}$\std{0.29} & $\mathbf{36.37}$\std{0.32} \\
        \bottomrule
    \end{tabular}
    \end{adjustbox}
    \label{supp:tab:retrieval:docci_iiw:copy}
\end{table*}

\begin{table*}[!t]
\footnotesize
\setlength{\tabcolsep}{3pt}
    \centering
     \caption{\textbf{Coarse-grained, short-caption retrieval without re-ranking on MSCOCO  and Flickr30k.} All the models are trained on CC12M dataset. We report R@1 and R@5 for both image-to-text and text-to-image retrieval. For \ours and No-SAE we report mean $\pm$ std over 5 random seeds. We see that \ours improves retrieval performance across models and datasets.}
    \begin{adjustbox}{width=\textwidth}
    \begin{tabular}{C{2.9cm}C{1.5cm}C{1.5cm}C{1.5cm}C{1.5cm}C{1.5cm}C{1.5cm}C{1.5cm}C{1.5cm}}
        \toprule
        \multirow{3}{*}{Model}
        & \multicolumn{4}{c}{COCO \cite{lin2014microsoft}}
        & \multicolumn{4}{c}{Flickr30k \cite{plummer2015flickr30k}}
         \\
        & \multicolumn{2}{c}{I $\rightarrow$ T} & \multicolumn{2}{c}{T $\rightarrow$ I}
        & \multicolumn{2}{c}{I $\rightarrow$ T} & \multicolumn{2}{c}{T $\rightarrow$ I} \\
        & R@1 & R@5 & R@1 & R@5 & R@1 & R@5 & R@1 & R@5 \\

        \midrule
        CLIP ViT-B/16 & $32.98$ & $59.02$ & $21.38$ & $45.30$ & $59.66$ & $83.73$ & $42.46$ & $70.34$ \\
        No SAE & $\mathbf{35.79}$\std{0.34} & $\mathbf{62.29}$\std{0.25} & $\mathbf{23.50}$\std{0.14} & $\mathbf{47.37}$\std{0.27} & $\mathbf{63.57}$\std{0.59} & $\mathbf{85.33}$\std{0.20} & $\mathbf{44.92}$\std{0.17} & $\mathbf{72.67}$\std{0.25} \\
        \rowcolor{hirow}
        \ours & $\mathbf{35.52}$\std{0.23} & $\mathbf{62.05}$\std{0.18} & $23.27$\std{0.09} & $\mathbf{47.17}$\std{0.09} & $\mathbf{64.54}$\std{0.64} & $\mathbf{85.33}$\std{0.26} & $\mathbf{44.77}$\std{0.11} & $71.98$\std{0.12} \\
        \hline
        CLIP ViT-L/14 & $36.52$ & $62.50$ & $23.85$ & $48.42$ & $63.71$ & $86.49$ & $46.51$ & $72.84$ \\
        \rowcolor{hirow}
        \ours & $\mathbf{38.08}$\std{0.23} & $\mathbf{64.24}$\std{0.10} & $\mathbf{24.85}$\std{0.12} & $\mathbf{49.70}$\std{0.26} & $\mathbf{65.27}$\std{0.45} & $\mathbf{87.75}$\std{0.28} & $\mathbf{47.79}$\std{0.10} & $\mathbf{74.54}$\std{0.14} \\
        \hline
        SigLIP ViT-B/16 & $33.88$ & $60.30$ & $22.06$ & $46.21$ & $62.23$ & $85.40$ & $\mathbf{44.69}$ & $70.61$ \\
        \rowcolor{hirow}
        \ours & $\mathbf{35.93}$\std{0.25} & $\mathbf{61.41}$\std{0.27} & $\mathbf{22.32}$\std{0.06} & $\mathbf{46.66}$\std{0.10} & $\mathbf{63.10}$\std{0.25} & $\mathbf{85.86}$\std{0.28} & $43.75$\std{0.22} & $\mathbf{71.54}$\std{0.11} \\
        \hline
        SharedCLIP & $32.62$ & $58.88$ & $21.54$ & $44.81$ & $\mathbf{60.75}$ & $\mathbf{84.22}$ & $43.31$ & $69.53$ \\
        \rowcolor{hirow}
        \ours & $\mathbf{34.06}$\std{0.15} & $\mathbf{60.10}$\std{0.16} & $\mathbf{21.92}$\std{0.14} & $\mathbf{45.78}$\std{0.11} & $59.71$\std{0.89} & $\mathbf{83.77}$\std{0.51} & $\mathbf{43.98}$\std{0.20} & $\mathbf{70.81}$\std{0.30} \\
        \hline
        AlignCLIP ViT-B/16 & $32.70$ & $58.92$ & $\mathbf{21.79}$ & $\mathbf{44.55}$ & $57.50$ & $82.35$ & $\mathbf{41.91}$ & $\mathbf{70.02}$ \\
        \rowcolor{hirow}
        \ours & $\mathbf{34.34}$\std{0.24} & $\mathbf{60.22}$\std{0.16} & $21.18$\std{0.19} & $44.12$\std{0.32} & $\mathbf{60.53}$\std{0.50} & $\mathbf{83.59}$\std{0.09} & $\mathbf{42.34}$\std{0.50} & $\mathbf{70.18}$\std{0.34} \\
        \bottomrule
    \end{tabular}
    \end{adjustbox}
    \label{supp:tab:retrieval:coco_flickr:copy}
\end{table*}

\subsection{Baseline Comparisons}
\label{supp:sec:cc12m:baseline}
 
\ours involves training a masking module for a small number of extra epochs while also fine-tuning the text encoder. For a fairer comparison accounting for this extra training budget, we report in \cref{supp:tab:retrieval:baseline:coco_flickr,supp:tab:retrieval:baseline:docci_iiw} the performance of CLIP models where the text encoder is fine-tuned for the same number of extra epochs. Similarly we also add the ablation of training the masking module on random SAE. We find that \ours  shows improved retrieval performance across datasets.

\begin{table*}[!ht]
\footnotesize
    \centering
    \caption{\textbf{Coarse-grained retrieval performance on MSCOCO \cite{lin2014microsoft} and Flickr30k \cite{plummer2015flickr30k} as compared to fine-tuning CLIP.} All the models are trained on CC12M dataset. We report R@1, R@5 and R@10 for both image-to-text and text-to-image retrieval. We see that \ours improves retrieval performance across models and datasets in comparison to baseline finetuning and random SAE variation as well.}
    \begin{tabular}{C{2.4cm}C{0.7cm}C{0.7cm}C{0.7cm}C{0.7cm}C{0.7cm}C{0.7cm}C{0.7cm}C{0.7cm}C{0.7cm}C{0.7cm}C{0.7cm}C{0.7cm}}
        \toprule
        \multirow{3}{*}{Model} 
        & \multicolumn{6}{c}{MS COCO \cite{lin2014microsoft}} 
        & \multicolumn{6}{c}{Flickr30k \cite{plummer2015flickr30k}} 
         \\ 
        & \multicolumn{3}{c}{I $\rightarrow$ T} & \multicolumn{3}{c}{T $\rightarrow$ I}
        & \multicolumn{3}{c}{I $\rightarrow$ T} & \multicolumn{3}{c}{T $\rightarrow$ I} \\ 
        & R@1 & R@5 & R@10 & R@1 & R@5 & R@10 & R@1 & R@5 & R@10 & R@1 & R@5 & R@10 \\
        \midrule
        CLIP ViT-B/16 &32.98&59.02&70.04&21.38&45.30&57.03& 59.66&	83.73&	90.14&	42.46&	70.33&	79.31\\ 
        \rowcolor{hirow}
        \ours &\textbf{35.66}&\textbf{61.82}&\textbf{72.94}&\textbf{23.12}&\textbf{47.03}&\textbf{58.67}&\textbf{64.20}&	85.70 &\textbf{91.12}&\textbf{44.75}&\textbf{72.10}&\textbf{81.22}  \\ 
        Random SAE &31.36&56.24&68.04&18.12&40.09&51.87&56.71&81.36&88.36&37.65&65.09&75.80 \\ 
        Fine-tuned &35.60&60.78&71.62&22.50&46.40&58.13&63.21	&\textbf{86.29}&	90.93&44.12	&71.50&80.47 \\ 
        
        \bottomrule
    \end{tabular}
    \label{supp:tab:retrieval:baseline:coco_flickr}
\end{table*}

\begin{table*}[!ht]
\footnotesize
    \centering
     \caption{\textbf{Fine-grained retrieval performance on DOCCI \cite{onoe2024docci} and IIW \cite{garg2024imageinwords} as compared to fine-tuning CLIP.} All the models are trained on CC12M dataset. We report R@1, R@5 and R@10 for both image-to-text and text-to-image retrieval. We see that \ours improves retrieval performance across models and datasets in comparison to baseline finetuning and random SAE variation as well.}
     \begin{adjustbox}{width=\textwidth}
    \begin{tabular}{C{2.4cm}C{0.7cm}C{0.7cm}C{0.7cm}C{0.7cm}C{0.7cm}C{0.7cm}C{0.7cm}C{0.7cm}C{0.7cm}C{0.7cm}C{0.7cm}C{0.7cm}}
        \toprule
        \multirow{3}{*}{Model} 
        & \multicolumn{6}{c}{DOCCI \cite{onoe2024docci}} 
        & \multicolumn{6}{c}{IIW \cite{garg2024imageinwords}} 
         \\ 
        & \multicolumn{3}{c}{I $\rightarrow$ T} & \multicolumn{3}{c}{T $\rightarrow$ I}
        & \multicolumn{3}{c}{I $\rightarrow$ T} & \multicolumn{3}{c}{T $\rightarrow$ I} \\ 
        & R@1 & R@5 & R@10 & R@1 & R@5 & R@10 & R@1 & R@5 & R@10 & R@1 & R@5 & R@10 \\
        \midrule
        CLIP ViT-B/16 &20.38&42.36&53.52&7.16&16.96&22.86 &50.98&77.94&\textbf{88.89}&16.88&32.66&41.46\\ 
        \rowcolor{hirow}
        \ours &\textbf{24.20}&\textbf{48.68}&\textbf{60.06}&\textbf{8.55}&\textbf{19.98}&\textbf{26.80}&\textbf{55.72}&\textbf{81.21}&88.72&\textbf{19.37}&\textbf{36.44}&\textbf{45.47}\\ 
        Random SAE &21.30&43.54&54.52&6.82&16.97&23.33&54.74&77.94&86.11&17.01&34.11&43.02 \\ 
        Fine-tuned & 23.54&46.48&56.92&	7.79&18.57&25.00&53.10&80.56&88.23&18.34&35.99&45.20 \\ 
        
        \bottomrule    
    \end{tabular}
    \end{adjustbox}
    \label{supp:tab:retrieval:baseline:docci_iiw}
\end{table*}

\subsection{Additional Results on RoCOCO}
\label{supp:sec:cc12m:rococo_additional}
In addition to the robust retrieval results provided in \cref{tab:retrieval:robustness_full} on the CLIP ViT-B/16 backbone, in \cref{tab:retrieval:robustness_full:additional} we provide additional results comparing \ours against SigLIP and AlignCLIP. 

\input{tables/robust_retrieval_results_additional}

\subsection{Zeroshot Classification Results}
\label{supp:sec:cc12m:zeroshot}
Similar to SmartCLIP (Tab.~3 in \citealp{xie2025smartclip}), we find (\cref{tab:zeroshot}) that the zeroshot classification performance is comparable to the baseline, possibly since zeroshot does not use long captions with multiple attributes, unlike retrieval. Since our method is post hoc, one can achieve the best of both worlds by using the original CLIP embeddings for zeroshot classification.

\input{tables/zeroshot_results}

\subsection{Ablation without Negative Conditioning}
\label{supp:sec:cc12m:without_negative_conditioning_retrieval}
As noted in \cref{sec:real_data:opt_obj}, we use \cref{eq:learntlossours_real} when applying \ours on CLIP models trained with CC12M, to additionally provide conditioning signal on negative captions. In \cref{supp:tab:retrieval:non_neg:flickr_coco,supp:tab:retrieval:non_neg:docci_iiw}, we empirically validate this by performing an ablation study against models trained without negative conditioning, \ie using \cref{eq:learntlossours}, and find that the performance degrades significantly.

\input{tables/without_negative_conditioning_retrieval}

\subsection{Cross-modal Alignment}
\label{supp:sec:cc12m:crossmodal}

\label{supp:sec:cc12m_results:histograms}

\input{figures/tex/alignment_plot_real_data}
Similar to \cref{fig:alignment_plot_cc12m}, \cref{fig:alignment_plot_alignclip,fig:alignment_plot_sharedclip} show cross-modal alignment for SharedCLIP and AlignCLIP across four downstream datasets before and after applying \ours, and find that it similarly improves alignment.
\input{figures/tex/supplement/alignment_plot_sharedclip}

\subsection{Comparison against SmartCLIP}
\label{supp:sec:cc12m:smartclip}

In \cref{tab:retrieval_smartclip:coco_flickr,tab:retrieval_smartclip:docci_iiw}, we compare our approach against SmartCLIP~\cite{xie2025smartclip}, applied on the same CLIP models trained on CC12M. We see that \ours performs comparably on MS COCO and Flickr30k and outperforms on the fine-grained DOCCI and IIW datasets, possibly due to explicit disentanglement from SAEs.

\begin{table*}[!t]
\footnotesize
    \centering
     \caption{\textbf{Fine-grained retrieval performance on DOCCI \cite{onoe2024docci} and IIW \cite{garg2024imageinwords} as compared to SmartCLIP.} All the models are trained on CC12M dataset. We report R@1, R@5 and R@10 for both image-to-text and text-to-image retrieval.}
     \begin{adjustbox}{width=\textwidth}
    \begin{tabular}{cC{0.75cm}C{0.75cm}C{0.75cm}C{0.75cm}C{0.75cm}C{0.75cm}C{0.75cm}C{0.75cm}C{0.75cm}C{0.75cm}C{0.75cm}C{0.75cm}}
        \toprule
        \multirow{4}{*}{Model} 
        & \multicolumn{6}{c}{DOCCI \cite{onoe2024docci}} 
        & \multicolumn{6}{c}{IIW \cite{garg2024imageinwords}} 
         \\ 
        & \multicolumn{3}{c}{I $\rightarrow$ T} & \multicolumn{3}{c}{T $\rightarrow$ I}
        & \multicolumn{3}{c}{I $\rightarrow$ T} & \multicolumn{3}{c}{T $\rightarrow$ I} \\ 
        & R@1 & R@5 & R@10 & R@1 & R@5 & R@10 & R@1 & R@5 & R@10 & R@1 & R@5 & R@10 \\
        
        \midrule
        CLIP ViT-B/16 &20.38&42.36&53.52&7.16&16.96&22.86 &50.98&77.94& 88.89 &16.88&32.66&41.46\\ 
        SmartCLIP & 21.76 & 45.04  & 55.64 & 7.47  & 17.83 & 23.80 & 54.74 & \textbf{82.35} & \textbf{89.05} & 17.62 & 33.90 & 42.15 \\ 
        \rowcolor{hirow}
        \ours &\textbf{24.20}&\textbf{48.68}&\textbf{60.06}&\textbf{8.55}&\textbf{19.98}&\textbf{26.80}&\textbf{55.72}& 81.21 &88.72&\textbf{19.37}&\textbf{36.44}&\textbf{45.47}\\ 
        
        \bottomrule
        
    \end{tabular}
    \end{adjustbox}
    \label{tab:retrieval_smartclip:docci_iiw}
\end{table*}

\begin{table*}[!t]
\footnotesize
    \centering
    \caption{\textbf{Coarse-grained retrieval performance on MSCOCO \cite{lin2014microsoft} and Flickr30k \cite{plummer2015flickr30k} as compared to SmartCLIP.} All the models are trained on CC12M dataset. We report R@1, R@5 and R@10 for both image-to-text and text-to-image retrieval.}
    \begin{adjustbox}{width=\textwidth}
    \begin{tabular}{cC{0.75cm}C{0.75cm}C{0.75cm}C{0.75cm}C{0.75cm}C{0.75cm}C{0.75cm}C{0.75cm}C{0.75cm}C{0.75cm}C{0.75cm}C{0.75cm}}
        \toprule
        \multirow{3}{*}{Model} 
        & \multicolumn{6}{c}{MS COCO \cite{lin2014microsoft}} 
        & \multicolumn{6}{c}{Flickr30k \cite{plummer2015flickr30k}} 
         \\ 
        & \multicolumn{3}{c}{I $\rightarrow$ T} & \multicolumn{3}{c}{T $\rightarrow$ I}
        & \multicolumn{3}{c}{I $\rightarrow$ T} & \multicolumn{3}{c}{T $\rightarrow$ I} \\ 
        & R@1 & R@5 & R@10 & R@1 & R@5 & R@10 & R@1 & R@5 & R@10 & R@1 & R@5 & R@10 \\
        \midrule
        CLIP ViT-B/16 &32.98&59.02&70.04&21.38&45.30&57.03& 59.66&	83.73&	90.14&	42.46&	70.33&	79.31\\ 
        SmartCLIP  & 35.54 & \textbf{61.98} & 72.90 & \textbf{23.89} & \textbf{48.24} & \textbf{59.93} & 61.05 & 85.11 & \textbf{91.32} & \textbf{45.29} & \textbf{72.78} & \textbf{81.24} \\
        
        \rowcolor{hirow}
        \ours &\textbf{35.66}& 61.82 &\textbf{72.94}& 23.12 & 47.03 & 58.67 &\textbf{64.20} &	\textbf{85.70}& 91.12 & 44.75 & 72.10 & 81.22   \\ 
        
        \bottomrule
    \end{tabular}
    \end{adjustbox}
    \label{tab:retrieval_smartclip:coco_flickr}
\end{table*}

\myparagraph{Additional mask analysis.}
\looseness=-1
We provide the full results for the mask analysis of \cref{sec:real_data:results}.
\Cref{supp:table:mask_sparsity} reports the mask sparsity,
\cref{supp:table:ablation_full} reports the latent ablations in both retrieval
directions, and \cref{supp:table:coco_ordering_i2i} repeats the conditioning test
of \cref{fig:coco_qualitative} for image-to-image retrieval, where the gallery
consists of the other images instead of the captions. We find that \ours still
removes the control object, but neither arm promotes the conditioned object,
which makes this the weaker of the two directions. We find that out of K=128 dimensions, the mask
preserves a small number with high value. Interestingly, fine-grained datasets
(DOCCI, IIW) have more latents preserved as compared to coarse-grained ones
(COCO, Flickr), which could be attributed to them having more attributes per
caption. We also report the coefficient of variation of mask values, computed per
image. A uniform mask would simply scale the embedding, and so the high
coefficient of variation ($\sim$0.4) shows that the mask is selective, actively
changing the representation's direction based on the text.

\input{tables/supplement/mask_analysis_page}

\subsection{Additional Qualitative Examples}
\label{supp:sec:cc12m:qualitative}
We provide additional randomly sampled qualitative examples in
\cref{fig:supp:quali_examples}. For each of the four datasets (DOCCI, IIW,
Flickr30k, MS COCO), we provide examples of instances both where \ours
retrieves correctly and the baseline does not, and vice-versa. Note that
some of the retrieved captions (\eg by the baseline model for the first MS
COCO example) appear correct but do not match the image as per the dataset,
which shows the challenging nature of the benchmarks. We show additional
examples in \cref{fig:coco_qualitative_page}, similar to
\cref{fig:coco_qualitative}.

\input{figures/tex/supplement/coco_qualitative_page}
\input{figures/tex/supplement/quali_examples}

%% file: figures/tex/supplement/sae_topk_ablation.tex
\begin{figure}[!t]
    \centering
    \begin{subfigure}[t]{0.49\linewidth}
        \includegraphics[width=\linewidth]{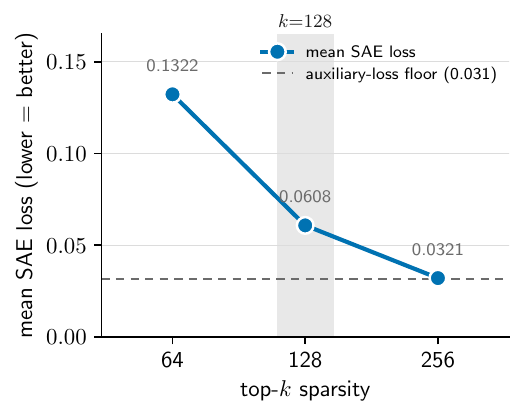}
        \caption{Training objective}
        \label{fig:sae_topk:loss}
    \end{subfigure}\hfill
    \begin{subfigure}[t]{0.49\linewidth}
        \includegraphics[width=\linewidth]{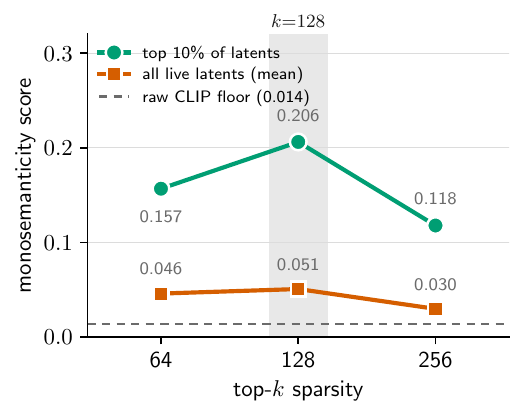}
        \caption{Monosemanticity}
        \label{fig:sae_topk:ms}
    \end{subfigure}
    \caption{\textbf{Choosing the SAE sparsity $K$ for CLIP ViT-B/16 on CC12M.} Left: SAE loss vs K plot. Right: Monosemanticity score vs K. 
    }
    \label{fig:sae_topk}
\end{figure}

%% file: tables/retrieval_results_coarse.tex
        
\begin{table*}[!t]
\footnotesize
\setlength{\tabcolsep}{3pt}
    \centering
     \caption{\textbf{Coarse-grained, short-caption retrieval on MSCOCO and Flickr30k.} All the models are trained on CC12M dataset. We report R@1 and R@5 for both image-to-text and text-to-image retrieval. For \ours and No-SAE, we use re-ranking, and report mean $\pm$ std over 5 random seeds. We see that \ours improves retrieval performance across models and datasets.}
    \begin{adjustbox}{width=\textwidth}
    \begin{tabular}{C{2.9cm}C{1.5cm}C{1.5cm}C{1.5cm}C{1.5cm}C{1.5cm}C{1.5cm}C{1.5cm}C{1.5cm}}
        \toprule
        \multirow{3}{*}{Model}
        & \multicolumn{4}{c}{COCO \cite{lin2014microsoft}}
        & \multicolumn{4}{c}{Flickr30k \cite{plummer2015flickr30k}}
         \\
        & \multicolumn{2}{c}{I $\rightarrow$ T} & \multicolumn{2}{c}{T $\rightarrow$ I}
        & \multicolumn{2}{c}{I $\rightarrow$ T} & \multicolumn{2}{c}{T $\rightarrow$ I} \\
        & R@1 & R@5 & R@1 & R@5 & R@1 & R@5 & R@1 & R@5 \\

        \midrule
        CLIP ViT-B/16 & $32.98$ & $59.02$ & $21.38$ & $45.30$ & $59.66$ & $83.73$ & $42.46$ & $70.34$ \\
        No SAE & $\mathbf{36.26}$\std{0.37} & $\mathbf{62.09}$\std{0.18} & $\mathbf{23.59}$\std{0.12} & $\mathbf{47.31}$\std{0.09} & $63.06$\std{0.40} & $\mathbf{85.25}$\std{0.27} & $\mathbf{45.00}$\std{0.14} & $\mathbf{72.02}$\std{0.15} \\
        \rowcolor{hirow}
        \ours & $\mathbf{36.25}$\std{0.23} & $\mathbf{61.98}$\std{0.18} & $\mathbf{23.48}$\std{0.04} & $\mathbf{47.43}$\std{0.05} & $\mathbf{64.22}$\std{0.37} & $\mathbf{85.17}$\std{0.11} & $\mathbf{44.79}$\std{0.13} & $\mathbf{71.74}$\std{0.17} \\
        \hline
        CLIP ViT-L/14 & $36.52$ & $62.50$ & $23.85$ & $48.42$ & $63.71$ & $86.49$ & $46.51$ & $72.84$ \\
        \rowcolor{hirow}
        \ours & $\mathbf{38.41}$\std{0.24} & $\mathbf{64.27}$\std{0.16} & $\mathbf{24.98}$\std{0.10} & $\mathbf{49.75}$\std{0.15} & $\mathbf{65.92}$\std{0.43} & $\mathbf{87.46}$\std{0.34} & $\mathbf{47.92}$\std{0.10} & $\mathbf{74.54}$\std{0.08} \\
        \hline
        SigLIP ViT-B/16 & $33.88$ & $60.30$ & $22.06$ & $46.21$ & $62.23$ & $85.40$ & $\mathbf{44.69}$ & $70.61$ \\
        \rowcolor{hirow}
        \ours & $\mathbf{37.22}$\std{0.34} & $\mathbf{62.48}$\std{0.14} & $\mathbf{22.53}$\std{0.06} & $\mathbf{47.67}$\std{0.02} & $\mathbf{64.48}$\std{0.40} & $\mathbf{87.08}$\std{0.21} & $43.96$\std{0.23} & $\mathbf{71.75}$\std{0.08} \\
        \hline
        SharedCLIP ViT-B/16 & $32.62$ & $58.88$ & $21.54$ & $44.81$ & $\mathbf{60.75}$ & $84.22$ & $43.31$ & $69.53$ \\
        \rowcolor{hirow}
        \ours & $\mathbf{34.80}$\std{0.13} & $\mathbf{60.79}$\std{0.14} & $\mathbf{22.15}$\std{0.11} & $\mathbf{46.56}$\std{0.13} & $\mathbf{60.87}$\std{0.61} & $\mathbf{85.50}$\std{0.42} & $\mathbf{44.17}$\std{0.16} & $\mathbf{71.22}$\std{0.25} \\
        \hline
        AlignCLIP ViT-B/16 & $32.70$ & $58.92$ & $\mathbf{21.79}$ & $44.55$ & $57.50$ & $82.35$ & $41.91$ & $70.02$ \\
        \rowcolor{hirow}
        \ours & $\mathbf{35.53}$\std{0.13} & $\mathbf{61.56}$\std{0.20} & $21.63$\std{0.16} & $\mathbf{45.70}$\std{0.13} & $\mathbf{61.56}$\std{0.55} & $\mathbf{83.85}$\std{0.32} & $\mathbf{42.85}$\std{0.44} & $\mathbf{71.21}$\std{0.32} \\
        \bottomrule
    \end{tabular}
    \end{adjustbox}
    \label{tab:retrieval:coco_flickr}
\end{table*}

%% file: tables/robust_retrieval_results_additional.tex
\begin{table*}[!ht]
\footnotesize
    \centering
    \caption{\textbf{Robust retrieval performance of \ours on the RoCOCO benchmark~\cite{park2024rococo}}.
    }
    \begin{adjustbox}{width=\textwidth}
    \begin{tabular}{L{1.1cm}C{0.7cm}C{0.5cm}C{1.15cm}C{0.6cm}C{0.5cm}C{1.15cm}C{0.6cm}C{0.5cm}C{1.15cm}C{0.6cm}C{0.5cm}C{1.15cm}C{0.6cm}}
        \toprule
        \multirow{2}{*}{Model} & \multirow{2}{*}{\shortstack{COCO\\R@1}}
        & \multicolumn{3}{c}{Rand-voca} 
        & \multicolumn{3}{c}{Same-concept}
        & \multicolumn{3}{c}{Diff-concept}
        & \multicolumn{3}{c}{Danger} 
         \\ 
        & & R@1 $(\uparrow)$ & drop rate $(\downarrow)$ & RSMS $(\downarrow)$ & R@1 $(\uparrow)$ & drop rate $(\downarrow)$ & RSMS $(\downarrow)$ & R@1 $(\uparrow)$ & drop rate $(\downarrow)$ & RSMS $(\downarrow)$ & R@1 $(\uparrow)$ & drop rate $(\downarrow)$ & RSMS $(\downarrow)$\\
        \midrule
        SigLIP & 33.66 & 21.98 & 11.68 & 44.16 & 22.38 & 11.28 & 42.12 & 22.66 & 11.00 & 41.84 & 24.98 & 8.68 & 31.96 \\
        \rowcolor{hirow}
        \ours & \textbf{36.46} & \textbf{26.32} & \textbf{10.14} & \textbf{33.58} & \textbf{26.14} & \textbf{10.32} & \textbf{35.90} & \textbf{27.06} & \textbf{9.40} & \textbf{32.74} & \textbf{29.14} & \textbf{7.32} & \textbf{24.28} \\
        \hline
        AlignCLIP & 32.86 & 21.46 & \textbf{11.40} & \textbf{43.66} & \textbf{22.72} & \textbf{10.14} & \textbf{40.96} & 21.94 & \textbf{10.92} & 41.20 & 23.44 & 9.42 & 37.52 \\
        \rowcolor{hirow}
        \ours & \textbf{34.70} & \textbf{22.32} & 12.38 & 43.70 & 22.58 & 12.12 & 44.16 & \textbf{22.74} & 11.96 & \textbf{44.48} & \textbf{25.48} & \textbf{9.22} & \textbf{35.20} \\
        \bottomrule
    \end{tabular}
    \end{adjustbox}
    \label{tab:retrieval:robustness_full:additional}
\end{table*}

%% file: tables/zeroshot_results.tex
 \begin{table*}[!ht]
 \footnotesize
    \centering
    \caption{\textbf{Zeroshot performance across benchmarks.} We evaluate on ImageNet-V2~\cite{recht2019imagenet}, ImageNet-Sketch~\cite{wang2019learningrobust}, ImageNet-A~\cite{hendrycks2019natural}, ImageNet-O~\cite{hendrycks2019natural}, ImageNet-R~\cite{hendrycks2020many}, and CIFAR-100~\cite{krizhevsky2009learning}.}
    \begin{tabular}{@{}l ccccc c@{}}
      \toprule
      Model & IMN V2 & Sketch & -A & -O & -R & CIF100 \\
      \midrule
      CLIP ViT-B/16
      & \textbf{29.24} & \textbf{21.93} & \textbf{7.64} & \textbf{26.20} & 42.69 & 28.07 \\
      \rowcolor{hirow}
      +\ours
      & 28.60 & \textbf{21.93} & 7.42 & 25.90 & \textbf{42.76} & \textbf{30.66} \\
      \bottomrule
    \end{tabular}
    \label{tab:zeroshot}
  \end{table*}%

%% file: tables/without_negative_conditioning_retrieval.tex
\begin{table*}[!ht]
\footnotesize
    \centering
     \caption{\textbf{Coarse-grained retrieval performance of CLIP ViT-B/16 when we do not use the negative conditioning on MSCOCO \cite{lin2014microsoft} and Flickr30k \cite{plummer2015flickr30k}}. All the models are trained on CC12M dataset. We report R@1, R@5 and R@10 for image-to-text and text-to-image retrieval.}
     \begin{adjustbox}{width=\textwidth}
    \begin{tabular}{cC{0.75cm}C{0.75cm}C{0.75cm}C{0.75cm}C{0.75cm}C{0.75cm}C{0.75cm}C{0.75cm}C{0.75cm}C{0.75cm}C{0.75cm}C{0.75cm}}
        \toprule
        \multirow{3}{*}{Model} 
        & \multicolumn{6}{c}{MS COCO \cite{lin2014microsoft}} 
        & \multicolumn{6}{c}{Flickr30k \cite{plummer2015flickr30k}} 
         \\ 
        & \multicolumn{3}{c}{I $\rightarrow$ T} & \multicolumn{3}{c}{T $\rightarrow$ I}
        & \multicolumn{3}{c}{I $\rightarrow$ T} & \multicolumn{3}{c}{T $\rightarrow$ I} \\ 
        & R@1 & R@5 & R@10 & R@1 & R@5 & R@10 & R@1 & R@5 & R@10 & R@1 & R@5 & R@10 \\
        \midrule
        CLIP ViT-B/16 &\textbf{32.98}&\textbf{59.02}&\textbf{70.04}&\textbf{21.38}&\textbf{45.30}&\textbf{57.03}& \textbf{59.66}&	\textbf{83.73}&	\textbf{90.14}&	\textbf{42.46}&	\textbf{70.33}&	\textbf{79.31}  \\ 
        \rowcolor{hirow}
        No neg.  & 10.28 & 26.24 & 36.48 & 9.65 & 23.77 & 32.84 & 26.04 & 53.65 & 63.81 & 22.47 & 46.07 & 56.82\\ 
        \bottomrule
    \end{tabular}
    \end{adjustbox}
    \label{supp:tab:retrieval:non_neg:flickr_coco}
\end{table*}

\begin{table*}[!ht]
\footnotesize
    \centering
     \caption{\textbf{Fine-grained retrieval performance of CLIP ViT-B/16 when we do not use the negative conditioning on DOCCI \cite{onoe2024docci} and IIW \cite{garg2024imageinwords}.} All the models are trained on CC12M dataset. We report R@1, R@5 and R@10 for image-to-text and text-to-image retrieval.}
    \begin{adjustbox}{width=\textwidth}
    \begin{tabular}{cC{0.75cm}C{0.75cm}C{0.75cm}C{0.75cm}C{0.75cm}C{0.75cm}C{0.75cm}C{0.75cm}C{0.75cm}C{0.75cm}C{0.75cm}C{0.75cm}}
        \toprule
        \multirow{3}{*}{Model} 
        & \multicolumn{6}{c}{DOCCI \cite{onoe2024docci}} 
        & \multicolumn{6}{c}{IIW \cite{garg2024imageinwords}} 
         \\ 
        & \multicolumn{3}{c}{I $\rightarrow$ T} & \multicolumn{3}{c}{T $\rightarrow$ I}
        & \multicolumn{3}{c}{I $\rightarrow$ T} & \multicolumn{3}{c}{T $\rightarrow$ I} \\ 
        & R@1 & R@5 & R@10 & R@1 & R@5 & R@10 & R@1 & R@5 & R@10 & R@1 & R@5 & R@10 \\
        \midrule
        CLIP ViT-B/16 &\textbf{20.38}&\textbf{42.36}&\textbf{53.52}&\textbf{7.16}&\textbf{16.96}&\textbf{22.86} &\textbf{50.98}&\textbf{77.94}&\textbf{88.89}&\textbf{16.88}&\textbf{32.66}&\textbf{41.46}\\ 
        \rowcolor{hirow}
        No neg.  & 5.54 & 14.50 & 20.88 & 3.07 & 8.29 & 11.90 & 17.65 & 38.56 & 52.12 & 7.80 & 19.44 & 26.82 \\ 
        \bottomrule
    \end{tabular}
    \end{adjustbox}
    \label{supp:tab:retrieval:non_neg:docci_iiw}
\end{table*}

%% file: figures/tex/alignment_plot_real_data.tex
\begin{figure*}[!t]
    \centering
    \begin{subfigure}[t]{0.24\linewidth}
        \includegraphics[trim={0 0 0 0cm},clip,width=\linewidth]{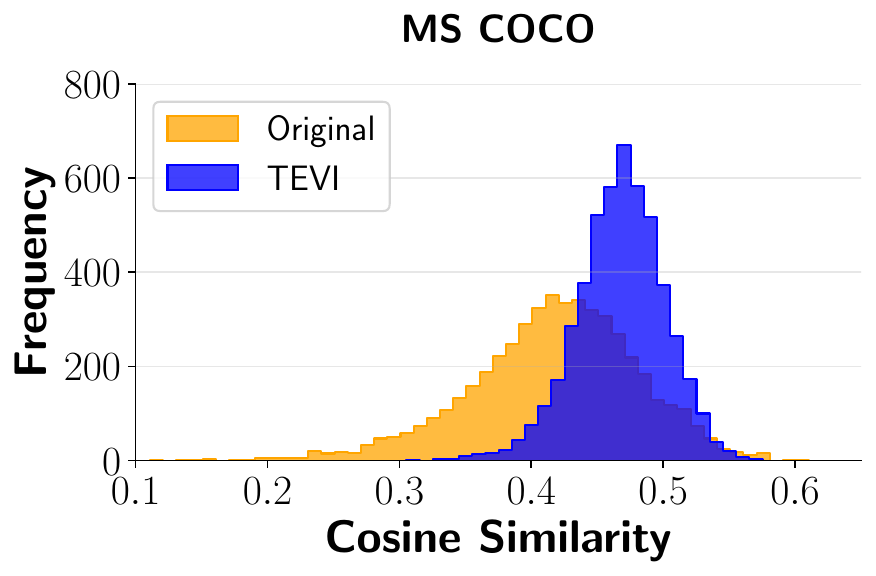}
    \end{subfigure}\hfill
    \begin{subfigure}[t]{0.24\linewidth}
        \includegraphics[trim={0 0 0 0cm},clip,width=\linewidth]{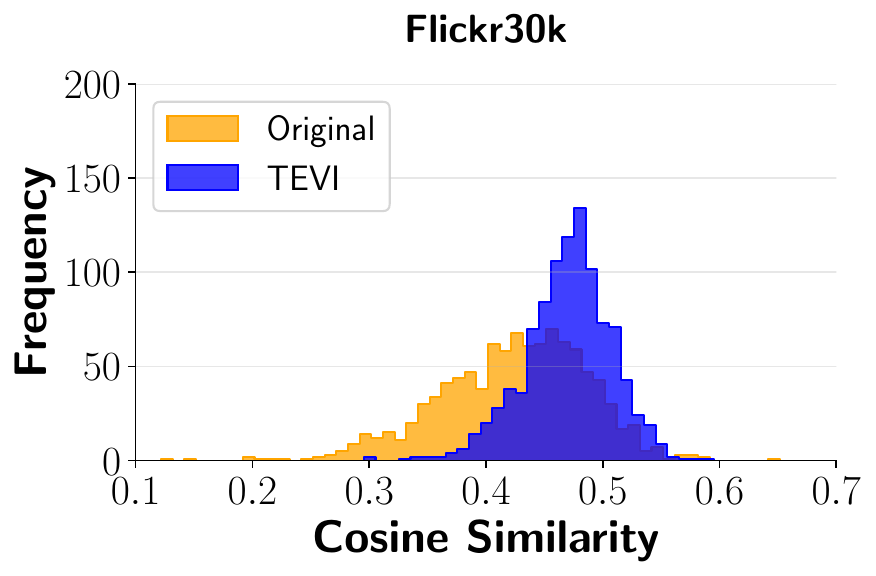}
    \end{subfigure}\hfill
    \begin{subfigure}[t]{0.24\linewidth}
        \includegraphics[trim={0 0 0 0cm},clip,width=\linewidth]{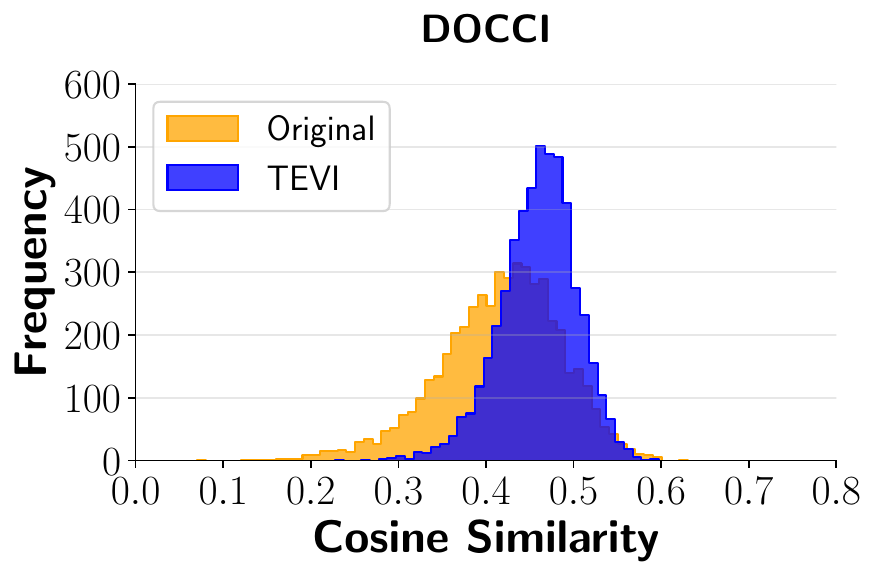}
    \end{subfigure}
    \hfill
    \begin{subfigure}[t]{0.24\linewidth}
        \includegraphics[trim={0 0 0 0cm},clip,width=\linewidth]{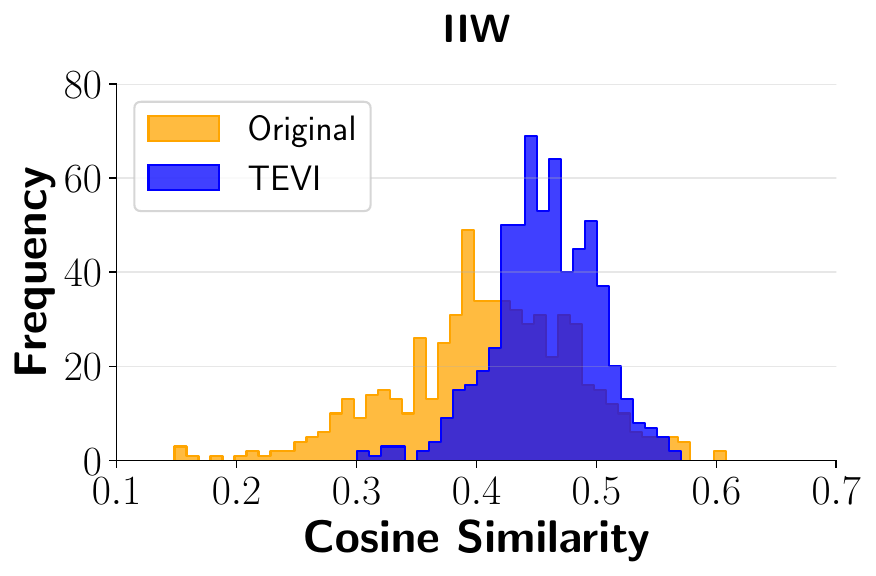}
    \end{subfigure}
    \caption{\textbf{Cross-modal alignment} across datasets using CLIP ViT-B/16. We find that the alignment between image-text pairs increases after applying \ours.
    }
    \label{fig:alignment_plot_cc12m}
\end{figure*}

%% file: figures/tex/supplement/alignment_plot_sharedclip.tex
\begin{figure*}[!ht]
    \centering
    \begin{subfigure}[t]{0.24\linewidth}
        \includegraphics[trim={0 0 0 0cm},clip,width=\linewidth]{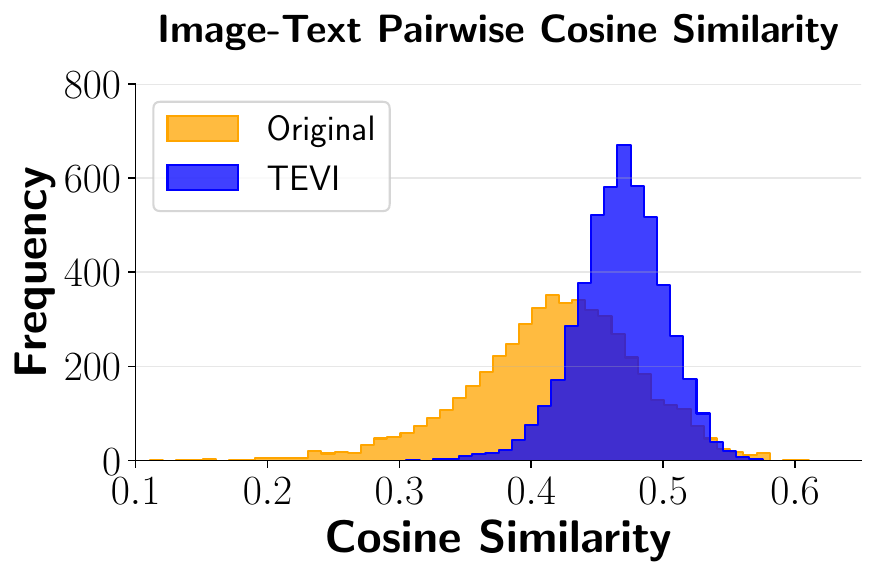}
    \end{subfigure}\hfill
    \begin{subfigure}[t]{0.24\linewidth}
        \includegraphics[trim={0 0 0 0cm},clip,width=\linewidth]{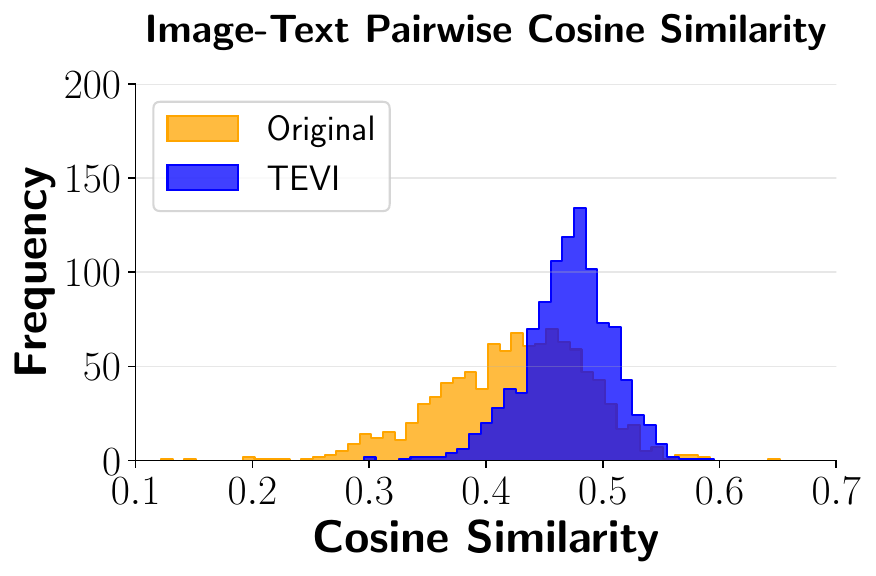}
    \end{subfigure}\hfill
    \begin{subfigure}[t]{0.24\linewidth}
        \includegraphics[trim={0 0 0 0cm},clip,width=\linewidth]{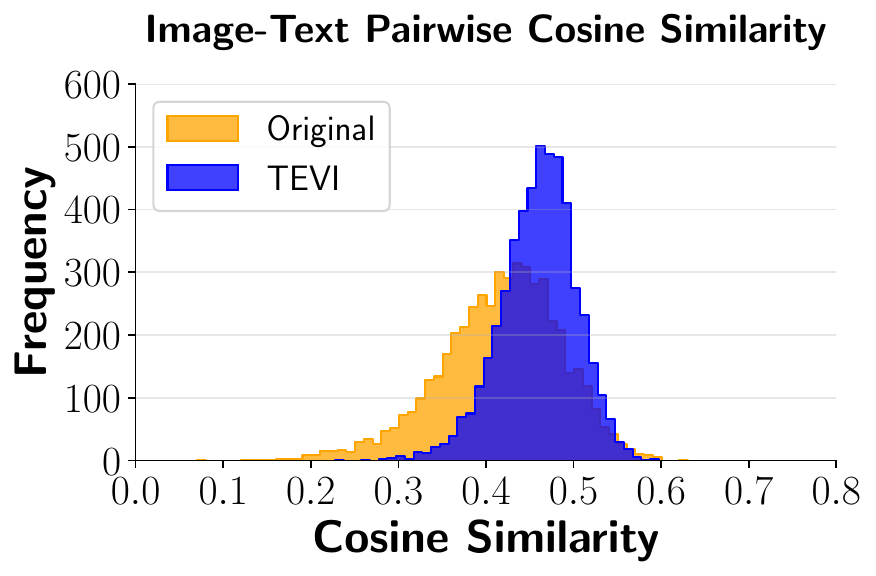}
    \end{subfigure}\hfill
    \begin{subfigure}[t]{0.24\linewidth}
        \includegraphics[trim={0 0 0 0cm},clip,width=\linewidth]{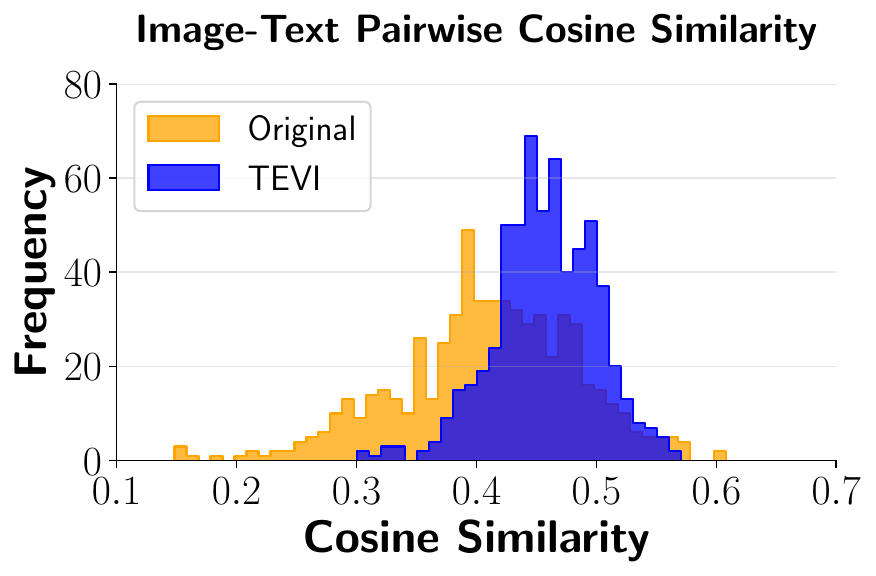}
    \end{subfigure}
    \caption{\textbf{Cross-modal alignment with SharedCLIP} on (1) MSCOCO \cite{lin2014microsoft}, (2) Flickr30k \cite{plummer2015flickr30k}, (3) DOCCI \cite{onoe2024docci}, (4) IIW \cite{garg2024imageinwords}. The alignment is measured by the cosine similarity between the positive image-text pairs, and the y-axis denotes the number of data points for each alignment score. We see the distribution after applying \ours (\textcolor{blue}{blue}) shifts to the right as compared to the baseline (\textcolor{orange}{orange}), showing improved alignment.
    }
    \label{fig:alignment_plot_sharedclip}
\end{figure*}

\begin{figure*}[!t]
    \centering
    \begin{subfigure}[t]{0.24\linewidth}
        \includegraphics[trim={0 0 0 0cm},clip,width=\linewidth]{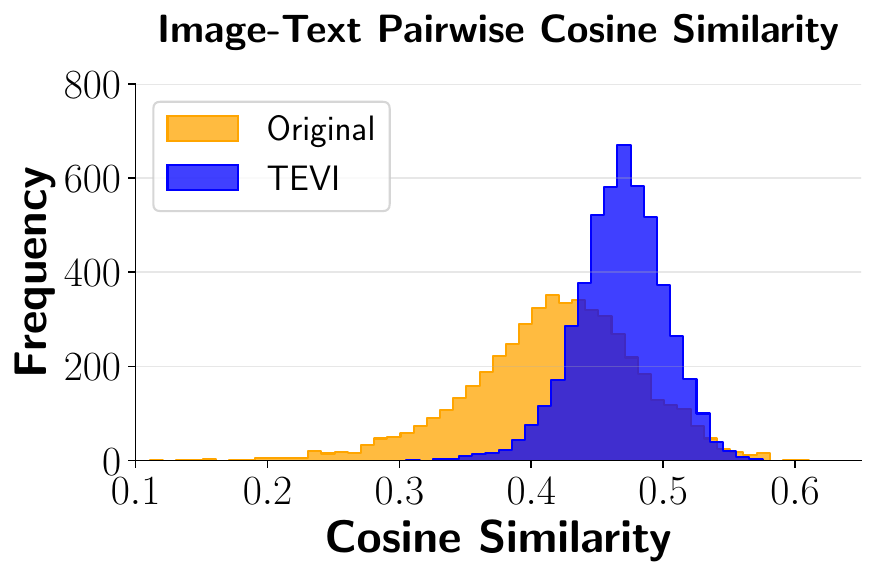}
    \end{subfigure}\hfill
    \begin{subfigure}[t]{0.24\linewidth}
        \includegraphics[trim={0 0 0 0cm},clip,width=\linewidth]{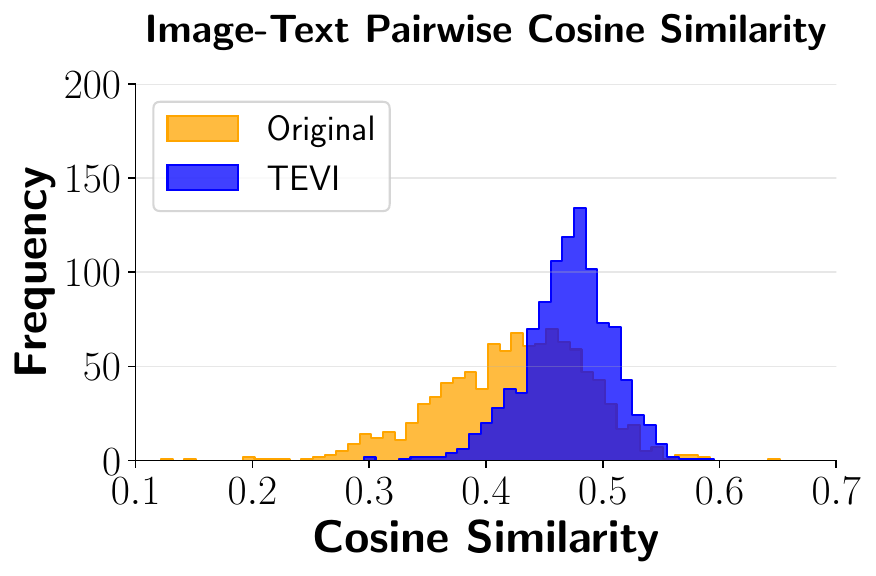}
    \end{subfigure}\hfill
    \begin{subfigure}[t]{0.24\linewidth}
        \includegraphics[trim={0 0 0 0cm},clip,width=\linewidth]{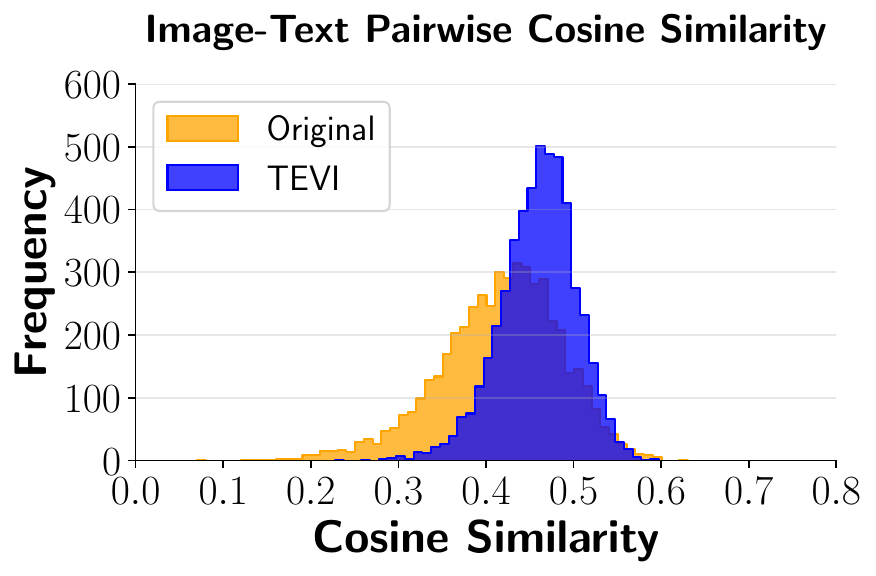}
    \end{subfigure}\hfill
    \begin{subfigure}[t]{0.24\linewidth}
        \includegraphics[trim={0 0 0 0cm},clip,width=\linewidth]{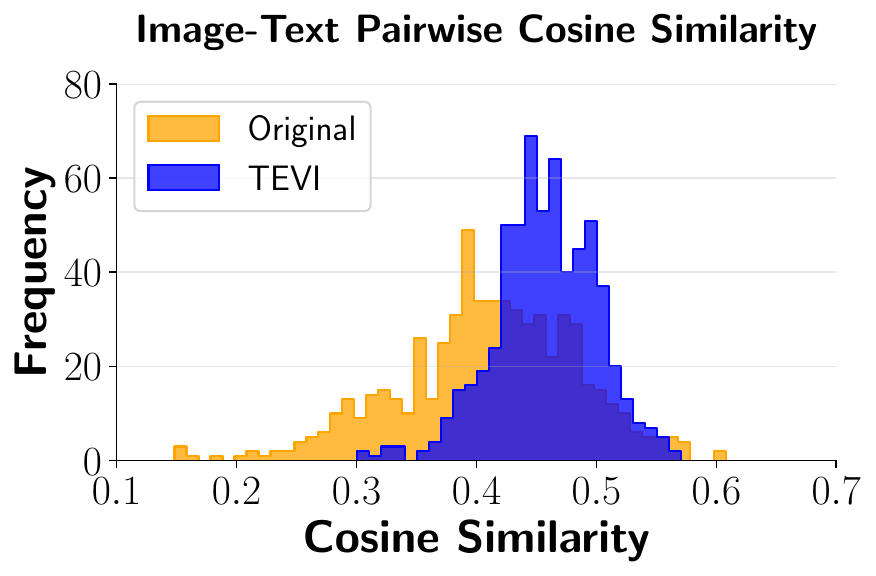}
    \end{subfigure}
    \caption{\textbf{Cross-modal alignment with AlignCLIP} on (1) MSCOCO \cite{lin2014microsoft}, (2) Flickr30k \cite{plummer2015flickr30k}, (3) DOCCI \cite{onoe2024docci}, (4) IIW \cite{garg2024imageinwords}. The alignment is measured by the cosine similarity between the positive image-text pairs, and the y-axis denotes the number of data points for each alignment score. We see the distribution after applying \ours (\textcolor{blue}{blue}) shifts to the right as compared to the baseline (\textcolor{orange}{orange}), showing improved alignment.
    }
    \label{fig:alignment_plot_alignclip}
\end{figure*}

%% file: tables/supplement/mask_analysis_page.tex
\input{tables/supplement/mask_sparsity}
\input{tables/supplement/ablation_full}
\input{tables/supplement/coco_ordering_i2i}

%% file: tables/supplement/mask_sparsity.tex
\begin{table}[!htbp]
\footnotesize
\setlength{\tabcolsep}{4pt}
    \centering
    \caption{\textbf{Mask sparsity.} Active latents retained and coefficient of variation of the learned mask.}
    \begin{tabular}{@{}lcc@{}}
        \toprule
        Split & active latents kept & coeff.\ of variation \\
        \midrule
        MS COCO   & \phantom{0}8.15\std{4.24} & 0.456\std{0.062} \\
        Flickr30k & \phantom{0}8.74\std{4.98} & 0.429\std{0.060} \\
        DOCCI     & 11.26\std{6.88} & 0.423\std{0.070} \\
        IIW       & 12.01\std{7.31} & 0.417\std{0.066} \\
        \bottomrule
    \end{tabular}
    \label{supp:table:mask_sparsity}
\end{table}

%% file: tables/supplement/ablation_full.tex
\begin{table}[!htbp]
\footnotesize
\setlength{\tabcolsep}{5pt}
    \centering
    \caption{\textbf{Ablating mask latents by ranking.}}
    \begin{tabular}{@{}llccc@{}}
        \toprule
        Direction & $m$ & top & random & bottom \\
        \midrule
        \multirow{3}{*}{I $\rightarrow$ T (24.20)}
          & 32  & \textbf{8.62} & 15.58 & 22.94 \\
          & 64  & \textbf{0.80} & \phantom{0}7.24  & 16.64 \\
          & 128 & 0.02 & 0.02  & 0.02  \\
        \midrule
        \multirow{3}{*}{T $\rightarrow$ I (8.55)}
          & 32  & \textbf{4.76} & \phantom{0}6.00  & \phantom{0}8.30  \\
          & 64  & \textbf{2.11} & \phantom{0}3.75  & \phantom{0}7.22  \\
          & 128 & 0.01 & 0.01  & 0.01  \\
        \bottomrule
    \end{tabular}
    \label{supp:table:ablation_full}
\end{table}

%% file: tables/supplement/coco_ordering_i2i.tex
\begin{table}[!htbp]
\footnotesize
\setlength{\tabcolsep}{4pt}
    \centering
    \caption{\textbf{Conditioning helps image-to-image retrieval on MS COCO.}}
    \begin{tabular}{@{}lcccccc@{}}
        \toprule
        & & \multicolumn{3}{c}{$\Delta$} & \multicolumn{2}{c}{Classes} \\
        \cmidrule(lr){3-5} \cmidrule(lr){6-7}
        & Exp. & Orig. & no-SAE & \ours & no-SAE & \ours \\
        \midrule
        $\Delta_1$ & $\downarrow$ & 0.101 & 0.102 & \textbf{0.086} & 30/61 & \textbf{52/61} \\
        $\Delta_2$ & $\uparrow$   & 0.067 & 0.065 & \textbf{0.081} & 29/61 & \textbf{45/61} \\
        $\Delta_3$ & $\downarrow$ & 0.212 & 0.206 & \textbf{0.202} & \textbf{51/61} & 50/61 \\
        \bottomrule
    \end{tabular}
    \label{supp:table:coco_ordering_i2i}
\end{table}

%% file: figures/tex/supplement/coco_qualitative_page.tex
\begin{figure*}[!p]
    \centering
    \includegraphics[width=\linewidth]{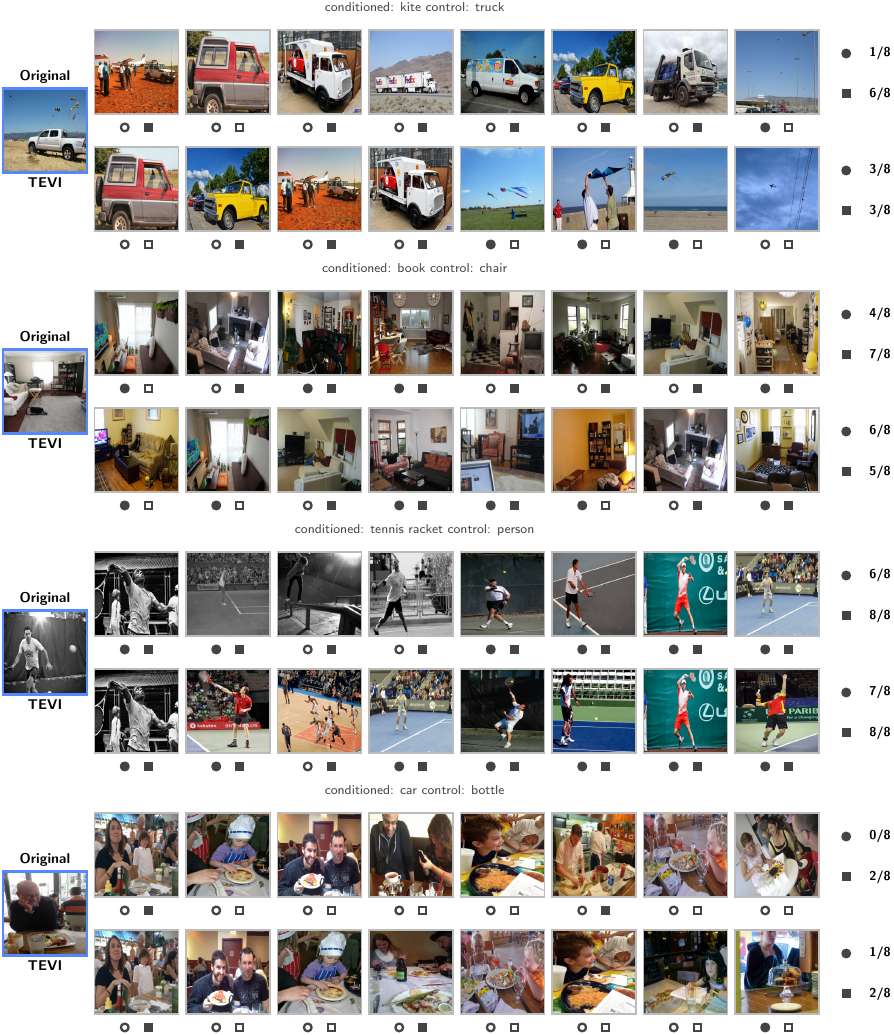}
    \caption{ \textbf{Conditioning helps retrieval on MS COCO captions given by attributes.} We show additional examples here. \cref{fig:coco_qualitative} shows one example at full size.
   }
    \label{fig:coco_qualitative_page}
\end{figure*}

%% file: figures/tex/supplement/quali_examples.tex
\begin{figure*}[!p]
\centering
\begin{subfigure}[c]{0.49\linewidth}
    \includegraphics[width=\linewidth]{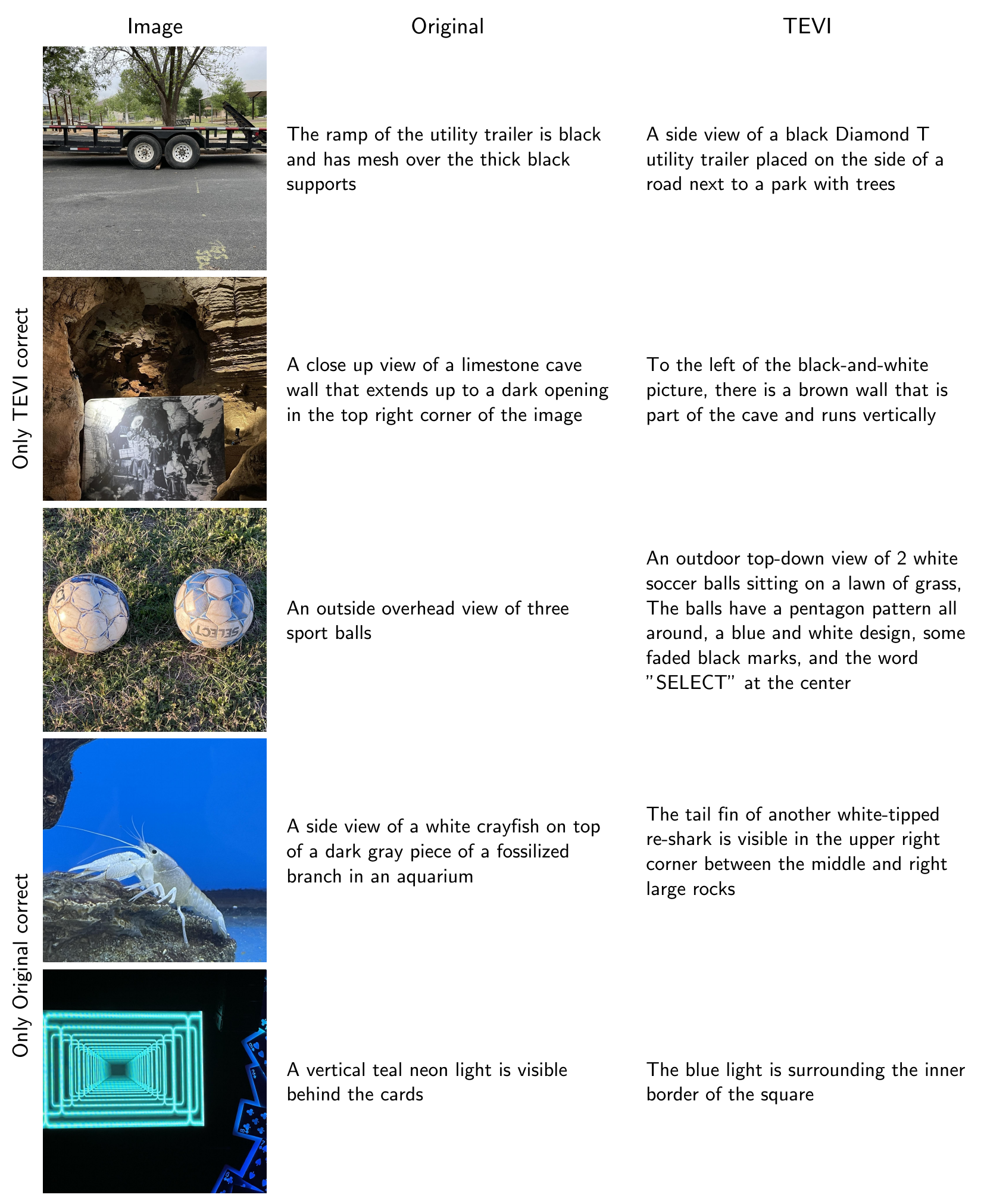}
    \caption{DOCCI}
\end{subfigure}
\hfill
\begin{subfigure}[c]{0.49\linewidth}
    \includegraphics[width=\linewidth]{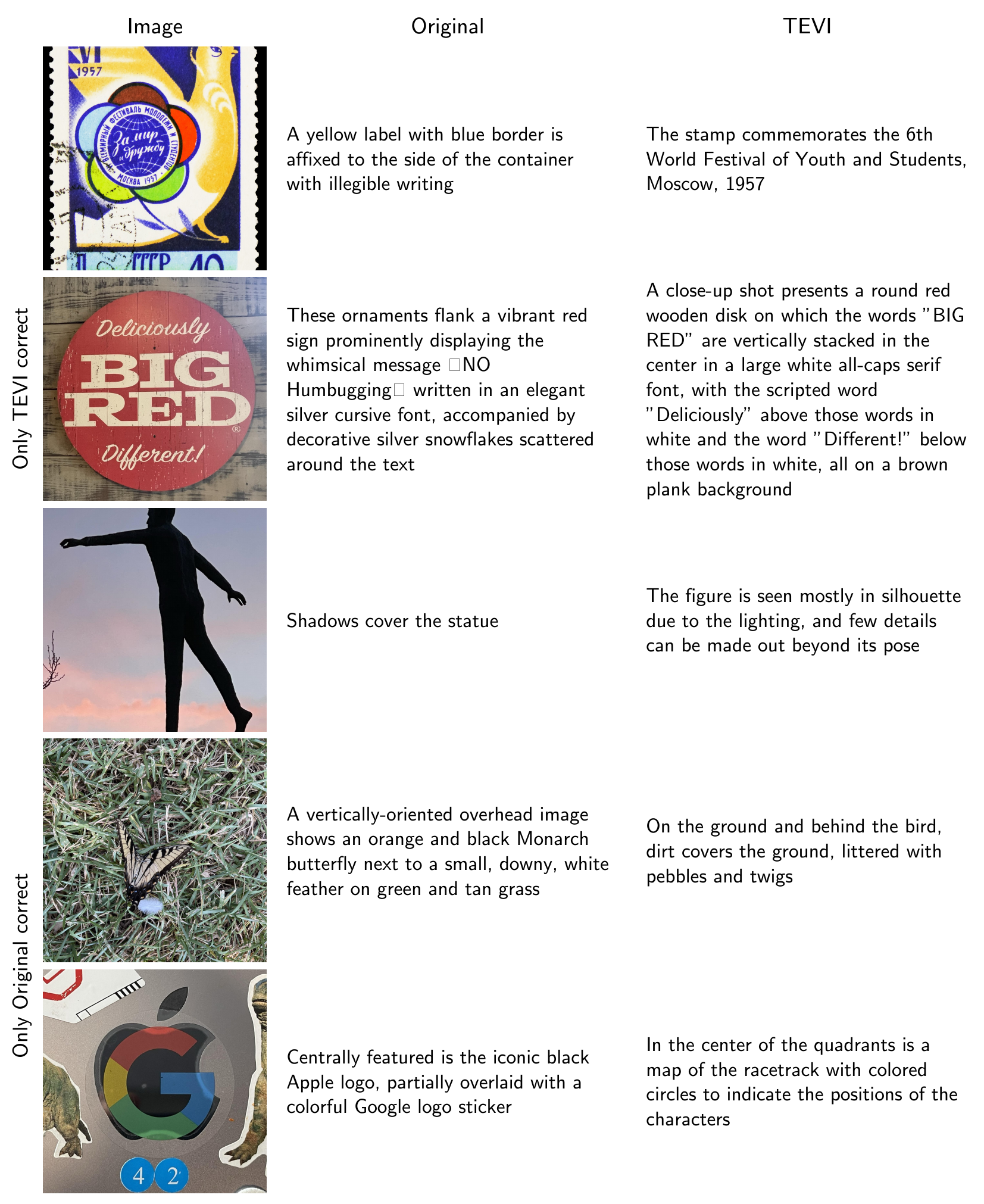}
    \caption{IIW}
\end{subfigure}
\hfill
\begin{subfigure}[c]{0.49\linewidth}
    \includegraphics[width=\linewidth]{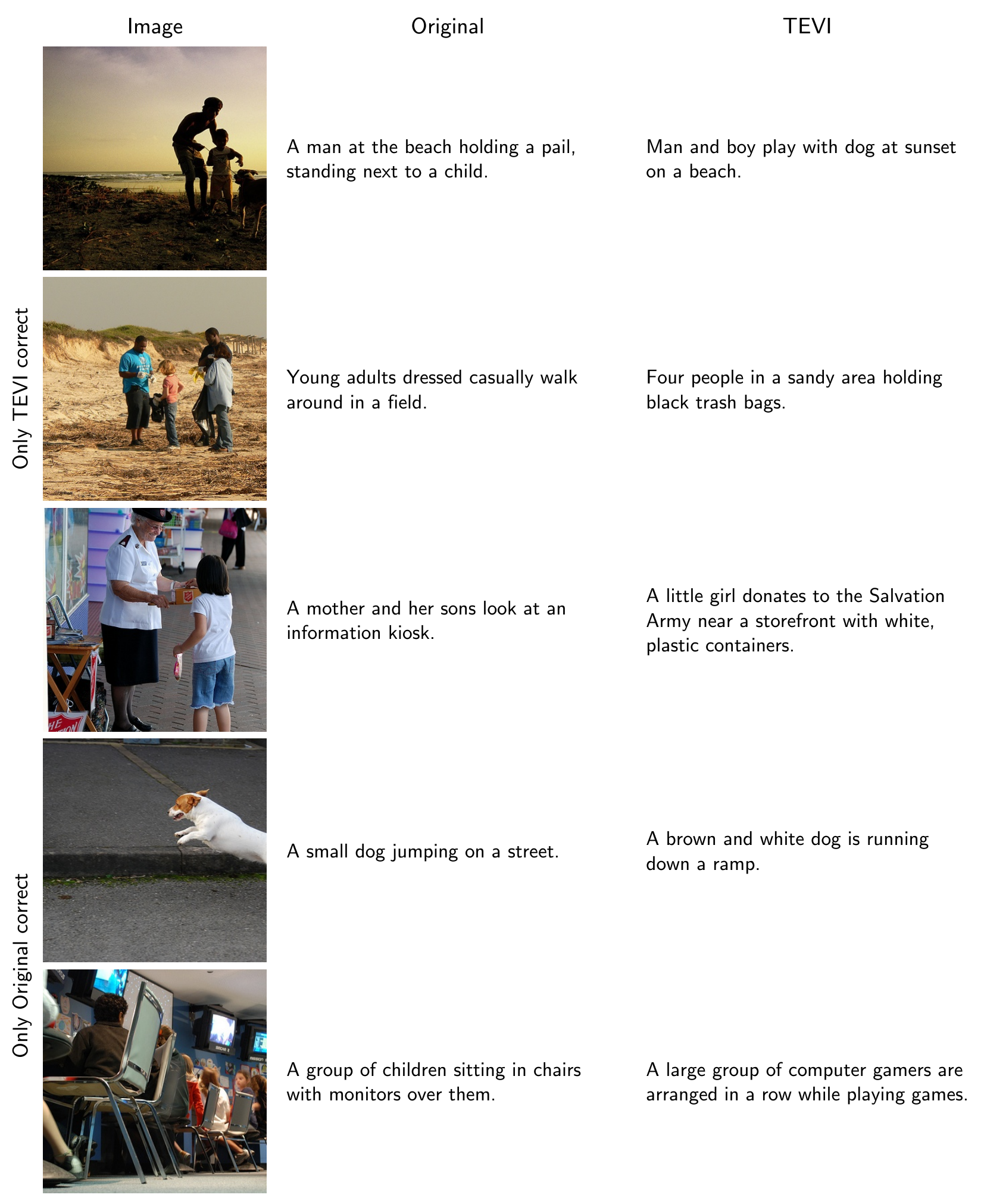}
    \caption{Flickr30k}
\end{subfigure}
\hfill
\begin{subfigure}[c]{0.49\linewidth}
    \includegraphics[width=\linewidth]{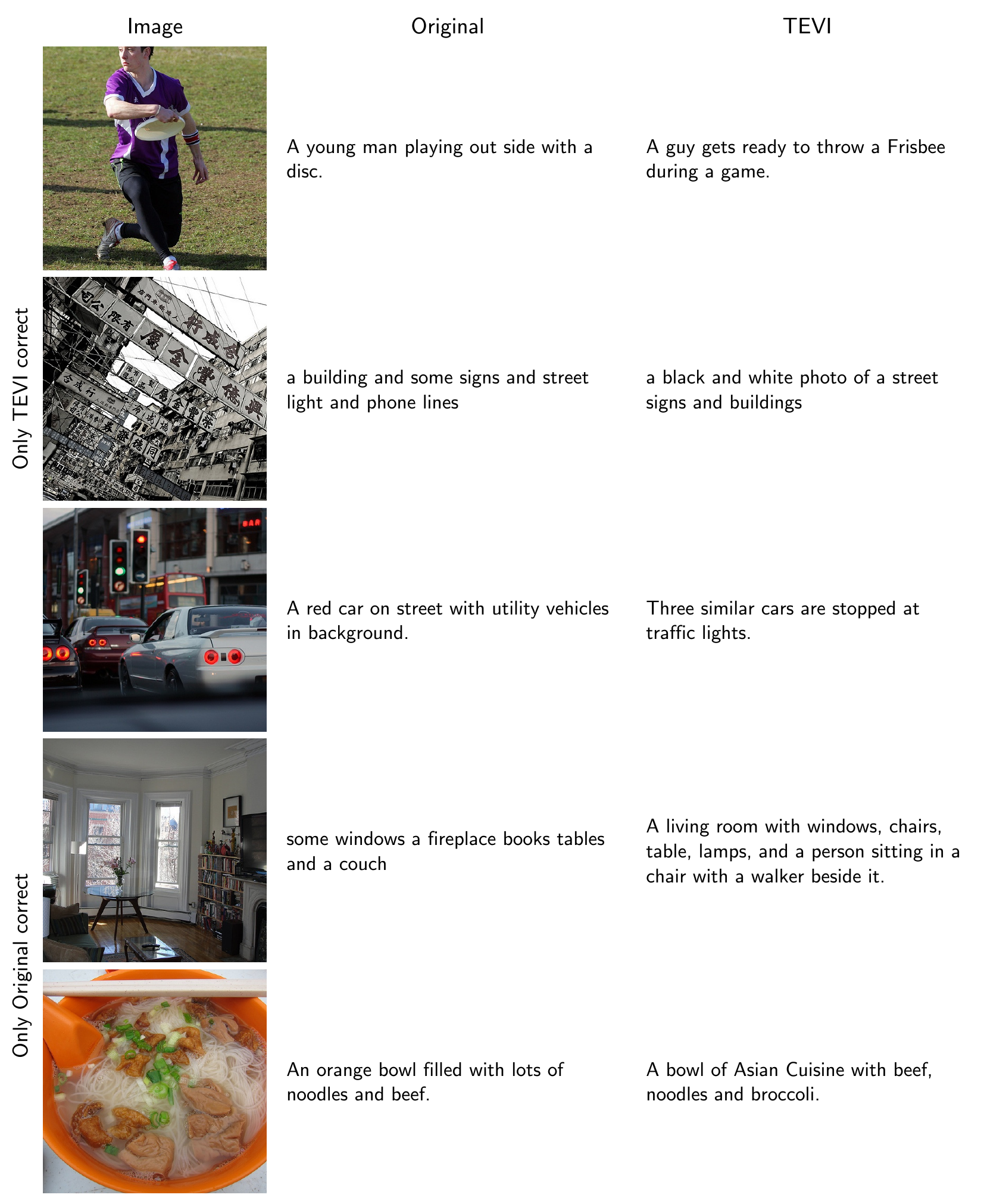}
    \caption{MS COCO}
\end{subfigure}
\caption{\textbf{Additional qualitative examples.} For each dataset, we show examples of both instances where \ours retrieves correctly when the baseline does not, and vice-versa. All examples are sampled randomly. }
\label{fig:supp:quali_examples}
\end{figure*}

%% file: supplement/limitations_broader_impact.tex
\balance
\section{Broader Impact}
\label{supp:sec:limitations:broader_impact}

The use of vision-language models for multimodal tasks is widespread, from tasks like image retrieval to visual question answering to image generation. This makes it increasingly important that such models work reliably. Lack of proper semantic alignment can lead to biased outputs~\cite{liang2022mind} or poor robustness to distribution shifts~\cite{eslami2024mitigate}. Improved visual-textual alignment can enhance downstream performance of such models and help alleviate the above mentioned drawbacks. Our work is fundamental research on addressing vision-language alignment and deals with relatively small-scale models, and therefore has low direct risk. However, like many multimodal frameworks, our methods could be co-opted for malicious purposes if deployed without safeguards.

\section{Licenses of Artifacts Used}
\label{supp:sec:licenses}
We adhere to the licenses provided by the artifacts we use in our work. OpenCLIP and RoCOCO are provided under the MIT license; SharedCLIP and AlignCLIP are released under the CC-BY-NC-ND-4.0 license; DOCCI, MS COCO, IIW are under the CC-BY-4.0 license; and SmartCLIP is released under the Apache 2.0 license. Flickr30k permits use for non-commercial research purposes. Our usage of all artifacts is in accordance with their allowed and intended use.